# Improved SVRG for Non-Strongly-Convex or Sum-of-Non-Convex Objectives

Zeyuan Allen-Zhu
zeyuan@csail.mit.edu
Princeton University

Yang Yuan
yangyuan@cs.cornell.edu
Cornell University

February 5, 2016*

**Abstract**

Many classical algorithms are found until several years later to outlive the confines in which they were conceived, and continue to be relevant in unforeseen settings. In this paper, we show that SVRG is one such method: being originally designed for strongly convex objectives, it is also very robust in non-strongly convex or sum-of-non-convex settings.

More precisely, we provide new analysis to improve the state-of-the-art running times in both settings by either applying SVRG or its novel variant. Since non-strongly convex objectives include important examples such as Lasso or logistic regression, and sum-of-non-convex objectives include famous examples such as stochastic PCA and is even believed to be related to training deep neural nets, our results also imply better performances in these applications.

## 1 Introduction

The fundamental algorithmic problem in optimization is to design efficient algorithms for solving certain *classes* of problems. By distinguishing between smooth and non-smooth functions, between weakly-convex and strongly-convex functions, between proximal and non-proximal functions, or even between convex and non-convex functions, the number of classes grows exponentially and it may be unrealistic to design a new algorithm for each specific class. Taking into account such "design complexity", it is beneficial to design a single method the works for multiple classes, or perhaps even more beneficial if this method is already widely used and happens to outlive the confines it was originally designed for. Easier done in practice, providing a support *theory* unifying the underlying classes for a specific method is particularly exciting, challenging, and sometimes even enlightening: the theoretical findings may further suggest experimentalists regarding how such a method should be best tuned in practice.

In this paper, we revisit the SVRG method by Johnson and Zhang [13] and explore its applications to either a non-strongly convex objective, or a *sum-of-non-convex* objective, or even both. We show faster convergence results for minimizing such objectives by either directly applying SVRG or modifying it in a novel manner.

Consider the following composite convex minimization:

$$\min_{x\in\mathbb{R}^d}\Big\{F(x)\stackrel{\text{def}}{=}f(x)+\Psi(x)\stackrel{\text{def}}{=}\frac{1}{n}\sum_{i=1}^{n}f_i(x)+\Psi(x)\Big\}\ . \tag{1.1}$$

*The current version polishes the writing and adds more experiments.

Here, $f(x) = \frac{1}{n}\sum_{i=1}^n f_i(x)$ is a convex function that is written as a finite average of $n$ smooth functions $f_i(x)$,[1] and $\Psi(x)$ is a relatively simple (but possibly non-differentiable) convex function, sometimes referred to as the *proximal* function. Suppose we are interested in finding an approximate minimizer $x \in \mathbb{R}^d$ satisfying $F(x) \le F(x^*) + \varepsilon$, where $x^*$ is a minimizer of $F(x)$.

**Examples.** Problems of this form arise in many places in machine learning, statistics, and operations research. For instance, many *regularized empirical risk minimization (ERM)* problems fall into this category with convex $f_i(\cdot)$. In such problems, we are given $n$ training examples $\{(a_1,\ell_1),\dots(a_n,\ell_n)\}$, where each $a_i \in \mathbb{R}^d$ is the feature vector of example $i$, and each $\ell_i \in \mathbb{R}$ is the label of example $i$. The following classification and regression problems are well-known examples of ERM:

- Ridge Regression: $f_i(x) = \frac{1}{2}(\langle a_i, x\rangle - \ell_i)^2 + \frac{\sigma}{2}\|x\|_2^2$ and $\Psi(x) = 0$.
- Lasso: $f_i(x) = \frac{1}{2}(\langle a_i, x\rangle - \ell_i)^2$ and $\Psi(x) = \sigma\|x\|_1$.
- $\ell_1$-Regularized Logistic Regression: $f_i(x) = \log(1 + \exp(-\ell_i\langle a_i, x\rangle))$ and $\Psi(x) = \sigma\|x\|_1$.

Another important problem that falls into this category is the *principle component analysis (PCA)* problem. Suppose we are given $n$ data vectors $a_1,\dots,a_n \in \mathbb{R}^d$, denoting by $A = \frac{1}{n}\sum_{i=1}^n a_i a_i^T$ the normalized covariance matrix, Garber and Hazan [8] showed that approximately finding the principle component of $A$ is equivalent to minimizing $f(x) = \frac{1}{2}x^T(\mu I - A)x$ for some suitably chosen parameter $\mu > 0$. Therefore, defining $f_i(x) \stackrel{\text{def}}{=} \frac{1}{2}x^T(\mu I - a_i a_i^T)x$ and $\Psi(x) = 0$, this falls into Problem (1.1) with non-convex functions $f_i(\cdot)$.

**Background of SVRG.** Stochastic first-order methods perform the following updates to solve Problem (1.1):

$$x_{t+1} \leftarrow \arg\min_{y\in\mathbb{R}^d}\Big\{\frac{1}{2\eta}\|y - x_t\|_2^2 + \langle \xi_t, y\rangle + \Psi(y)\Big\} ,$$

where $\eta$ is the step length, and $\xi_t$ is a random vector satisfying $\mathbb{E}[\xi_t] = \nabla f(x_t)$ which is referred to as the *stochastic gradient*. If the proximal function $\Psi(y)$ equals zero, the update simply reduces to $x_{t+1} \leftarrow x_t - \eta\xi_t$.

Given the "finite average" structure $f(x) = \frac{1}{n}\sum_{i=1}^n f_i(x)$, a classical choice is to set $\xi_t = \nabla f_i(x_t)$ for some random index $i \in [n]$ per iteration. Methods based on this choice are known as *stochastic gradient descent (SGD)*.

More recently, the convergence speed of SGD has been further improved with the *variance-reduction* technique [5, 13, 18, 21, 24, 25, 29]. In all of these cited results, the authors have, in one way or another, shown that SGD can converge much faster if one makes a better choice of the stochastic gradient $\xi_t$, so that its variance $\mathbb{E}[\|\xi_t - \nabla f(x_t)\|_2^2]$ reduces as $t$ increases.

One particular way to reduce the variance is the SVRG method described as follows [13]. Keep a snapshot $\widetilde{x} = x_t$ after every $m$ stochastic update steps (where $m$ is some parameter), and compute the full gradient $\nabla f(\widetilde{x})$ only for such snapshots. Then, set $\xi_t = \nabla f_i(x_t) - \nabla f_i(\widetilde{x}) + \nabla f(\widetilde{x})$ as the stochastic gradient. One can verify that, under this choice of $\xi_t$, it satisfies $\mathbb{E}[\xi_t] = \nabla f(x_t)$ and $\lim_{t\to\infty}\mathbb{E}[\|\xi_t - \nabla f(x_t)\|_2^2] = 0$.

**Non-Strongly Convex Objectives.** Although many variance-reduction based methods have been proposed, most of them, including SVRG, only has convergence guarantee of Problem (1.1) when the objective $F(x)$ is strongly convex. However, in many machine learning applications, $F(x)$ is simply *not* strongly convex. This is particularly true for Lasso [28] and $\ell_1$-Regularized Logistic Regression [20], two cornerstone problems extensively used for feature selections.

[1] In fact, even if each $f_i(x)$ is not smooth but only Lipschitz continuous, standard smoothing techniques such as Chapter 2.3 of [11] can make each $f_i(x)$ smooth without sacrificing too much accuracy.

One way to get around this is to add a dummy regularizer $\frac{\lambda}{2}\|x\|_2^2$ to $F(x)$, and then apply any of the above methods. However, the weight of this regularizer, $\lambda$, needs to be chosen before the algorithm starts. This adds a lot of difficulty when applying such methods to real life: (1) one needs to tune $\lambda$ by repeatedly executing the algorithm, and (2) the error of the algorithm does not converge to zero as time goes (in fact, it converges to $O(\lambda)$ so one needs to know the desired accuracy before the algorithm starts). Perhaps more importantly, adding the dummy regularizer hurts the performance of the algorithm both in theory and practice.

Another possible solution is to tackle the non-strongly convex case *directly* [5, 18, 21], without using any dummy regularizer. These methods are the so-called *anytime* algorithms: they can be interrupted at any time, and the training error tends to zero as the number of iterations increases.

While direct methods are much more convenient for practical uses, existing direct methods are much slower than indirect methods (i.e., methods via dummy regularization) at least in theory. More specifically, if the desired accuracy is $\varepsilon$ and the smoothness of each $f_i(x)$ is $L$, then the *gradient complexities* [2] of the best known direct and indirect methods are respectively

$$O\Big(\frac{n+L}{\varepsilon}\Big) \qquad \text{and} \qquad O\Big((n+\frac{L}{\varepsilon})\log\frac{1}{\varepsilon}\Big) \ .$$

Therefore in theory, when $n$ is usually dominating, indirect methods are faster but less convenient, while direct methods are slower but more convenient.

In this paper, we propose $\texttt{SVRG}^{++}$, a new method that solves the non-strongly convex case of Problem (1.1) *directly* with gradient complexity $O(n\log\frac{1}{\varepsilon}+\frac{L}{\varepsilon})$, outperforming both known direct and indirect methods. In particular, our complexity outperforms known direct methods (e.g., SAGA or SAG) by a factor $\widetilde{\Omega}(n/L)$ in the case when $L\leq n$. Since $L$ is usually on the order of $O(s)$ for large-scale machine learning problems where $s$ is the sparsity of feature vectors and $s$ can be much smaller than $n$, we claim that this outperformance may be significant in theory. On the practical side, $\texttt{SVRG}^{++}$ is a direct, anytime method, which is convenient to use. We describe $\texttt{SVRG}^{++}$ and the main techniques we use in Section 4.

**Sum-of-Non-Convex Objectives.** If $f(x)$ is $\sigma$-strongly convex while each $f_i(x)$ is non-convex but $L$-smooth, Shalev-Shwartz discovered that the SVRG method admits a gradient complexity of $O\big((n+\frac{L^2}{\sigma^2})\log\frac{1}{\varepsilon}\big)$ for minimizing $F(x)$ [22] in the case of $\Psi(x)=0$. A similar result has been independently re-discovered by Garber and Hazan [8] and applied to the PCA problem. This setting is also believed to be happening (at least locally) on training deep neural nets [3, 13, 22].

Despite the missing proximal term $\Psi(x)$ in their analysis, the running time above is imperfect for two reasons.

- First, this complexity is not stable: even if we modify only one of $f_i(x)$ from convex to (a little bit) non-convex, the best known gradient complexity for SVRG immediately worsens to $O\big((n+\frac{L^2}{\sigma^2})\log\frac{1}{\varepsilon}\big)$ from $O\big((n+\frac{L}{\sigma})\log\frac{1}{\varepsilon}\big)$. In contrast, one should expect a more graceful decay of the performance as a function on the "magnitude" of the non-convexity, or perhaps even a *threshold* where the performance is totally unaffected if the magnitude is "below" this threshold.
- Second, the complexity does not take into account the asymmetry in smoothness. For instance, in PCA applications, each $f_i(x)$ can be very non-convex and its Hessian has eigenvalues between $-l < 0$ and $L > 0$ where $l$ can be significantly larger than $L$. Can we take advantage of this asymmetry to get better running time?

[2]Throughout this paper, we will use *gradient complexity* as an effective measure of an algorithm's running time. Usually, the total running time of an algorithm is $O(d)$ multiplied with its gradient complexity, because each $\nabla f_i(x)$ can be computed in $O(d)$ time.

In this paper, we prove that if each $f_i(x)$ is $L$-upper smooth and $l$-lower smooth (which means the Hessian of $f_i(x)$ has eigenvalues bounded between $[-l, L]$), the same SVRG method admits a gradient complexity of $O\big((n + \frac{L}{\sigma} + \frac{Ll}{\sigma^2}) \log \frac{1}{\varepsilon}\big)$. This resolves both our aforementioned concerns. First, if $l = O(\sigma)$, our new result suggests that the convergence of SVRG is asymptotically the *same* as the convex case, meaning there is a threshold $O(\delta)$ that SVRG allows each $f_i(x)$ to be non-convex below this threshold for free. Second, in the $l > L$ case, our result implies a linear dependence on the non-convexity parameter $l$, rather than the quadratic one $O\big((n + \frac{l^2}{\sigma^2}) \log \frac{1}{\varepsilon}\big)$ shown by prior work [8, 22]. To the best of our knowledge, this is the first time that upper and lower smoothness parameters are distinguished in order to prove convergence results for minimizing (1.1).

Our improvement on SVRG immediately leads to faster stochastic algorithms for PCA [8, 27]. Assume that $A = \frac{1}{n}\sum_{i=1}^n a_i a_i^T$ is a normalized covariance matrix where each $a_i \in \mathbb{R}^d$ has Euclidean norm at most 1. Let $\lambda \in [0, 1]$ be the largest eigenvalue of $A$. Garber and Hazan showed that computing the leading eigenvector of $A$ is, up to binary search preprocessing, equivalent to the sum-of-non-convex form of Problem (1.1), with upper smoothness $L = \lambda$ and lower smoothness $l = 1$.[3] Garber and Hazan further applied SVRG to minimize this objective and proved an overall running time $O\big((nd + \frac{d}{\delta^2}) \log \frac{1}{\varepsilon}\big)$. Our result improves this running time to $O\big((nd + \frac{\lambda d}{\delta^2}) \log \frac{1}{\varepsilon}\big)$. Since $\lambda$ may be as small as $1/d$, this speed up is significant in theory.[4]

Since the original publication of this paper, our above PCA speed-up has also been translated to $k$-SVD, which is to compute the first $k$ singular vectors of a given matrix [4].

Our results above are non-accelerated for the sum-of-non-convex setting. One can apply Catalyst [7, 15] to further improve its running time when $\sigma$ is very small. Not surprisingly, our performance improvement carries to the accelerated setting as well.

Finally, we also prove that our proposed improvements on SVRG (for non-strongly convex objectives and for sum-of-non-convex objectives) can be put together, leading to a new algorithm $\mathtt{SVRG}^{++}_{\mathtt{nc}}$ that works for both non-strongly convex and sum-of-non-convex objectives. This gives faster algorithms than known results as well.

**Roadmap.** We discuss related work in Section 2 and provide notational background in Section 3. We state our result for non-strongly convex objectives in Section 4, for sum-of-non-convex objectives in Section 5 and 6, and for both non-strongly convex and sum-of-non-convex objectives in Section 7. In Section 8 and Section 9 we perform experiments supporting our theory. Most of the technical proofs are included in the appendix.

## 2 Other Related Work

The first published variance-reduction method is SAG [21]. SAG obtains an $O(\log(1/\varepsilon))$ convergence (i.e., linear convergence) for strongly convex and smooth objectives, comparing to the $O(1/\varepsilon)$ rate of SGD [12, 23]. This $O(\log(1/\varepsilon))$ rate has also been obtained by several concurrent or subsequent works, such as SVRG, MISO and SAGA [5, 13, 18]. SDCA [25] has also been discovered to be intrinsically performing some "variance reduction" procedure [22].

---

[3] Suppose that the eigengap between largest and second largest eigenvalues of $A$ is $\delta = \lambda - \lambda_2$. Garber and Hazan showed that computing the principle component of $A$ is, up to binary search preprocessing, equivalent to minimizing the objective $f(x) \stackrel{\text{def}}{=} \frac{1}{2}x^T(\mu I - A)x + b^T x$ where $\mu = \lambda + \delta$. If one defines $f_i(x) \stackrel{\text{def}}{=} \frac{1}{2}x^T(\mu I - a_i a_i^T)x + b^T x$, this minimization problem falls into the sum-of-non-convex setting of Problem (1.1), with upper smoothness $L = \mu \approx \lambda$ and lower smoothness $l = 1$.

[4] Garber and Hazan also applied acceleration schemes on top of SVRG, and obtained a running time $\widetilde{O}(\frac{n^{3/4}d}{\sqrt{\delta}})$. We can do the same thing here and improve their running time to $\widetilde{O}(\frac{n^{3/4}\lambda^{1/4}d}{\sqrt{\delta}})$ in the accelerated setting.

Among the variance-reduction algorithms, only SAG, MISO, and SAGA can provide theoretical guarantees for directly solving non-strongly convex objectives (i.e., without adding a dummy regularizer). The best gradient complexity for direct methods before our work is $O(\frac{n+L}{\varepsilon})$ due to SAG and SAGA. On the other hand, if one uses indirect methods, the best gradient complexity is $O\big((n+\frac{L}{\varepsilon})\log\frac{1}{\varepsilon}\big)$, where the asymptotic dependence on $\varepsilon$ is weakened to $\frac{\log(1/\varepsilon)}{\varepsilon}$.

We work directly with smooth functions $f_i(x)$ rather than the more structured $f_i(x) \stackrel{\text{def}}{=} \phi_i(\langle x, a_i\rangle)$. In the structured case, AccSDCA [26], along with subsequent works [16, 31], obtains a slightly better gradient complexity $O\big(\big(n+\min\big\{L/\varepsilon, \sqrt{nL/\varepsilon}\big\}\big)\log\frac{1}{\varepsilon}\big)$ for non-strongly convex objectives. This class of methods require one to work with the dual of the objective, require one to add dummy regularizer for non-strongly convex objectives (i.e., are indirect), and run only faster than the variance-reduction based methods when $n < \sqrt{L/\varepsilon}$.

Since the original submission of this paper, we learned several other related works from the anonymous reviewers. First, the SVRG method was independently discovered and published also by [30]. Second, the result of [17] also uses doubling-epoch technique and can partially infer our results on $\texttt{SVRG}^{++}$ with a slightly more complicated proof and different algorithm.[5] Third, in a concurrent accepted paper to this ICML, Garber et al. [9] improved the original Garber-Hazan PCA result [8] and thus solved a special case of our Theorem 6.1; their result has nothing to do with other theorems in this paper, especially Theorem 5.1 and 7.1.[6]

In some concurrent works, the authors of [2] obtained our same running time on $\texttt{SVRG}^{++}$ through reductions. However, their algorithm is not a direct one so cannot be practically as good as $\texttt{SVRG}^{++}$. Also after this paper is accepted, the author of [1] provided a direct method for solving (1.1) but in an accelerated speed. As mentioned in [1], his method can be combined with the technique in this paper to obtain a non-strongly convex accelerated running time.

## 3 Notations

Throughout this paper, we denote by $\|\cdot\|$ the Euclidean norm. We assume that each $f_i(\cdot)$ is differentiable and $\Psi(\cdot)$ is convex and lower semicontinuous.

We say that a differentiable function $f_i(\cdot)$ is *L-smooth* (or has $L$-Lipschitz continuous gradient) if:

$$\|\nabla f_i(x) - \nabla f_i(y)\| \le L\|x-y\| \quad \forall x, y \in \mathbb{R}^d \enspace .$$

The above definition has several equivalent forms, and one of them says for all $x, y \in \mathbb{R}^d$:

$$-\frac{L}{2}\|y-x\|^2 \le f(y) - \big(f(x) + \langle \nabla f(x), y-x\rangle\big) \le \frac{L}{2}\|y-x\|^2 \enspace .$$

In this paper, we say $f_i(\cdot)$ is $L$-upper smooth if it satisfies

$$f(y) - \big(f(x) + \langle \nabla f(x), y-x\rangle\big) \le \frac{L}{2}\|y-x\|^2 \quad \forall x, y \in \mathbb{R}^d \enspace ,$$

[5] Mahdavi et al. studied an oracle model where there are two gradient oracles, a stochastic one and a full-gradient one. Then, they prove comparable bounds to $\texttt{SVRG}^{++}$ but without supporting proximal terms and therefore do not directly apply to ERM problems such as Lasso or logistic regression.

[6] For the PCA problem, they produced the same $O\big((nd+\frac{\lambda d}{\delta^2})\log\frac{1}{\varepsilon}\big)$ running time as we do; however, their result is only about PCA so does not solve general sum-of-non-convex objectives; they also did not introduce upper or lower smoothness like we do.

and $f_i(\cdot)$ is $l$-lower smooth if it satisfies

$$f(y) - \big(f(x) + \langle \nabla f(x), y - x\rangle\big) \geq -\frac{l}{2}\|y-x\|^2 \quad \forall x, y \in \mathbb{R}^d \ .$$

Let us give a few examples: a convex differentiable function is 0-lower smooth; an $L$-smooth function is $L$-upper and $L$-lower smooth; a convex $L$-smooth function is $L$-upper and 0-lower smooth.

We say a function $f(\cdot)$ is $\sigma$-strongly convex if

$$f(y) - \big(f(x) + \langle \nabla f(x), y - x\rangle\big) \geq \frac{\sigma}{2}\|y-x\|^2 \quad \forall x, y \in \mathbb{R}^d \ .$$

Note that for a twice differentiable function $f$, the above definitions are equivalent to the corresponding statements about the eigenvalues of $\nabla^2 f(x)$. Indeed, $L$-upper smoothness is equivalent to saying all eigenvalues are no more than $L$, $l$-lower smoothness is equivalent to saying all eigenvalues are no less than $-l$, and $\sigma$-strong convexity is saying all eigenvalues are at least $\sigma$.

## 4 $\texttt{SVRG}^{++}$ for Non-Strongly Convex Objectives

In this section we consider the case of Problem (1.1) when each $f_i(x)$ is a convex function and the objective is not necessarily strongly convex. Recall that this class of problems include Lasso and logistic regression as notable examples.

We propose our $\texttt{SVRG}^{++}$ algorithm for solving this case, see Algorithm 1. Given an initial vector $x^\phi$, our algorithm is divided into $S$ epochs. The $s$-th epoch consists of $m_s$ stochastic gradient steps (see Line 8 of $\texttt{SVRG}^{++}$), where $m_s$ doubles between every consecutive two epochs. This "doubling" feature distinguishes our method from all of the cited variance-reduction based methods.

Within each epoch, similar to SVRG, we compute the full gradient $\widetilde{\mu}_{s-1} = \nabla f(\widetilde{x}^{s-1})$ where $\widetilde{x}^{s-1}$ is the average point of the previous epoch. We then use $\widetilde{\mu}_{s-1}$ to define the variance-reduced stochastic gradient $\xi$, see Line 7 of $\texttt{SVRG}^{++}$. Unlike SVRG, our starting vector $x_0^s$ of each epoch is set to be the ending vector $x_{m_{s-1}}^{s-1}$ of the previous epoch, rather than the average of the previous epoch.[7]

We state our main result for $\texttt{SVRG}^{++}$ as follows:

**Theorem 4.1.** *If each $f_i(x)$ is convex in Problem (1.1), then $\texttt{SVRG}^{++}(x^\phi, m_0, S, \eta)$ satisfies if $m_0$ and $S$ are positive integers and $\eta = 1/(7L)$, then*

$$\mathbb{E}[F(\widetilde{x}^S) - F(x^*)] \leq O\Big(\frac{F(x^\phi) - F(x^*)}{2^S} + \frac{L\|x^\phi - x^*\|^2}{2^S m_0}\Big) \ . \tag{4.1}$$

*In addition, $\texttt{SVRG}^{++}$ has a gradient complexity of $O(S \cdot n + 2^S \cdot m_0)$.*

As a result, given an initial vector $x^\phi$ satisfying $\|x^\phi - x^*\|^2 \leq \Theta$ and $F(x^\phi) - F(x^*) \leq \Delta$ for parameters $\Theta, \Delta \in \mathbb{R}_+$, by setting $S = \log_2(\Delta/\varepsilon)$, $m_0 = L\Theta/\Delta$, and $\eta = 1/(7L)$, we obtain an $O(\varepsilon)$ approximate minimizer of $F(\cdot)$ with a total gradient complexity $O\big(n \log\big(\frac{\Delta}{\varepsilon}\big) + \frac{L\Theta}{\varepsilon}\big)$.

[7] The theoretical convergence of SVRG relies on its Option II, that is to set the beginning vector of each epoch to be the *average* (or a random) vector of the previous epoch. However, the authors of SVRG conduct their experiment using the last vector rather than the average because it is more "natural". This present paper partially shows that this natural choice also has competitive performance, and therefore confirms the empirical finding of SVRG. (Similar result can also be obtained for the strongly convex case, which we exclude for simplicity.)

**Algorithm 1** $\texttt{SVRG}^{++}(x^\phi, m_0, S, \eta)$

$\widetilde{x}^0 \leftarrow x^\phi$, $x_0^1 \leftarrow x^\phi$
**for** $s \leftarrow 1$ **to** $S$ **do**
  $\widetilde{\mu}_{s-1} \leftarrow \nabla f(\widetilde{x}^{s-1})$
  $m_s \leftarrow 2^s \cdot m_0$
  **for** $t \leftarrow 0$ **to** $m_s - 1$ **do**
    Pick $i$ uniformly at random in $\{1, \cdots, n\}$.
    $\xi \leftarrow \nabla f_i(x_t^s) - \nabla f_i(\widetilde{x}^{s-1}) + \widetilde{\mu}_{s-1}$
    $x_{t+1}^s = \arg\min_{y \in \mathbb{R}^d} \left\{ \frac{1}{2\eta}\|x_t^s - y\|^2 + \Psi(y) + \langle \xi, y \rangle \right\}$
  **end for**
  $\widetilde{x}^s \leftarrow \frac{1}{m_s} \sum_{t=1}^{m_s} x_t^s$
  $x_0^{s+1} \leftarrow x_{m_s}^s$
**end for**
**return** $\widetilde{x}^S$.

**High-Level Techniques.** Our proof is based on a new way to telescope regret inequalities that is specially designed for growing-epoch methods. Unlike the analysis of SVRG, we telescope not only across iterations, see (A.2), but also across epochs, see (A.3). In contrast, the original SVRG has to rely on the strong convexity of $f(\cdot)$ in order to combine different epochs — this is why SVRG cannot directly solve non-strongly convex objectives. Our technique is also very different from known direct methods such as SAG or SAGA: to some extent, these methods can be viewed as having "equivalent" epoch length $n$, because each stochastic gradient in SAG or SAGA is updated once every $n$ iterations on average. As a result, it may be hard to grow their epoch length. Finally, it is the telescoping across all epochs and all iterations that requires the starting vector of an epoch to be the last one from the previous epoch (which is different from SVRG). We shall demonstrate in our experiment section that these modifications on top of SVRG are also useful in practice.

Our full proof of Theorem 4.1 is included in Appendix A.

## 4.1 Additional Improvements

Inspired by $\texttt{SVRG}^{++}$, we also introduce `SVRG_Auto_Epoch`, a variant of $\texttt{SVRG}^{++}$ where epoch length is automatically determined instead of doubled every epoch. Auto epoch is an attractive feature in practice because it enables the algorithm to perform well for different types of objectives.

The criterion we use to determine the termination of epoch $s$ in `SVRG_Auto_Epoch` is based on the quality of the snapshot full gradient $\nabla f(\widetilde{x}^{s-1})$. Intuitively, if epoch length is too long, an algorithm may move too far from the snapshot point, meaning that the gradient estimator $\xi$ may have a large variance. Following this intuition, for every iteration $t$, we record $\texttt{diff}_t = \|\nabla f_i(x_t^s) - \nabla f_i(\widetilde{x}^{s-1})\|_2^2$ because $\mathbb{E}_i[\texttt{diff}_t]$ is a very tight upper bound on the variance of the gradient estimator (see the proof of Lemma A.2). Under this notion, we decide the epoch termination of `SVRG_Auto_Epoch` as follows. Each epoch has a minimum length of $n/4$. From iteration $t = n/4$ onwards, we keep track of the average $\texttt{diff}_t$ in the last $n/4$ iterations, i.e., $\sum_{j=t-n/4+1}^{t} \texttt{diff}_j$. If this quantity is greater than half of the average $\texttt{diff}_j$ recorded from the previous epoch, we terminate the current epoch and start a new one.[8] `SVRG_Auto_Epoch` shows good performance in our experiments, and we leave it as an open question to prove a complexity result for this method.

In addition to auto epoch, $\texttt{SVRG}^{++}$ can also be combined with other enhancements proposed for `SVRG`. For example, [10] saves the time to compute full gradients at snapshot points by making

[8] We always set the first epoch to be of length $n/4$ and the second to be of length $n/2$.

**Algorithm 2** $\texttt{SVRG}(x^{\phi}, m, S, \eta)$ [13]

1: $\widetilde{x}^0 \leftarrow x^{\phi}$, $x_0^1 \leftarrow x^{\phi}$
2: **for** $s \leftarrow 1$ **to** $S$ **do**
3: $\quad \widetilde{\mu}_{s-1} \leftarrow \nabla f(\widetilde{x}^{s-1})$
4: $\quad$ **for** $t \leftarrow 0$ **to** $m-1$ **do**
5: $\quad\quad$ Pick $i$ uniformly at random in $\{1, \cdots, n\}$.
6: $\quad\quad \xi \leftarrow \nabla f_i(x_t^s) - \nabla f_i(\widetilde{x}^{s-1}) + \widetilde{\mu}_{s-1}$
7: $\quad\quad x_{t+1}^s = \arg\min_{y \in \mathbb{R}^d} \big\{ \frac{1}{2\eta} \|x_t^s - y\|^2 + \Psi(y) + \langle \xi, y \rangle \big\}$
8: $\quad$ **end for**
9: $\quad \widetilde{x}^s \leftarrow \frac{1}{m} \sum_{t=1}^{m} x_t^s$
10: $\quad x_0^{s+1} \leftarrow \widetilde{x}^s$
11: **end for**
12: **return** $\widetilde{x}^S$.

them less accurate in the first a few epochs. [14] uses mini-batch gradients per iteration to further decrease the variance. These ideas are orthogonal to our proposed techniques and therefore can be applied to further improve the performance of $\texttt{SVRG}^{++}$.

# 5 SVRG for Sum-of-Non-Convex Objectives I: Small Lower Smoothness

In this section we consider Problem (1.1) when each $f_i(x)$ is not necessarily convex, $L$-upper smooth, and $l$-lower smooth for some $0 \le l \le L$. We assume that $f(\cdot)$ is $\sigma$-strongly convex. For this class of objectives, the best known gradient complexity for stochastic gradient methods is $O\big((n + \frac{L^2}{\sigma^2}) \log \frac{1}{\varepsilon}\big)$ due to SVRG [22].

This gradient complexity is essentially a factor $L/\sigma$ greater than that for the convex case, that is $O\big((n + \frac{L}{\sigma}) \log \frac{1}{\varepsilon}\big)$. Following the intuition discussed in the introduction, we improve it to $O\big((n + \frac{L}{\sigma} + \frac{Ll}{\sigma^2}) \log \frac{1}{\varepsilon}\big)$, a quantity that is asymptotically the same as the convex setting when $l \le O(\sigma)$, and linearly degrades as $l$ increases.

Recall that the original SVRG (Option II) works as follows (see Algorithm 2 for completeness). Given an initial vector $x^{\phi}$, SVRG is divided into $S$ epochs, each of length $m$ for the same $m$ across epochs. Within each epoch, SVRG computes the full gradient $\widetilde{\mu}_{s-1} = \nabla f(\widetilde{x}^{s-1})$ where $\widetilde{x}^{s-1}$ is the average point of the previous epoch. Then, SVRG uses $\widetilde{\mu}_{s-1}$ to define the variance-reduced version of the stochastic gradient $\xi$, see Line 6 of Algorithm 2. The starting vector $x_0^s$ of each epoch is set to be the average vector of the previous epoch.[9]

We state our main result for $\texttt{SVRG}$ in this section as follows:

[9]This choice of the starting vector is different from $\texttt{SVRG}^{++}$, but was the original choice made by SVRG. Similar result can also be obtained using the choice from $\texttt{SVRG}^{++}$.

**Theorem 5.1.** *If each $f_i(x)$ is $L$-upper and $l$-lower smooth in Problem (1.1) for $0 \le l \le L$, $f(x)$ is $\sigma$-strongly convex, $\eta = \min\{\frac{1}{21L}, \frac{\sigma}{63Ll}\}$ and $m \ge \frac{10}{\sigma\eta} = \Omega(\max\{\frac{L}{\sigma}, \frac{Ll}{\sigma^2}\})$, then $\texttt{SVRG}(x^\phi, m, S, \eta)$ satisfies*[a]

$$\mathbb{E}[F(\widetilde{x}^s) - F(x^*)] \le \frac{3}{4}\big(F(\widetilde{x}^{s-1}) - F(x^*)\big) \ . \tag{5.1}$$

*Therefore, by setting $S = \log_{4/3}\big(\frac{F(x^\phi)-F(x^*)}{\varepsilon}\big)$, in a total gradient complexity of*

$$O\Big(\Big(n + \frac{L}{\sigma}\max\big\{1, \frac{l}{\sigma}\big\}\Big)\log\frac{F(x^\phi) - F(x^*)}{\varepsilon}\Big) \ ,$$

*we obtain an output $\widetilde{x}^S$ satisfying $\mathbb{E}[F(\widetilde{x}^s) - F(x^*)] \le \varepsilon$.*

[a] Here we have assumed that the first $s-1$ epochs are fixed and the only randomness comes from epoch $s$.

Our technique for proving this theorem depends on the following new upper bound on the variance. Denoting by $\xi_t^s$ the stochastic gradient $\xi$ at epoch $s$ and iteration $t$, and denoting by $i_t^s$ the random index $i$ chosen at epoch $s$ and iteration $t$, we have

**Lemma 5.2.**

$$\begin{aligned}\mathbb{E}_{i_t^s}\big[\|\xi_t^s - \nabla f(x_t^s)\|^2\big] &\le 4(L+l)\cdot\big(F(x_t^s) - F(x^*) + F(\widetilde{x}^{s-1}) - F(x^*)\big) \\ &\quad + (8l^2 + 4Ll)\big(\|x_t^s - x^*\|^2 + \|\widetilde{x}^{s-1} - x^*\|^2\big) \ .\end{aligned}$$

This is different from Section 4.1 of [22], where the author only provided a weaker upper bound $O(L^2)\cdot\big(\|x_t^s - x^*\|^2 + \|\widetilde{x}^{s-1} - x^*\|^2\big)$. In the event that $l$ is very small, our new upper bound reduces to the variance upper bound in the convex setting, see for instance Eq. (8) of [13].

The full proof of Theorem 5.1 is included in Appendix B.

# 6 SVRG for Sum-of-Non-Convex Objectives II: Large Lower Smoothness

In this section we consider Problem (1.1) when each $f_i(x)$ is not necessarily convex, $L$-upper smooth, and $l$-lower smooth function for some $l \ge L$. We assume $f(\cdot)$ is $\sigma$-strongly convex. For this class of objectives, the best known gradient complexity for stochastic gradient methods is $O\big((n + \frac{l^2}{\sigma^2})\log\frac{1}{\varepsilon}\big)$ due to [22].

This known gradient complexity is essentially a factor $l^2/L^2 \ge 1$ worse than that of the symmetric case (i.e., the case when $l = L$). In this section, we improve this factor to $l/L$ which is quadratically faster than $l^2/L^2$. As we have explained in the introduction, this result improves the convergence for the best known stochastic algorithm for PCA.

We state our main result for `SVRG` in this section as follows.

**Theorem 6.1.** *If each $f_i(x)$ is $L$-upper and $l$-lower smooth in Problem (1.1) for $l \geq L \geq 0$, $f(x)$ is $\sigma$-strongly convex, $\eta = \frac{\sigma}{25Ll}$ and $m \geq \frac{4}{\sigma\eta} = \Omega(\frac{Ll}{\sigma^2})$, then $\mathtt{SVRG}(x^\phi, m, S, \eta)$ satisfies*

$$\mathbb{E}[F(\widetilde{x}^s) - F(x^*)] \leq \frac{3}{4}\big(F(\widetilde{x}^{s-1}) - F(x^*)\big) \ . \tag{6.1}$$

*Therefore, by setting $S = \log_{4/3}\big(\frac{F(x^\phi)-F(x^*)}{\varepsilon}\big)$, in a total gradient complexity of*

$$O\Big(\Big(n + \frac{Ll}{\sigma^2}\Big)\log\frac{F(x^\phi) - F(x^*)}{\varepsilon}\Big) \ ,$$

*we obtain an output $\widetilde{x}^S$ satisfying $\mathbb{E}[F(\widetilde{x}^s) - F(x^*)] \leq \varepsilon$.*

Although Theorem 6.1 (for the large $l$ setting) has the same form as Theorem 5.1 (for the small $l$ setting), its proof is quite different. In order to provide a variance bound without paying the $l^2$ factor as in Lemma 5.2, we negate the objective for analysis purpose only. This is reasonable because $-f_i(\cdot)$ becomes $l$ upper smooth but only $L$ lower smooth for $L \leq l$. By applying the smoothness lemmas for minimizing $-f_i(\cdot)$ (and thus maximizing $f_i(x)$), we obtain a better variance upper bound without paying the factor $l^2$.

Details of the proof is included in Appendix C.

# 7 $\mathtt{SVRG}^{++}_{\mathtt{nc}}$ for Non-Strongly Convex AND Sum-of-Non-Convex Objectives

In this section we show that our improvements for (1) non-strongly convex objectives in Section 4 and for (2) sum-of-non-convex objectives in Section 5 and 6 can be non-trivially put together. That is, we consider the case of Problem (1.1) when each $f_i(x)$ is a not-necessarily convex function but $L$-upper and $l$-lower smooth for $l \geq 0$. We assume that $f$, the average of functions $f_i$, is simply convex but not necessarily strongly convex.

For this class of objectives, if one applies a classical regularization (by adding a dummy $\frac{\sigma}{2}\|x\|^2$ regularizer for $\sigma \stackrel{\text{def}}{=} \frac{\varepsilon}{\|x_0 - x^*\|^2}$) reduction to that of Shalev-Shwartz [22], we can obtain a gradient complexity of essentially $O\big((n + \frac{L^2}{\varepsilon^2})\log\frac{1}{\varepsilon}\big)$. If one applies the same reduction to our new analysis in Section 5 and 6, we can obtain a gradient complexity of essentially $O\big((n + \frac{L}{\varepsilon} + \frac{Ll}{\varepsilon^2})\log\frac{1}{\varepsilon}\big)$. Note that the so-obtained algorithms are indirect and biased.

We propose a direct algorithm $\mathtt{SVRG}^{++}_{\mathtt{nc}}$ for solving this class of objectives with a gradient complexity of $O\big(n\log\frac{1}{\varepsilon} + \frac{L}{\varepsilon} + \frac{Ll}{\varepsilon^2}\big)$.

Our $\mathtt{SVRG}^{++}_{\mathtt{nc}}$ algorithm for this case is analogous to $\mathtt{SVRG}^{++}$ in Section 4. Given an initial vector $x^\phi$, our algorithm is divided into $S$ epochs. The $s$-th epoch consists of $m_s$ stochastic gradient steps, where $m_s$ doubles between every consecutive two epochs. As before, within each epoch we compute the full gradient $\widetilde{\mu}_{s-1} = \nabla f(\widetilde{x}^{s-1})$ where $\widetilde{x}^{s-1}$ is the average point of the previous epoch. We use also $\widetilde{\mu}_{s-1}$ to define the variance-reduced version of the stochastic gradient $\xi$. Unlike $\mathtt{SVRG}^{++}$, for analysis purpose the step length $\eta$ is no longer a constant throughout the iterations. However, it will almost remain a constant.

More precisely, define $T = m_1 + \cdots + m_S \leq 2m_0 \cdot 2^S$ to be the total number of iterations. Then,

**Algorithm 3** $\mathtt{SVRG}^{++}_{\mathtt{nc}}(x^\phi, m_0, S, \eta)$

$\widetilde{x}^0 \leftarrow x^\phi$, $x_0^1 \leftarrow x^\phi$
**for** $s \leftarrow 1$ **to** $S$ **do**
  $\widetilde{\mu}_{s-1} \leftarrow \nabla f(\widetilde{x}^{s-1})$
  $m_s \leftarrow 2^s \cdot m_0$
  $k \leftarrow 0$ and $T \leftarrow m_1 + \cdots + m_S$
  **for** $t \leftarrow 0$ **to** $m_s - 1$ **do**
    Pick $i$ uniformly at random in $\{1, \cdots, n\}$.
    $\xi \leftarrow \nabla f_i(x_t^s) - \nabla f_i(\widetilde{x}^{s-1}) + \widetilde{\mu}_{s-1}$
    $k \leftarrow k + 1$ and $\eta_{t+1}^s \leftarrow \eta \cdot \frac{\sqrt{T}}{\sqrt{2T-k}}$.
    $x_{t+1}^s = \arg\min_{y \in \mathbb{R}^d} \left\{ \frac{1}{2\eta_{t+1}^s} \|x_t^s - y\|^2 + \Psi(y) + \langle \xi, y \rangle \right\}$
  **end for**
  $\widetilde{x}^s \leftarrow \frac{1}{m_s} \sum_{t=0}^{m_s - 1} x_t^s$
  $x_0^{s+1} \leftarrow x_{m_s}^s$
**end for**
**return** $\widetilde{x}^S$.

for some parameter $\eta > 0$ to be chosen later, we define the sequence of step lengths

$$\left(\eta_0^1, \eta_1^1, \ldots \eta_{m^1}^1 (= \eta_0^2), \eta_1^2, \ldots, \eta_{m^2}^2 (= \eta_0^3), \eta_1^3, \cdots \eta_{m^S}^S\right) \stackrel{\text{def}}{=} \left(\frac{\eta\sqrt{T}}{\sqrt{2T}}, \frac{\eta\sqrt{T}}{\sqrt{2T-1}}, \ldots, \frac{\eta\sqrt{T}}{\sqrt{T}}\right) \ .$$

Note that in the above definition, the last step length $\eta_{m_s}^s$ is chosen as the same as the first step length $\eta_0^{s+1}$ of the next epoch. We also have $\frac{\eta}{\sqrt{2}} \leq \eta_t^s \leq \eta$ for all epochs $s$ and all iterations $t \in \{0, 1, \ldots, m_s\}$. Since for every real $k \geq 1$ we have $\sqrt{k} - \sqrt{k-1} \geq \frac{1}{2\sqrt{k}}$, it satisfies that

$$\frac{1}{\eta_{t+1}^s} - \frac{1}{\eta_t^s} \geq \frac{1}{2\eta\sqrt{T}\sqrt{2T}} = \frac{1}{2\sqrt{2}\eta T} \ . \tag{7.1}$$

We state our main convergence result for $\mathtt{SVRG}^{++}_{\mathtt{nc}}$ in this section as follows:

**Theorem 7.1.** *If $f(x)$ is convex, each $f_i(x)$ is $L$-upper and $l$-lower smooth in Problem (1.1) for $l, L \geq 0$, and we are an initial vector $x^\phi$ satisfying $\|x^\phi - x^*\|^2 \leq \Theta$ and $F(x^\phi) - F(x^*) \leq \Delta$ for parameters $\Theta, \Delta \in \mathbb{R}_+$. Then, $\mathtt{SVRG}^{++}_{\mathtt{nc}}(x^\phi, m_0, S, \eta)$ satisfies if $\eta = \min\left\{\frac{1}{13L}, \frac{\varepsilon}{312\sqrt{2}\Theta L l}\right\}$, $m_0 = \frac{\Theta}{\eta\Delta}$, and $S = \log_2(\Delta/\varepsilon)$, we have*

$$\mathbb{E}[F(\widetilde{x}^S) - F(x^*)] \leq O(\varepsilon) \ .$$

*The total gradient complexity is $O(S \cdot n + 2^S \cdot m_0) = O\big(n \log \frac{\Delta}{\varepsilon} + \frac{L\Theta}{\varepsilon} + \frac{Ll\Theta^2}{\varepsilon^2}\big)$.*

The proof can be found in Appendix D.

# 8 Experiments on Empirical Risk Minimization

We confirm our theoretical findings using four real-life datasets: (1) the `Adult` dataset (32,561 examples and 123 features), (2) the `Covtype` dataset (581,012 examples and 54 features), (3) the `Ijcnn1` dataset (49990 examples and 22 features), and (4) the 2nd class of the `MNIST` dataset (60,000 examples and 780 features) [6]. In order to make easy comparisons between different datasets, we scale each data vector down by the *average* Euclidean norm of the whole data set. This step is for comparison only and not necessary in practice.

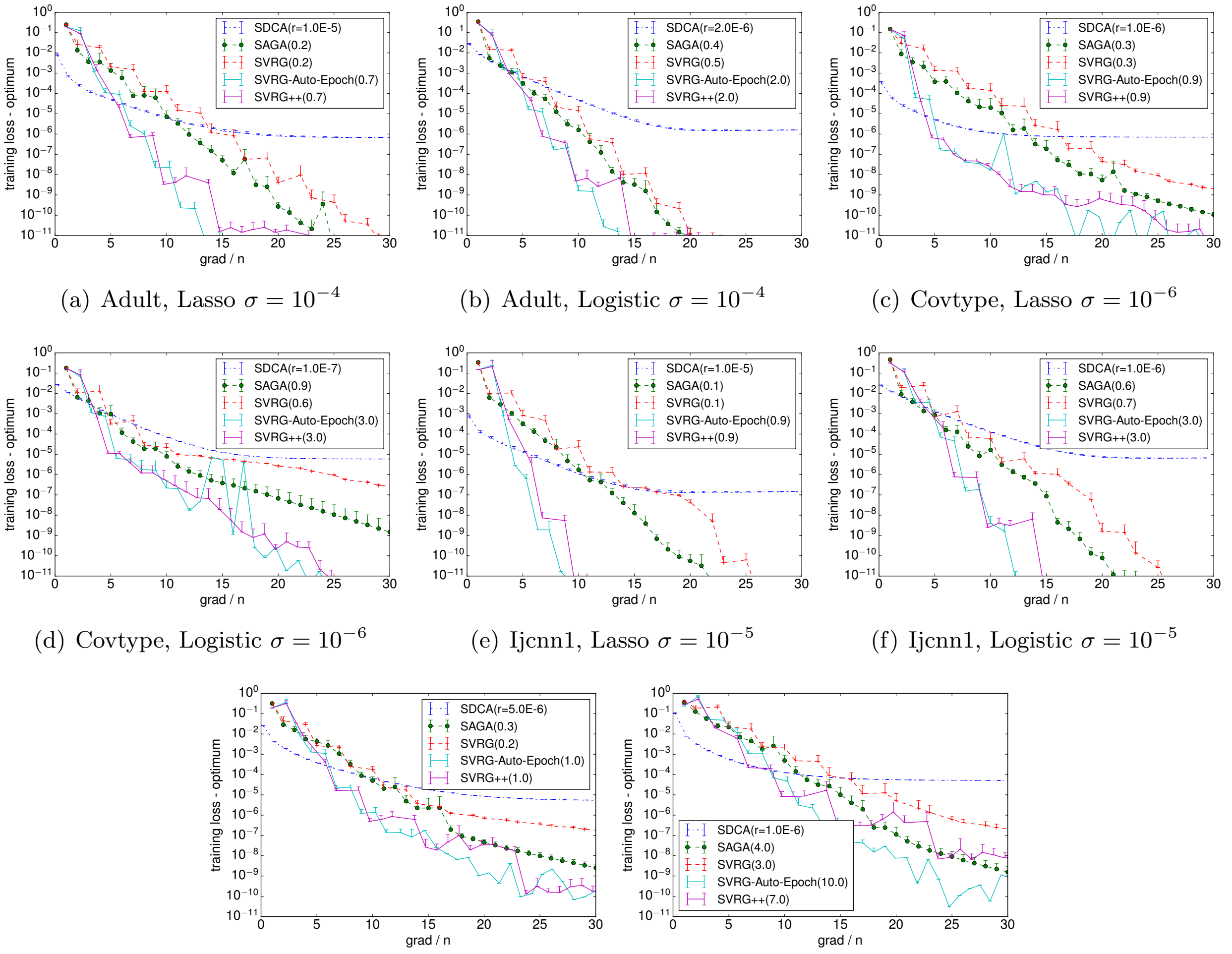

(a) Adult, Lasso $\sigma = 10^{-4}$ (b) Adult, Logistic $\sigma = 10^{-4}$ (c) Covtype, Lasso $\sigma = 10^{-6}$

(d) Covtype, Logistic $\sigma = 10^{-6}$ (e) Ijcnn1, Lasso $\sigma = 10^{-5}$ (f) Ijcnn1, Logistic $\sigma = 10^{-5}$

(g) Mnist, Lasso $\sigma = 10^{-4}$ (h) Mnist, Logistic $\sigma = 10^{-4}$

Figure 1: Selected performance comparisons for lasso and logistic regression using Tuning Type I. Our comprehensive comparisons for other regularizer weights as well as ridge regression can be found in Figure 3, 4, 5, and 6 in the appendix.

We perform 3 classification tasks: *Lasso, ridge regression*, and *$\ell_1$-regularized logistic regression*. As described in the introduction, Lasso and logistic regression do *not* admit strongly convex objectives, while the ridge objective is strongly convex. We consider four different values $\sigma \in \{10^{-3}, 10^{-4}, 10^{-5}, 10^{-6}\}$, where $\sigma$ is either the weight in regularizer $\frac{\sigma}{2}\|x\|_2^2$ for ridge, or that in regularizer $\sigma\|x\|_1^2$ for Lasso and logistic regression.

We have implemented the following algorithms:

- $\texttt{SVRG}^{++}$ with initial epoch length $m_0 = n/4$.
- SVRG_Auto_Epoch as we described in Section 4.1.
- SVRG [13, 29] with (their suggested) epoch length $m = 2n$. (Recall that, in theory, SVRG is not designed for non-strongly convex objectives and $F(\cdot)$ needs to be added by a dummy regularizer for Lasso and logistic regression. However, in our experiments, we observed that this dummy regularizer is not necessary, so have neglected the regularized version of `SVRG` for a clean comparison.)

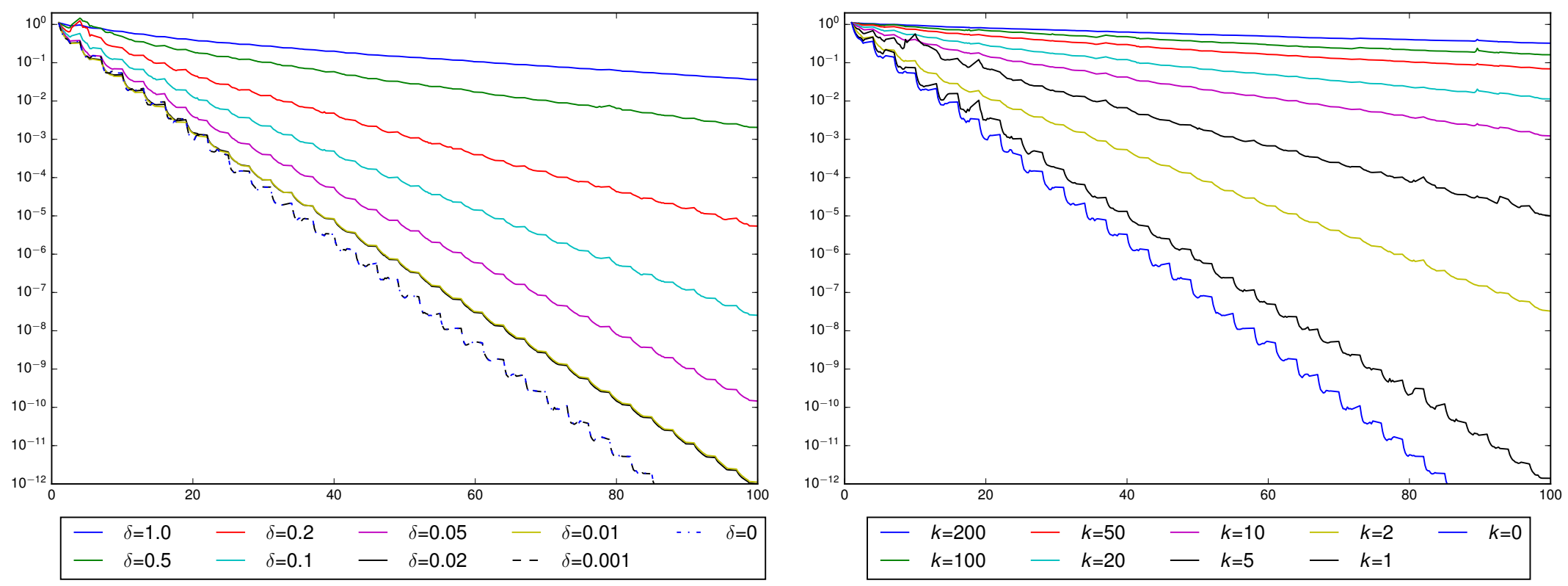

(a) parameterized by $\delta$

(b) parameterized by $k$

Figure 2: Performance analysis on sum-of-non-convex objectives. Note that the curves for $\delta = 0.001, 0.01, 0.02$ have overlapped in (a).

- `SAGA` [5].
- `SDCA` [24, 25] with Option I (steepest descent). Since `SDCA` works only with strongly convex objectives, a dummy regularizer has to be introduced for Lasso and Logistic regression.

For each algorithm above except `SDCA`, we tune the step length carefully from the set $\{a \times 10^{-k} : a \in \{1, 2, \ldots, 9\}, k \in \mathbb{Z}\}$ for each plot. For `SDCA` on Lasso and logistic regression, we also tune the weight of its dummy regularizer from the set $\{10^{-k}, 2 \times 10^{-k}, 5 \times 10^{-k} : k \in \mathbb{Z}\}$. To make our comparison stronger, we adopt an anonymous reviewer's suggestion and consider two types of parameter tuning. In Tuning Type I, we select the best curve based on the training objective performance in the entire 30 passes to the dataset. In Tuning Type II, we select the best parameter only based on method's performance in the first 4 passes to the dataset. Tuning Type II might be more realistic for experimentalists who need to quickly pick the best parameters of the algorithms.

In each plot, we run 10 times the experiments and plot both the mean and the variance. Since our plots are in log scale, we only keep the upper error bar to make the plots easier to read. In other words, the *lower* end of each error bar represents the *mean* of each data point.

**Performance Comparison.** We have picked a representative regularizer weight $\sigma$ for each of the eight analysis tasks (lasso or logistic regression on one of the four datasets), and presented the performance plots using Tuning Type I in Figure 1. For the results on other values of $\sigma$ as well as those for ridge regression, see Figure 3, 4, 5, and 6 in the appendix. We have also included plots using Tuning Type II in Figure 7, 8, 9, and 10 in the appendix.

In all of our plots, the $y$-axis represents the training objective value minus the minimum, and the $x$-axis represents the number of passes to the dataset. Here, following the tradition, one iteration of each algorithm counts as $1/n$ pass of the dataset, and the snapshot full-gradient computation of `SVRG`, `SVRG`$^{++}$, and `SVRG_Auto_Epoch` counts as one additional pass.

In the legend of each plot, we use `SDCA`$(r = r_0)$ to denote that $r_0$ is the weight of the best-tuned dummy regularizer. For every other algorithm, we use `Alg`$(\eta)$ to denote that $\eta$ is the best-tuned step length for algorithm `Alg`.

We make the following observations from this experiment:

- SVRG$^{++}$ and SVRG_Auto_Epoch consistently outperform SVRG in all the plots, indicating that they do improve over SVRG in non-strongly convex settings.
- SVRG$^{++}$ and SVRG_Auto_Epoch outperform SAGA in most cases, and are at least comparable to SAGA in the rest cases. This is not surprising because SAGA is also a direct algorithm for non-strongly convex objectives.
- SVRG$^{++}$ and SVRG_Auto_Epoch significantly outperform indirect methods via dummy regularization (i.e., SDCA) in the non-strongly convex settings. For ridge regression which is strongly convex, SDCA is comparable to other methods (see the figures in the appendix).

# 9 Experiments for Sum-of-Non-Convex Objectives

To verify our theoretical findings in Section 5 and 6, we run SVRG on a sum-of-non-convex objective built from synthetically generated data. We generate $n = 500$ random vectors $a_1, \dots, a_{500} \in \mathbb{R}^d$ from the $d = 200$ dimensional unit cube and then normalize them to have Euclidean norm 1. Define the covariance matrix $A \stackrel{\text{def}}{=} \frac{1}{n}\sum_{i=1}^n a_i a_i^T$, and we consider the minimization problem[10]

$$\min_{x\in\mathbb{R}^d} \Big\{ f(x) \stackrel{\text{def}}{=} \frac{x^T A x}{2} + bx \Big\}$$

for some randomly generated vector $b$.

The matrix $A$ we generated has minimum eigenvalue equal to $7.02 \times 10^{-4}$, and thus $f(x)$ is strongly convex with parameter $7.02 \times 10^{-4}$. Next, we decompose $f(x)$ into an average of $f_i(x)$, each being non-convex with upper and lower smoothness parameters that we can control.

More specifically, given $n$ diagonal matrices $D_1, \cdots, D_n$ satisfying $D_1 + \cdots + D_n = 0$, by setting $f_i(x) \stackrel{\text{def}}{=} \frac{x^T(a_i^T a_i + D_i)x}{2} + bx$, we have $f(x) = \frac{1}{n}\sum_i f_i(x)$. Under this construction, each $f_i$ is non-convex if $D_i$ has negative entries in the diagonals. We now consider two different ways to build $D_1, \dots, D_n$.

**Remark 9.1.** We do not perform real-life PCA experiments for the following reason. Recall Garber and Hazan reduced PCA to minimizing $f(x) = \frac{1}{2}x^T(\mu I - A)x + b^T x$. For all interesting choices of $\mu$, our result in Theorem 6.1 is faster than theirs by the same constant factor $\lambda \in [1/d, 1]$, which is the largest eigenvalue of $A$. Therefore, by varying $\mu$ and comparing the plots, it is impossible to observe anything interesting: in particular, one cannot conclude our theoretical bound is tighter in practice. In contrast, our carefully designed synthetic experiment allows us to control the upper and lower smoothness parameters, and therefore to observe the improvements of our theorems directly.

Our first experiment is parameterized by a given value $\delta \in [0, 1]$. For each $j \in [d]$, we randomly select half of the indices $i \in [n]$ and assign its $j$-th diagonal $(D_i)_{jj}$ to be $\delta$; for the other half of the indices $i$ we assign $(D_i)_{jj}$ to be $-\delta$. In this way, we satisfy $D_1 + \cdots + D_n = 0$ and for each $i \in [n]$, we have $-\delta I \leq \nabla^2 f_i(x) \leq (1+\delta)I$. In other words, each function $f_i(x)$ is $L \approx 1$ upper smooth and exactly $l = \delta$ lower smooth. This corresponds to the $l \leq L$ regime studied by Section 5.

Our second experiment is parameterized by a given value $k \in [1, n]$. For each $j \in [d]$, consider the $j$-th diagonal entry of all the matrices, $(D_1)_{jj}, (D_2)_{jj}, \dots (D_n)_{jj}$. We randomly select one of these entries and set it to be $-k$, and the rest $n-1$ of them to be $\frac{k}{n-1}$. Under this definition, we have $D_1 + \cdots + D_n = 0$ and for each $i \in [n]$, we have $-kI \le \nabla^2 f_i(x) \le (1 + k/(n-1))I$. In other words, each function $f_i(x)$ is approximately $L \approx 1$ upper smooth and $l = k$ lower smooth. This corresponds to the $l \ge L$ regime studied by Section 6.

[10] Since $x^* = A^{-1}b$ this is a linear system problem.

We run SVRG (with the best tuned step length) for both experiments, and plot the performance in Figure 2. We make the following observations from the plots:

- In Figure 2(a), we observe that the performance SVRG is approximately linearly proportional to $lL = O(\delta)$ for large $\delta$, as compared to $L^2 = O(1)$ from prior work. More importantly, SVRG is robust against small non-convexity parameter $l$. Indeed, for $l = \delta \le 0.02$, the convergence of SVRG is as fast as the convex case (i.e., $\delta = 0$ case). This confirms our theoretical finding in Section 5 and particularly confirms the existence of a threshold $O(\sigma)$ where the performance of SVRG only starts to degrade when $l$ is above this threshold.

- In Figure 2(b), we see that the performance of SVRG is approximately linearly proportional to $lL = O(k)$, as compared to $l^2 = O(k^2)$ from prior work. This confirms our finding in Section 6.

# Appendix

## A Convergence Analysis for Section 4

For each outer iteration $s \in [S]$ and inner iteration $t \in \{0, 1, \dots, m_s - 1\}$ of $\mathtt{SVRG}^{++}$, we denote by $i_t^s$ the selected random index $i \in [n]$ and $\xi_t^s$ the stochastic gradient $\xi = \nabla f_{i_t^s}(x_t^s) - \nabla f_{i_t^s}(\widetilde{x}^{s-1}) + \widetilde{\mu}_{s-1}$. Then, using the convexity and smoothness of our objective, as well as the definition of our stochastic gradient step, we obtain the following lemma:

**Lemma A.1.** *For every $u \in \mathbb{R}^d$ and $t \in \{0, 1, \dots, m_s - 1\}$, fixing $x_t^s$ and letting $i = i_t^s$ be the random variable, we have*

$$\mathbb{E}_{i_t^s}\big[F(x_{t+1}^s) - F(u)\big] \le \mathbb{E}_{i_t^s}\Big[\frac{\eta}{2(1-\eta L)}\|\xi_t^s - \nabla f(x_t^s)\|^2 + \frac{\|x_t^s - u\|^2 - \|x_{t+1}^s - u\|^2}{2\eta}\Big] \ .$$

*Proof.* We first upper bound the left hand side:

$$\begin{aligned}
&\mathbb{E}_{i_t^s}\big[F(x_{t+1}^s) - F(u)\big] = \mathbb{E}_{i_t^s}\big[f(x_{t+1}^s) - f(u) + \Psi(x_{t+1}^s) - \Psi(u)\big] \\
&\overset{①}{\le} \mathbb{E}_{i_t^s}\big[f(x_t^s) + \langle \nabla f(x_t^s), x_{t+1}^s - x_t^s\rangle + \tfrac{L}{2}\|x_t^s - x_{t+1}^s\|^2 - f(u) + \Psi(x_{t+1}^s) - \Psi(u)\big] \\
&\overset{②}{\le} \mathbb{E}_{i_t^s}\big[\langle \nabla f(x_t^s), x_t^s - u\rangle + \langle \nabla f(x_t^s), x_{t+1}^s - x_t^s\rangle + \tfrac{L}{2}\|x_t^s - x_{t+1}^s\|^2 + \Psi(x_{t+1}^s) - \Psi(u)\big] \\
&\overset{③}{=} \mathbb{E}_{i_t^s}\big[\langle \xi_t^s, x_t^s - u\rangle + \langle \nabla f(x_t^s), x_{t+1}^s - x_t^s\rangle + \tfrac{L}{2}\|x_t^s - x_{t+1}^s\|^2 + \Psi(x_{t+1}^s) - \Psi(u)\big] \ . \qquad \text{(A.1)}
\end{aligned}$$

Above, inequalities ① and ② are respectively due to the smoothness and convexity of $f(\cdot)$, and ③ is because $\mathbb{E}_{i_t^s}[\xi_t^s] = \nabla f(x_t^s)$. Next, using the definition of $x_{t+1}^s$ we have

$$\begin{aligned}
\langle \xi_t^s, x_t^s - u\rangle + \Psi(x_{t+1}^s) - \Psi(u) &= \langle \xi_t^s, x_t^s - x_{t+1}^s\rangle + \langle \xi_t^s, x_{t+1}^s - u\rangle + \Psi(x_{t+1}^s) - \Psi(u) \\
&\overset{④}{\le} \langle \xi_t^s, x_t^s - x_{t+1}^s\rangle + \langle -\frac{1}{\eta}(x_{t+1}^s - x_t^s), x_{t+1}^s - u\rangle \\
&\overset{⑤}{=} \langle \xi_t^s, x_t^s - x_{t+1}^s\rangle + \frac{\|x_t^s - u\|^2}{2\eta} - \frac{\|x_{t+1}^s - u\|^2}{2\eta} - \frac{\|x_{t+1}^s - x_t^s\|^2}{2\eta} \ .
\end{aligned}$$

Above, inequality ④ holds for the following reason. Recall that the minimality of $x^s_{t+1} = \arg\min_{y\in\mathbb{R}^d}\{\frac{1}{2\eta}\|y - x^s_t\|^2 + \Psi(y) + \langle \xi^s_t, y\rangle\}$ implies the existence of some subgradient $g \in \partial\Psi(x^s_{t+1})$ which satisfies $\frac{1}{\eta}(x^s_{t+1} - x^s_t) + \xi^s_t + g = 0$. Combining this with $\Psi(u) - \Psi(x^s_{t+1}) \geq \langle g, u - x^s_{t+1}\rangle$, which is due to the convexity of $\Psi(\cdot)$, we immediately have $\Psi(u) - \Psi(x^s_{t+1}) + \langle \frac{1}{\eta}(x^s_{t+1} - x^s_t) + \xi^s_t, u - x^s_{t+1}\rangle \geq \langle \frac{1}{\eta}(x^s_{t+1} - x^s_t) + \xi^s_t + g, u - x^s_{t+1}\rangle = 0$. This gives inequality ④. In addition, ⑤ can be verified by expanding the Euclidean norms.

Combining the above two inequalities, we have

$$
\begin{aligned}
&\mathbb{E}_{i^s_t}\big[F(x^s_{t+1}) - F(u)\big] \\
&\leq \mathbb{E}_{i^s_t}\Big[\langle \xi^s_t - \nabla f(x^s_t), x^s_t - x^s_{t+1}\rangle - \frac{1-\eta L}{2\eta}\|x^s_t - x^s_{t+1}\|^2 + \frac{\|x^s_t - u\|^2 - \|x^s_{t+1} - u\|^2}{2\eta}\Big] \\
&\overset{⑥}{\leq} \mathbb{E}_{i^s_t}\Big[\frac{\eta}{2(1-\eta L)}\|\xi^s_t - \nabla f(x^s_t)\|^2 + \frac{\|x^s_t - u\|^2 - \|x^s_{t+1} - u\|^2}{2\eta}\Big] \ .
\end{aligned}
$$

Above, ⑥ is by Young's inequality. □

The next lemma is classical and analogous to most of the variance reduction literatures (cf. [5, 13, 29]). We include it here for the sake of completeness.

**Lemma A.2.** $\mathbb{E}_{i^s_t}\big[\|\xi^s_t - \nabla f(x^s_t)\|^2\big] \leq 4L \cdot \big(F(x^s_t) - F(x^*) + F(\widetilde{x}^{s-1}) - F(x^*)\big)$.

*Proof.* The proof of this lemma is classical and is analogous to most of the variance reduction literatures (cf. [5, 13, 29]). Indeed,

$$
\begin{aligned}
\mathbb{E}_{i^s_t}\big[\|\xi^s_t - \nabla f(x^s_t)\|^2\big] &= \mathbb{E}_{i^s_t}\Big[\big\|\big(\nabla f_{i^s_t}(x^s_t) - \nabla f_{i^s_t}(\widetilde{x}^{s-1})\big) - \big(\nabla f(x^s_t) - \nabla f(\widetilde{x}^{s-1})\big)\big\|^2\Big] \\
&\overset{①}{\leq} \mathbb{E}_{i^s_t}\Big[\big\|\nabla f_{i^s_t}(x^s_t) - \nabla f_{i^s_t}(\widetilde{x}^{s-1})\big\|^2\Big] \\
&= \mathbb{E}_{i^s_t}\Big[\big\|\big(\nabla f_{i^s_t}(x^s_t) - \nabla f_{i^s_t}(x^*)\big) - \big(\nabla f_{i^s_t}(\widetilde{x}^{s-1}) - \nabla f_{i^s_t}(x^*)\big)\big\|^2\Big] \\
&\overset{②}{\leq} 2 \cdot \mathbb{E}_{i^s_t}\Big[\big\|\nabla f_{i^s_t}(x^s_t) - \nabla f_{i^s_t}(x^*)\big\|^2 + \big\|\nabla f_{i^s_t}(\widetilde{x}^{s-1}) - \nabla f_{i^s_t}(x^*)\big\|^2\Big] \ .
\end{aligned}
$$

Above, ① is because for any random vector $\zeta \in \mathbb{R}^d$, it holds that $\mathbb{E}\|\zeta - \mathbb{E}\zeta\|^2 = \mathbb{E}\|\zeta\|^2 - \|\mathbb{E}\zeta\|^2$, and ② is because for any two vectors $a, b \in \mathbb{R}^d$, it holds that $\|a - b\|^2 \leq 2\|a\|^2 + 2\|b\|^2$.

Next, the classical smoothness assumption on a function $f_i$ yields (see for instance Theorem 2.1.5 in the textbook [19]) $\|\nabla f_i(x) - \nabla f_i(x^*)\|^2 \leq 2L\big[f_i(x) - f_i(x^*) - \langle \nabla f_i(x^*), x - x^*\rangle\big)$. Plugging this into the above inequality, we have

$$
\begin{aligned}
&\mathbb{E}_{i^s_t}\big[\|\xi^s_t - \nabla f(x^s_t)\|^2\big] \\
&\leq 4L \cdot \mathbb{E}_{i^s_t}\big[f_{i^s_t}(x^s_t) - f_{i^s_t}(x^*) - \langle \nabla f_{i^s_t}(x^*), x^s_t - x^*\rangle + f_{i^s_t}(\widetilde{x}^{s-1}) - f_{i^s_t}(x^*) - \langle \nabla f_{i^s_t}(x^*), \widetilde{x}^{s-1} - x^*\rangle\big] \\
&= 4L \cdot \big(f(x^s_t) - f(x^*) - \langle \nabla f(x^*), x^s_t - x^*\rangle + f(\widetilde{x}^{s-1}) - f(x^*) - \langle \nabla f(x^*), \widetilde{x}^{s-1} - x^*\rangle\big) \\
&= 4L \cdot \big(f(x^s_t) - f(x^*) + \langle g^*, x^s_t - x^*\rangle + f(\widetilde{x}^{s-1}) - f(x^*) + \langle g^*, \widetilde{x}^{s-1} - x^*\rangle\big) \\
&\leq 4L \cdot \big(f(x^s_t) - f(x^*) + \Psi(x^s_t) - \Psi(x^*) + f(\widetilde{x}^{s-1}) - f(x^*) + \Psi(\widetilde{x}^{s-1}) - \Psi(x^*)\big) \\
&= 4L \cdot \big(F(x^s_t) - F(x^*) + F(\widetilde{x}^{s-1}) - F(x^*)\big) \ .
\end{aligned}
$$

Above, $g^* \in \partial\Psi(x^*)$ is the subgradient of $\Psi$ at $x^*$ that satisfies $\nabla f(x^*) + g^* = 0$. □

We are now ready to prove the main theorem for the convergence of $\texttt{SVRG}^{++}$:

*Proof of Theorem 4.1.* Combining Lemma A.1 with $u = x^*$ and Lemma A.2, we have

$$\mathbb{E}_{i_t^s}\big[F(x_{t+1}^s) - F(x^*)\big] \le \frac{2\eta L}{(1-\eta L)}\big(F(x_t^s) - F(x^*) + F(\widetilde{x}^{s-1}) - F(x^*)\big) + \frac{\|x_t^s - x^*\|^2 - \mathbb{E}_{i_t^s}\|x_{t+1}^s - x^*\|^2}{2\eta} \enspace .$$

Choosing $\eta = 1/(7L)$ in the above inequality, summing it up over $t = 0, 1, \ldots, m_s - 1$, and dividing both sides by $m_s$, we arrive at

$$\mathbb{E}\Big[\sum_{t=0}^{m_s-1} \frac{F(x_{t+1}^s)}{m_s} - F(x^*)\Big] \le \mathbb{E}\Big[\frac{1}{3}\Big(\sum_{t=0}^{m_s-1} \frac{F(x_t^s)}{m_s} - F(x^*) + F(\widetilde{x}^{s-1}) - F(x^*)\Big) + \frac{\|x_0^s - x^*\|^2 - \|x^* - x_{m_s}^s\|^2}{2\eta \cdot m_s}\Big] \enspace . \tag{A.2}$$

After rearranging, this yields

$$\begin{aligned} 2\mathbb{E}\Big[\sum_{t=0}^{m_s-1} \frac{F(x_{t+1}^s)}{m_s} - F(x^*)\Big] \le \mathbb{E}\Big[&\frac{(F(x_0^s) - F(x^*)) - (F(x_{m_s}^s) - F(x^*))}{m_s} + F(\widetilde{x}^{s-1}) - F(x^*) \\ &+ \frac{\|x_0^s - x^*\|^2 - \|x^* - x_{m_s}^s\|^2}{2\eta/3 \cdot m_s}\Big] \enspace . \end{aligned}$$

Next, using the fact that $F(\widetilde{x}^s) \le \sum_{t=0}^{m_s-1} \frac{F(x_{t+1}^s)}{m_s}$ due to the convexity of $F$ and the definition $\widetilde{x}^s = \sum_{t=0}^{m_s-1} \frac{x_{t+1}^s}{m_s}$, as well as the choice $x_{m_s}^s = x_0^{s+1}$, we rewrite the above inequality as

$$\begin{aligned} 2\mathbb{E}\big[F(\widetilde{x}^s) - F(x^*)\big] \le \mathbb{E}\Big[&\frac{(F(x_0^s) - F(x^*)) - (F(x_0^{s+1}) - F(x^*))}{m_s} + F(\widetilde{x}^{s-1}) - F(x^*)\big) \\ &+ \frac{\|x_0^s - x^*\|^2 - \|x^* - x_0^{s+1}\|^2}{2\eta/3 \cdot m_s}\Big] \enspace . \end{aligned} \tag{A.3}$$

After rearranging and using the fact $m_s = 2m_{s-1}$, we conclude that

$$\begin{aligned} &2\mathbb{E}\big[F(\widetilde{x}^s) - F(x^*) + \frac{\|x^* - x_0^{s+1}\|^2}{4\eta/3 \cdot m_s} + \frac{F(x_0^{s+1}) - F(x^*)}{2m_s}\big] \\ &\le \mathbb{E}\Big[F(\widetilde{x}^{s-1}) - F(x^*) + \frac{\|x_0^s - x^*\|^2}{4\eta/3 \cdot m_{s-1}} + \frac{F(x_0^s) - F(x^*)}{2m_{s-1}}\Big] \enspace . \end{aligned}$$

In sum, after telescoping for $s = 1, 2, \ldots, S$, we have[11]

$$\begin{aligned} \mathbb{E}[F(\widetilde{x}^S) - F(x^*)] &\le 2^{-S} \cdot \Big(F(\widetilde{x}^0) - F(x^*) + \frac{\|x^* - x_0^1\|^2}{4\eta/3 \cdot m_0} + \frac{F(x_0^1) - F(x^*)}{2m_0}\Big) \\ &\le \frac{F(x^\phi) - F(x^*)}{2^{S-1}} + \frac{\|x^\phi - x^*\|^2}{2^S \cdot \frac{4\eta m_0}{3}} \enspace . \end{aligned}$$

This finishes the proof of (4.1) due to the choice of $\eta = 1/(7L)$. Finally, $\texttt{SVRG}^{++}$ computes $S$ times the full gradient $\nabla f(\cdot)$, and $\sum_{s=1}^S m_s = O(2^S m_0)$ times the gradient $\nabla f_i(\cdot)$. This gives a total gradient complexity $O(S \cdot n + 2^S \cdot m_0)$. ☐

[11] We can perform telescoping because we set our starting vector $x_0^{s+1}$ of each epoch to equal the ending vector $x_{m_s}^s$ of the previous epoch. This is different from SVRG, which chooses the average of the previous epoch as the starting vector. This difference is also beneficial in practice (see Section 8).

## B Convergence Analysis for Section 5

As in Section 4, for each outer iteration $s \in [S]$ and inner iteration $t \in \{0, 1, \dots, m-1\}$ of SVRG, we denote by $i_t^s$ the selected random index $i \in [n]$ and $\xi_t^s$ the stochastic gradient $\xi = \nabla f_{i_t^s}(x_t^s) - \nabla f_{i_t^s}(\widetilde{x}^{s-1}) + \widetilde{\mu}_{s-1}$. Then, the following lemma is a counterpart of Lemma A.1 where the only difference is the use of the strong convexity parameter $\sigma$:

**Lemma B.1.** *For every $u \in \mathbb{R}^d$ and $t \in \{0, 1, \dots, m-1\}$, fixing $x_t^s$ and letting $i = i_t^s$ be the random variable, we have*

$$\mathbb{E}_{i_t^s}\big[F(x_{t+1}^s) - F(u)\big] \leq \mathbb{E}_{i_t^s}\Big[\frac{\eta}{2(1-\eta L)}\|\xi_t^s - \nabla f(x_t^s)\|^2 + \frac{(1-\sigma\eta)\|x_t^s - u\|^2 - \|x_{t+1}^s - u\|^2}{2\eta}\Big] \ .$$

*Proof.* We first upper bound the left hand side using the strong convexity and smoothness of $f(\cdot)$:

$$\begin{aligned}
&\mathbb{E}_{i_t^s}\big[F(x_{t+1}^s) - F(u)\big] \\
&= \mathbb{E}_{i_t^s}\big[f(x_{t+1}^s) - f(u) + \Psi(x_{t+1}^s) - \Psi(u)\big] \\
&\leq \mathbb{E}_{i_t^s}\big[f(x_t^s) + \langle \nabla f(x_t^s), x_{t+1}^s - x_t^s\rangle + \frac{L}{2}\|x_t^s - x_{t+1}^s\|^2 - f(u) + \Psi(x_{t+1}^s) - \Psi(u)\big] \\
&\leq \mathbb{E}_{i_t^s}\big[\langle \nabla f(x_t^s), x_t^s - u\rangle - \boxed{\frac{\sigma}{2}\|x_t^s - u\|^2} + \langle \nabla f(x_t^s), x_{t+1}^s - x_t^s\rangle + \frac{L}{2}\|x_t^s - x_{t+1}^s\|^2 + \Psi(x_{t+1}^s) - \Psi(u)\big] \\
&= \mathbb{E}_{i_t^s}\big[\langle \xi_t^s, x_t^s - u\rangle - \boxed{\frac{\sigma}{2}\|x_t^s - u\|^2} + \langle \nabla f(x_t^s), x_{t+1}^s - x_t^s\rangle + \frac{L}{2}\|x_t^s - x_{t+1}^s\|^2 + \Psi(x_{t+1}^s) - \Psi(u)\big]
\end{aligned} \tag{B.1}$$

Above, the term $\frac{\sigma}{2}\|x_t^s - u\|^2$ is due to the $\sigma$-strong convexity of $f(\cdot)$, and this is the only difference between the inequalities (B.1) and (A.1). Therefore, Lemma B.1 can be proven using exactly the identical rest of the proof of Lemma A.1. □

We next state and prove a counterpart of Lemma A.2.

**Lemma 5.2.**

$$\begin{aligned}
\mathbb{E}_{i_t^s}\big[\|\xi_t^s - \nabla f(x_t^s)\|^2\big] &\leq 4(L+l)\cdot\big(F(x_t^s) - F(x^*) + F(\widetilde{x}^{s-1}) - F(x^*)\big) \\
&\quad + (8l^2 + 4Ll)\big(\|x_t^s - x^*\|^2 + \|\widetilde{x}^{s-1} - x^*\|^2\big) \ .
\end{aligned}$$

Before we prove this lemma let us make a few remarks. First, if $l = 0$ then Lemma 5.2 is identical to Lemma A.2. In general, the second term in the above upper bound has a factor $8l^2 + 4Ll$ in the front which increases as $l$ increases. We can also compare Lemma 5.2 to that obtained by Shalev-Shwartz for sum-of-non-convex objectives: he showed $\|\xi_t^s - \nabla f(x_t^s)\|^2 \leq O(L^2)\cdot\big(\|x_t^s - x^*\|^2 + \|\widetilde{x}^{s-1} - x^*\|^2\big)$ in [22] which is suboptimal to ours and exactly why the $L^2$ factor shows up in his final gradient complexity.

*Proof of Lemma 5.2.* The first step of the proof of this lemma is analogous to most of the variance reduction literatures (cf. [5, 13, 29]):

$$\begin{aligned}
\mathbb{E}_{i_t^s}\big[\|\xi_t^s - \nabla f(x_t^s)\|^2\big] &= \mathbb{E}_{i_t^s}\Big[\big\|\big(\nabla f_{i_t^s}(x_t^s) - \nabla f_{i_t^s}(\widetilde{x}^{s-1})\big) - \big(\nabla f(x_t^s) - \nabla f(\widetilde{x}^{s-1})\big)\big\|^2\Big] \\
&\overset{①}{\leq} \mathbb{E}_{i_t^s}\Big[\big\|\nabla f_{i_t^s}(x_t^s) - \nabla f_{i_t^s}(\widetilde{x}^{s-1})\big\|^2\Big] \\
&= \mathbb{E}_{i_t^s}\Big[\big\|\big(\nabla f_{i_t^s}(x_t^s) - \nabla f_{i_t^s}(x^*)\big) - \big(\nabla f_{i_t^s}(\widetilde{x}^{s-1}) - \nabla f_{i_t^s}(x^*)\big)\big\|^2\Big] \\
&\overset{②}{\leq} 2\cdot\mathbb{E}_{i_t^s}\Big[\big\|\nabla f_{i_t^s}(x_t^s) - \nabla f_{i_t^s}(x^*)\big\|^2 + \big\|\nabla f_{i_t^s}(\widetilde{x}^{s-1}) - \nabla f_{i_t^s}(x^*)\big\|^2\Big] \ .
\end{aligned} \tag{B.2}$$

Above, ① is because for any random vector $\zeta \in \mathbb{R}^d$, it holds that $\mathbb{E}\|\zeta - \mathbb{E}\zeta\|^2 = \mathbb{E}\|\zeta\|^2 - \|\mathbb{E}\zeta\|^2$, and ② is because for any two vectors $a, b \in \mathbb{R}^d$, it holds that $\|a-b\|^2 \le 2\|a\|^2 + 2\|b\|^2$.

For analysis-purpose only, we define $\phi_i(y) \stackrel{\text{def}}{=} f_i(y) - \langle \nabla f_i(x^*), y\rangle + \frac{l}{2}\|y - x^*\|^2$ for each $i \in [n]$. It is clear that $\phi_i(y)$ is a convex, $(L+l)$-smooth function that has a minimizer $y = x^*$ (which can be seen by taking the derivative). For this reason, we claim that

$$\phi_i(x^*) \le \phi_i(y) - \frac{1}{L+l}\|\nabla \phi_i(y)\|^2 \ , \tag{B.3}$$

for each $y$, and this inequality is classical for smooth functions (see for instance Theorem 2.1.5 in the textbook [19]). By expanding out the definition of $\phi_i(\cdot)$ in (B.3), we immediately have

$$\begin{aligned} f_i(x^*) - \langle \nabla f_i(x^*), x^* \rangle \le f_i(y) - \langle \nabla f_i(x^*), y\rangle + \frac{l}{2}\|y - x^*\|^2 \\ - \frac{1}{2(L+l)}\|\nabla f_i(y) - \nabla f_i(x^*) + l(y - x^*)\|^2 \end{aligned}$$

which then implies

$$\begin{aligned} \|\nabla f_i(y) - \nabla f_i(x^*)\|^2 &\le 2\|\nabla f_i(y) - \nabla f_i(x^*) + l(y-x^*)\|^2 + 2\|l(y-x^*)\|^2 \\ &\le 2(L+l)(f_i(y) - f_i(x^*) - \langle \nabla f_i(x^*), y - x^*\rangle) + (4l^2 + 2Ll)\|y - x^*\|^2 \ . \end{aligned} \tag{B.4}$$

Now, by choosing $y = x_t^s$ and $i = i_t^s$ in (B.4), we have

$$\begin{aligned} &\mathbb{E}_{i_t^s}\big[\big\|\nabla f_{i_t^s}(x_t^s) - \nabla f_{i_t^s}(x^*)\big\|^2\big] \\ &\le \mathbb{E}_{i_t^s}\big[2(L+l)(f_{i_t^s}(x_t^s) - f_{i_t^s}(x^*) - \langle \nabla f_{i_t^s}(x^*), x_t^s - x^*\rangle)\big] + (4l^2+2Ll)\|x_t^s - x^*\|^2 \\ &= 2(L+l)\big(f(x_t^s) - f(x^*) + \langle g^*, x_t^s - x^*\rangle\big) + (4l^2+2Ll)\|x_t^s - x^*\|^2 \\ &\le 2(L+l)\big(f(x_t^s) - f(x^*) + \psi(x_t^s) - \psi(x^*)\big) + (4l^2+2Ll)\|x_t^s - x^*\|^2 \\ &= 2(L+l)\big(F(x_t^s) - F(x^*)\big) + (4l^2+2Ll)\|x_t^s - x^*\|^2 \ . \end{aligned} \tag{B.5}$$

Above, $g^* \in \partial\Psi(x^*)$ is the subgradient of $\Psi$ at $x^*$ that satisfies $\nabla f(x^*) + g^* = 0$.

Similarly, by choosing $y = \widetilde{x}^{s-1}$ and $i = i_t^s$ in (B.4), we have

$$\mathbb{E}_{i_t^s}\big[\big\|\nabla f_{i_t^s}(\widetilde{x}^{s-1}) - \nabla f_{i_t^s}(x^*)\big\|^2\big] \le 2(L+l)\big(F(\widetilde{x}^{s-1}) - F(x^*)\big) + (4l^2+2Ll)\|\widetilde{x}^{s-1} - x^*\|^2 \ . \tag{B.6}$$

Finally, putting together (B.2), (B.5) and (B.6) we finish the proof of the desired lemma. □

Finally, we are ready to prove our main theorem of this section:

*Proof of Theorem 5.1.* Combining Lemma B.1 with $u = x^*$, Lemma 5.2, as well as the assumption that $l \le L$, we have

$$\begin{aligned} \mathbb{E}_{i_t^s}\big[F(x_{t+1}^s) - F(x^*)\big] \le \frac{4\eta L}{(1-\eta L)}\big(F(x_t^s) - F(x^*) + F(\widetilde{x}^{s-1}) - F(x^*) + \frac{3l}{2}\|x_t^s - x^*\|^2 + \frac{3l}{2}\|\widetilde{x}^{s-1} - x^*\|^2\big) \\ + \frac{(1-\sigma\eta)\|x_t^s - x^*\|^2 - \mathbb{E}_{i_t^s}\|x_{t+1}^s - x^*\|^2}{2\eta} \ . \end{aligned}$$

Choosing $\eta = \min\{\frac{1}{21L}, \frac{\sigma}{63Ll}\}$ in the above inequality, we conclude that

$$\mathbb{E}_{i_t^s}\big[F(x_{t+1}^s) - F(x^*)\big] \le \frac{1}{5}\big(F(x_t^s) - F(x^*) + F(\widetilde{x}^{s-1}) - F(x^*)\big) + \frac{\sigma}{10}\|\widetilde{x}^{s-1} - x^*\|^2 + \frac{\|x_t^s - x^*\|^2 - \mathbb{E}_{i_t^s}\|x_{t+1}^s - x^*\|^2}{2\eta} \ .$$

Summing it up over $t = 0, 1, \dots, m-1$, and dividing both sides by $m$, we arrive at

$$\mathbb{E}\Big[\sum_{t=0}^{m-1}\frac{F(x_{t+1}^s)}{m} - F(x^*)\Big] \le \mathbb{E}\Big[\frac{1}{5}\Big(\sum_{t=0}^{m-1}\frac{F(x_t^s)}{m} - F(x^*) + F(\widetilde{x}^{s-1}) - F(x^*)\Big) + \frac{\|x_0^s - x^*\|^2}{2\eta\cdot m} + \frac{\sigma}{10}\|\widetilde{x}^{s-1} - x^*\|^2\Big] \ .$$

After rearranging we have

$$\begin{aligned}4\mathbb{E}\Big[\sum_{t=0}^{m-1}\frac{F(x_{t+1}^s)}{m} - F(x^*)\Big] &\le \mathbb{E}\Big[\frac{(F(x_0^s) - F(x^*)) - (F(x_m^s) - F(x^*))}{m} + F(\widetilde{x}^{s-1}) - F(x^*) \\ &\qquad + \frac{\|x_0^s - x^*\|^2}{2\eta/5\cdot m} + \frac{\sigma}{2}\|\widetilde{x}^{s-1} - x^*\|^2\Big] \\ &\le \big(1 + \frac{1}{m}\big)(F(\widetilde{x}^{s-1}) - F(x^*)) + \big(\frac{5}{\sigma\eta m} + 1\big)\big(F(\widetilde{x}^{s-1}) - F(x^*)\big) \ .\end{aligned}$$

Above, the last inequality uses the fact that $x^*$ is a minimizer of $F(\cdot)$ as well as our choice $x_0^s = \widetilde{x}^{s-1}$. Using the convexity of $F(\cdot)$ we have $F(\widetilde{x}^s) \le \frac{1}{m}\sum_{t=1}^m F(x_t^s)$ and therefore the above inequality gives

$$\mathbb{E}[F(\widetilde{x}^s) - F(x^*)] \le \frac{2 + \frac{1}{m} + \frac{5}{\sigma\eta m}}{4}\big(F(\widetilde{x}^{s-1}) - F(x^*)\big) \ . \qquad \square$$

# C Convergence Analysis for Section 6

This section is devoted to proving Theorem 6.1. We use the same notation as in Section 5 and Lemma B.1 remains true here. We replace Lemma 5.2 with the following:

**Lemma C.1.**

$$\mathbb{E}_{i_t^s}\big[\|\xi_t^s - \nabla f(x_t^s)\|^2\big] \le (8L^2 + 4Ll)\big(\|x_t^s - x^*\|^2 + \|\widetilde{x}^{s-1} - x^*\|^2\big) \ .$$

*Proof.* We begin the proof by first recalling (B.2) from the proof of Lemma 5.2.

$$\mathbb{E}_{i_t^s}\big[\|\xi_t^s - \nabla f(x_t^s)\|^2\big] \le 2\cdot\mathbb{E}_{i_t^s}\Big[\big\|\nabla f_{i_t^s}(x_t^s) - \nabla f_{i_t^s}(x^*)\big\|^2 + \big\|\nabla f_{i_t^s}(\widetilde{x}^{s-1}) - \nabla f_{i_t^s}(x^*)\big\|^2\Big] \ . \qquad \text{(B.2)}$$

This time, we define $\phi_i(y) \stackrel{\text{def}}{=} -f_i(y) + \langle \nabla f_i(x^*), y\rangle + \frac{L}{2}\|y - x^*\|^2$ for each $i \in [n]$. It is clear that $\phi_i(y)$ is a convex, $(L+l)$-smooth function that has a minimizer $y = x^*$ (which can be seen by taking the derivative). For this reason, we claim that

$$\phi_i(x^*) \le \phi_i(y) - \frac{1}{L+l}\|\nabla\phi_i(y)\|^2 \ , \qquad \text{(C.1)}$$

for each $y$, and this inequality is classical for smooth functions (see for instance Theorem 2.1.5 in the textbook [19]). By expanding out the definition of $\phi_i(\cdot)$ in (C.1), we immediately have

$$\begin{aligned}-f_i(x^*) + \langle \nabla f_i(x^*), x^*\rangle \le -f_i(y) + \langle \nabla f_i(x^*), y\rangle + \frac{L}{2}\|y - x^*\|^2 \\ - \frac{1}{2(L+l)}\|\nabla f_i(y) - \nabla f_i(x^*) - L(y - x^*)\|^2\end{aligned}$$

which then implies that

$$\begin{aligned}\|\nabla f_i(y) - \nabla f_i(x^*)\|^2 &\leq 2\|\nabla f_i(y) - \nabla f_i(x^*) - L(y - x^*)\|^2 + 2\|l(y - x^*)\|^2 \\ &\leq 2(L + l)(f_i(x^*) - f_i(y) + \langle \nabla f_i(x^*), y - x^* \rangle) + (4L^2 + 2Ll)\|y - x^*\|^2 \ . \end{aligned} \tag{C.2}$$

Now by choosing $y = x_t^s$ and $i = i_t^s$ in (C.2), we have

$$\begin{aligned} &\mathbb{E}_{i_t^s}\big[\big\|\nabla f_{i_t^s}(x_t^s) - \nabla f_{i_t^s}(x^*)\big\|^2\big] \\ &\leq \mathbb{E}_{i_t^s}\big[2(L + l)(f_{i_t^s}(x^*) - f_{i_t^s}(x_t^s) + \langle \nabla f_{i_t^s}(x^*), x_t^s - x^* \rangle)\big] + (4L^2 + 2Ll)\|x_t^s - x^*\|^2 \\ &= 2(L + l)\big(f(x^*) - f(x_t^s) + \langle \nabla f(x^*), x_t^s - x^* \rangle\big) + (4L^2 + 2Ll)\|x_t^s - x^*\|^2 \\ &\leq (4L^2 + 2Ll)\|x_t^s - x^*\|^2 \ . \end{aligned} \tag{C.3}$$

Above, the second inequality uses the convexity of $f(\cdot)$. Similarly, by choosing $y = \widetilde{x}^{s-1}$ and $i = i_t^s$ in (C.2), we have

$$\mathbb{E}_{i_t^s}\big[\big\|\nabla f_{i_t^s}(\widetilde{x}^{s-1}) - \nabla f_{i_t^s}(x^*)\big\|^2\big] \leq (4L^2 + 2Ll)\|\widetilde{x}^{s-1} - x^*\|^2 \ . \tag{C.4}$$

Finally, putting together (B.2), (C.3) and (C.4) we finish the proof of the desired lemma. □

Finally, we are ready to prove our main theorem of this section:

*Proof of Theorem 6.1.* Combining Lemma B.1 with $u = x^*$, Lemma C.1, as well as the assumption that $L \leq l$, we have

$$\begin{aligned}\mathbb{E}_{i_t^s}\big[F(x_{t+1}^s) - F(x^*)\big] &\leq \frac{12\eta Ll}{(1 - \eta L)}\big(\frac{1}{2}\|x_t^s - x^*\|^2 + \frac{1}{2}\|\widetilde{x}^{s-1} - x^*\|^2\big) \\ &\quad + \frac{(1 - \sigma\eta)\|x_t^s - x^*\|^2 - \mathbb{E}_{i_t^s}\|x_{t+1}^s - x^*\|^2}{2\eta} \ . \end{aligned}$$

Choosing $\eta = \frac{\sigma}{25Ll} \leq \frac{1}{25L}$ in the above inequality, we obtain that

$$\mathbb{E}_{i_t^s}\big[F(x_{t+1}^s) - F(x^*)\big] \leq \frac{\sigma}{4}\|\widetilde{x}^{s-1} - x^*\|^2 + \frac{\|x_t^s - x^*\|^2 - \mathbb{E}_{i_t^s}\|x_{t+1}^s - x^*\|^2}{2\eta} \ .$$

Summing it up over $t = 0, 1, \dots, m - 1$, and dividing both sides by $m$, we arrive at

$$\mathbb{E}\Big[\sum_{t=0}^{m-1} \frac{F(x_{t+1}^s)}{m} - F(x^*)\Big] \leq \mathbb{E}\Big[\frac{\|x_0^s - x^*\|^2}{2\eta \cdot m} + \frac{\sigma}{4}\|\widetilde{x}^{s-1} - x^*\|^2\Big] \ .$$

Finally, using our choice $x_0^s = \widetilde{x}^{s-1}$, using the convexity of $F(\cdot)$ which tells us $F(\widetilde{x}^s) \leq \frac{1}{m}\sum_{t=1}^m F(x_t^s)$, and using the strong convexity of $F(\cdot)$ which tells us $\frac{\sigma}{2}\|\widetilde{x}^{s-1} - x^*\|^2 \leq F(\widetilde{x}^{s-1}) - F(x^*)$, we conclude from the above inequality that

$$\mathbb{E}[F(\widetilde{x}^s) - F(x^*)] \leq \frac{2 + \frac{4}{\sigma\eta m}}{4}\big(F(\widetilde{x}^{s-1}) - F(x^*)\big) \ .$$ □

# D Convergence Analysis for Section 7

We use the same notations of $i_t^s$ and $\xi_t^s$ as in previous sections. The following lemma is exactly Lemma A.1 where the step length $\eta$ is replaced with $\eta_{t+1}^s$:

**Lemma D.1** (Lemma A.1 revised)**.** *For every $u \in \mathbb{R}^d$ and $t \in \{0, 1, \ldots, m_s - 1\}$, fixing $x_t^s$ and letting $i = i_t^s$ be the random variable, we have*

$$\mathbb{E}_{i_t^s}\big[F(x_{t+1}^s) - F(u)\big] \le \mathbb{E}_{i_t^s}\Big[\frac{\eta_{t+1}^s}{2(1-\eta_{t+1}^s L)}\|\xi_t^s - \nabla f(x_t^s)\|^2 + \frac{\|x_t^s - u\|^2 - \|x_{t+1}^s - u\|^2}{2\eta_{t+1}^s}\Big] \ .$$

Also, by combining Lemma 5.2 (for $l \le L$) and Lemma C.1 (for $l \ge L$), we have that for every $l \ge 0$,

**Lemma D.2.**

$$\begin{aligned}\mathbb{E}_{i_t^s}\big[\|\xi_t^s - \nabla f(x_t^s)\|^2\big] \le\ & 8L \cdot \big(F(x_t^s) - F(x^*) + F(\widetilde{x}^{s-1}) - F(x^*)\big) \\ & + 12Ll\big(\|x_t^s - x^*\|^2 + \|\widetilde{x}^{s-1} - x^*\|^2\big) \ .\end{aligned}$$

Now we are ready to prove a lemma that is different from all previous sections.

**Lemma D.3.** *If $m^0 \ge 1$, $\eta \le 1/13L$, and $\frac{1}{4\sqrt{2}T\eta} \ge 39\eta Ll$, we have*

$$\mathbb{E}[F(\widetilde{x}^S) - F(x^*)] \le \frac{F(x^\phi) - F(x^*)}{2^{S-1}} + \frac{39\eta Ll\|x^\phi - x^*\|^2}{2^S} + \frac{\|x^\phi - x^*\|^2}{2^S \cdot \frac{4\eta_0^1 m_0}{3}} \ . \tag{D.1}$$

*Proof.* Combining Lemma D.1 with $u = x^*$ and Lemma D.2, as well as using the fact that $\eta_{t+1}^s \le \eta$, we have

$$\begin{aligned}\mathbb{E}_{i_t^s}\big[F(x_{t+1}^s) - F(x^*)\big] \le\ & \frac{4\eta L}{(1-\eta L)}\big(F(x_t^s) - F(x^*) + F(\widetilde{x}^{s-1}) - F(x^*) + 3l\|x_t^s - x^*\|^2 + 3l\|\widetilde{x}^{s-1} - x^*\|^2\big) \\ & + \frac{\|x_t^s - x^*\|^2 - \mathbb{E}_{i_t^s}\|x_{t+1}^s - x^*\|^2}{2\eta_{t+1}^s} \ .\end{aligned}$$

Choosing $\eta \le 1/13L$ in the above inequality, we have

$$\begin{aligned}\mathbb{E}_{i_t^s}\big[F(x_{t+1}^s) - F(x^*)\big] &\le \frac{1}{3}\big(F(x_t^s) - F(x^*) + F(\widetilde{x}^{s-1}) - F(x^*)\big) + 13\eta Ll\big(\|x_t^s - x^*\|^2 + \|\widetilde{x}^{s-1} - x^*\|^2\big) \\ &\quad + \frac{\|x_t^s - x^*\|^2 - \mathbb{E}_{i_t^s}\|x_{t+1}^s - x^*\|^2}{2\eta_{t+1}^s} \\ &\le \frac{1}{3}\big(F(x_t^s) - F(x^*) + F(\widetilde{x}^{s-1}) - F(x^*)\big) + 13\eta Ll\big(-2\|x_t^s - x^*\|^2 + \|\widetilde{x}^{s-1} - x^*\|^2\big) \\ &\quad + \frac{\|x_t^s - x^*\|^2}{2\eta_t^s} - \frac{\mathbb{E}_{i_t^s}\|x_{t+1}^s - x^*\|^2}{2\eta_{t+1}^s} \ .\end{aligned}$$

where the last inequality uses (7.1) and the assumption that $\frac{1}{4\sqrt{2}T\eta} \ge 39\eta Ll$.

Summing it up over $t = 0, 1, \ldots, m_s - 1$ and dividing both sides by $m_s$, we arrive at

$$\begin{aligned}\mathbb{E}\Big[\sum_{t=0}^{m_s-1}\frac{F(x_{t+1}^s) + 26\eta Ll\|x_t^s - x^*\|^2}{m_s} - F(x^*)\Big] \le\ & \mathbb{E}\Big[\frac{1}{3}\Big(\sum_{t=0}^{m_s-1}\frac{F(x_t^s)}{m_s} - F(x^*) + F(\widetilde{x}^{s-1}) - F(x^*)\Big) \\ & + 13 \cdot \|\widetilde{x}^{s-1} - x^*\|^2 + \frac{\|x_0^s - x^*\|^2}{2\eta_0^s \cdot m_s} - \frac{\|x^* - x_{m_s}^s\|^2}{2\eta_{m_s}^s \cdot m_s}\Big] \ .\end{aligned}$$

After rearranging, this yields

$$2\mathbb{E}\Big[\sum_{t=0}^{m_s-1}\frac{F(x_t^s)+39\eta Ll\|x_t^s-x^*\|^2}{m_s}-F(x^*)\Big]\le\mathbb{E}\Big[\frac{3(F(x_0^s)-F(x^*))-3(F(x_{m_s}^s)-F(x^*))}{m_s}+F(\widetilde{x}^{s-1})-F(x^*)$$
$$+39\cdot\|\widetilde{x}^{s-1}-x^*\|^2+\frac{\|x_0^s-x^*\|^2}{2\eta_0^s/3\cdot m_s}-\frac{\|x^*-x_{m_s}^s\|^2}{2\eta_{m_s}^s/3\cdot m_s}\Big]\ .$$

Next, using the fact that $F(\widetilde{x}^s)\le\sum_{t=0}^{m_s-1}\frac{F(x_t^s)}{m_s}$ and $\|\widetilde{x}^s-x^*\|^2\le\frac{1}{m_s}\sum_{t=0}^{m_s-1}\|x_t^s-x^*\|^2$ which follow from convexity and the definition $\widetilde{x}^s=\sum_{t=0}^{m_s-1}\frac{x_t^s}{m_s}$, we can we rewrite the above inequality as

$$2\mathbb{E}\big[F(\widetilde{x}^s)-F(x^*)+39\eta Ll\|\widetilde{x}^s-x^*\|^2\big]\le\mathbb{E}\Big[\frac{3(F(x_0^s)-F(x^*))-3(F(x_{m_s}^s)-F(x^*))}{m_s}+F(\widetilde{x}^{s-1})-F(x^*)\big)$$
$$+39\cdot\|\widetilde{x}^{s-1}-x^*\|^2+\frac{\|x_0^s-x^*\|^2}{2\eta_0^s/3\cdot m_s}-\frac{\|x^*-x_{m_s}^s\|^2}{2\eta_{m_s}^s/3\cdot m_s}\Big]$$

At this point, let us recall choice $x_{m_s}^s=x_0^{s+1}$, $\eta_{m_s}^s=\eta_0^{s+1}$, and $m_s=2m_{s-1}$, which yield

$$2\mathbb{E}\big[F(\widetilde{x}^s)-F(x^*)+39\eta Ll\|\widetilde{x}^s-x^*\|^2+\frac{\|x^*-x_0^{s+1}\|^2}{4\eta_0^{s+1}/3\cdot m_s}+\frac{F(x_0^{s+1})-F(x^*)}{2m_s/3}\big]$$
$$\le\mathbb{E}\Big[F(\widetilde{x}^{s-1})-F(x^*)+39\eta Ll\|\widetilde{x}^{s-1}-x^*\|^2+\frac{\|x_0^s-x^*\|^2}{4\eta_0^s/3\cdot m_{s-1}}+\frac{F(x_0^s)-F(x^*)}{2m_{s-1}/3}\Big]\ .$$

In sum, after telescoping for $s=1,2,\dots,S$, we have

$$\mathbb{E}[F(\widetilde{x}^S)-F(x^*)]\le 2^{-S}\cdot\Big(F(\widetilde{x}^0)-F(x^*)+39\eta Ll\|\widetilde{x}^0-x^*\|^2+\frac{\|x^*-x_0^1\|^2}{4\eta_0^1/3\cdot m_0}+\frac{F(x_0^1)-F(x^*)}{2m_0}\Big)$$
$$\le\frac{F(x^\phi)-F(x^*)}{2^{S-1}}+\frac{39\eta Ll\|x^\phi-x^*\|^2}{2^S}+\frac{\|x^\phi-x^*\|^2}{2^S\cdot\frac{4\eta m_0}{3\sqrt{2}}}\ .$$

□

Finally, the above lemma immediately yields our desired theorem:

*Proof of Theorem 7.1.* Under the given parameter choices, we first have

$$\frac{1}{4\sqrt{2}T\eta}\ge\frac{1}{4\sqrt{2}\eta\cdot 2m_0\cdot 2^S}=\frac{1}{8\sqrt{2}\eta m_0\cdot\frac{\Delta}{\varepsilon}}=\frac{\varepsilon}{8\sqrt{2}\Theta}=39\cdot\frac{\varepsilon}{312\sqrt{2}\Theta}\ge 39\eta Ll$$

so the preassumption of Lemma D.3 holds.

Now we consider the three terms on the right hand side of (D.1). The first term is no more than $\frac{2\Delta}{2^S}\le 2\varepsilon$. The second term is no more than

$$\frac{39\eta Ll\Theta}{2^S}=\frac{39\eta Ll\Theta}{\Delta}\varepsilon\le\frac{\varepsilon}{8\sqrt{2}\Delta}\varepsilon\le\frac{\varepsilon}{8\sqrt{2}}\ .$$

The third term is no more than

$$\frac{\Theta}{\Delta/\varepsilon\cdot\frac{4\eta m_0}{3\sqrt{2}}}=\frac{\Theta}{1/\varepsilon\cdot\frac{4\Theta}{3\sqrt{2}}}=\frac{3\sqrt{2}}{4}\varepsilon\ .$$

In sum, we conclude that $\mathbb{E}[F(\widetilde{x}^S)-F(x^*)]\le O(\varepsilon)$. □

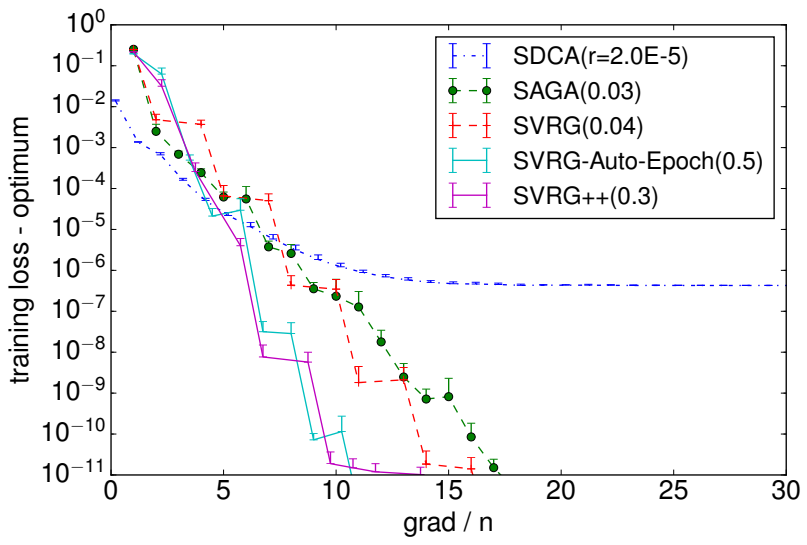

(a) Adult, Lasso $\sigma = 10^{-3}$

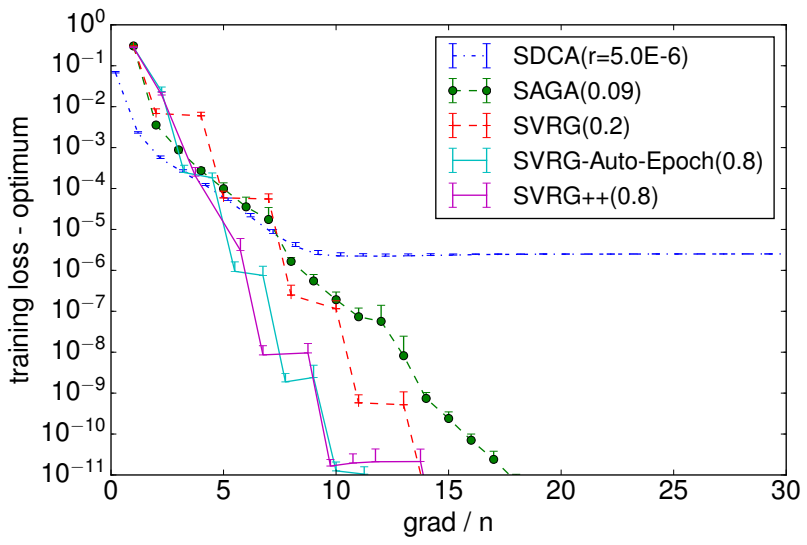

(b) Adult, Logistic $\sigma = 10^{-3}$

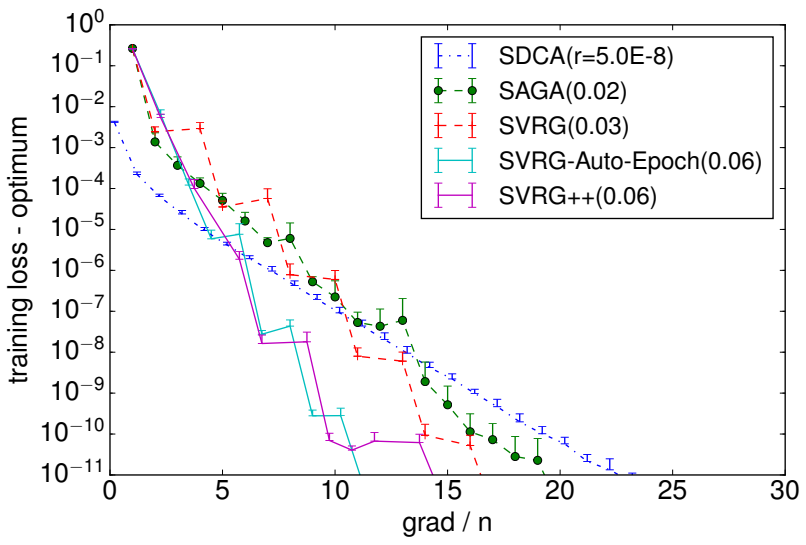

(c) Adult, Ridge $\sigma = 10^{-3}$

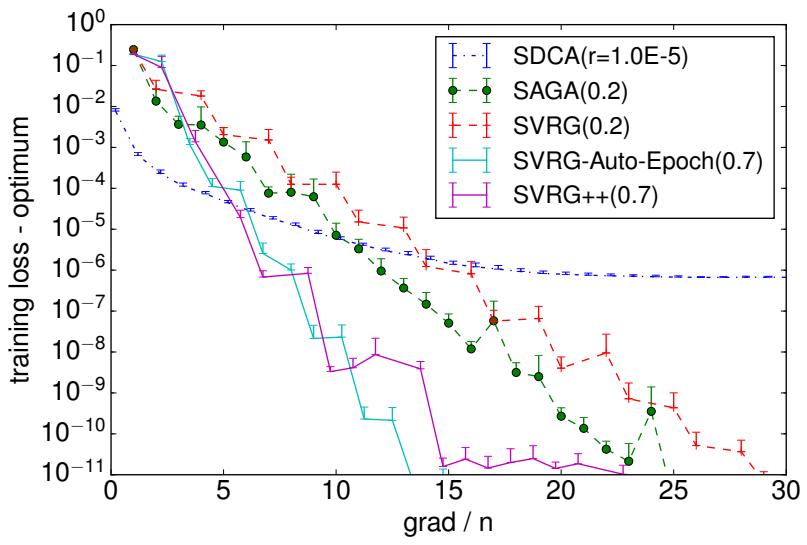

(d) Adult, Lasso $\sigma = 10^{-4}$

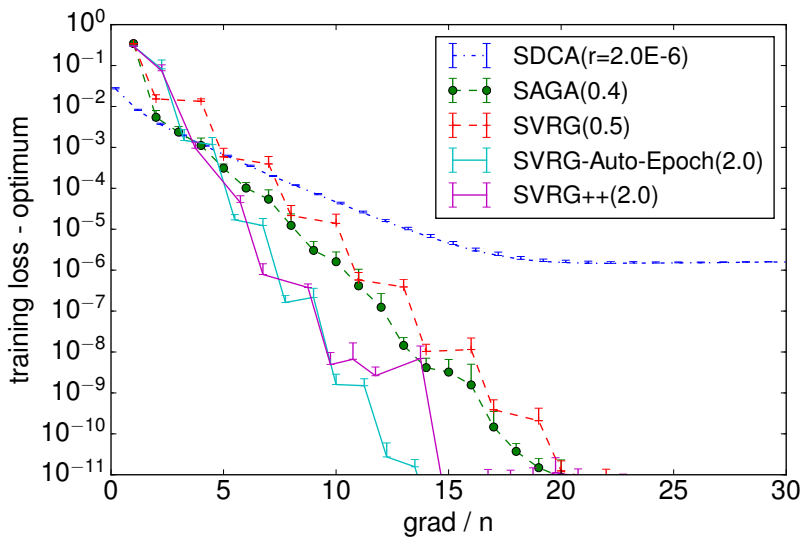

(e) Adult, Logistic $\sigma = 10^{-4}$

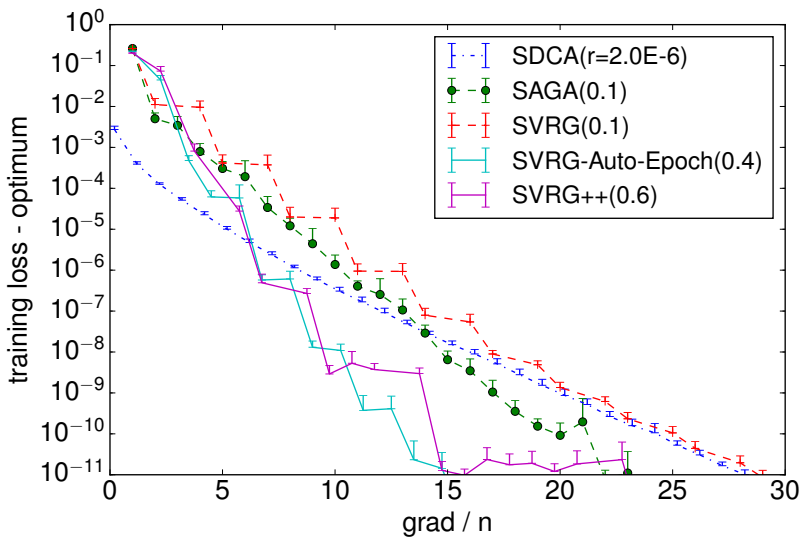

(f) Adult, Ridge $\sigma = 10^{-4}$

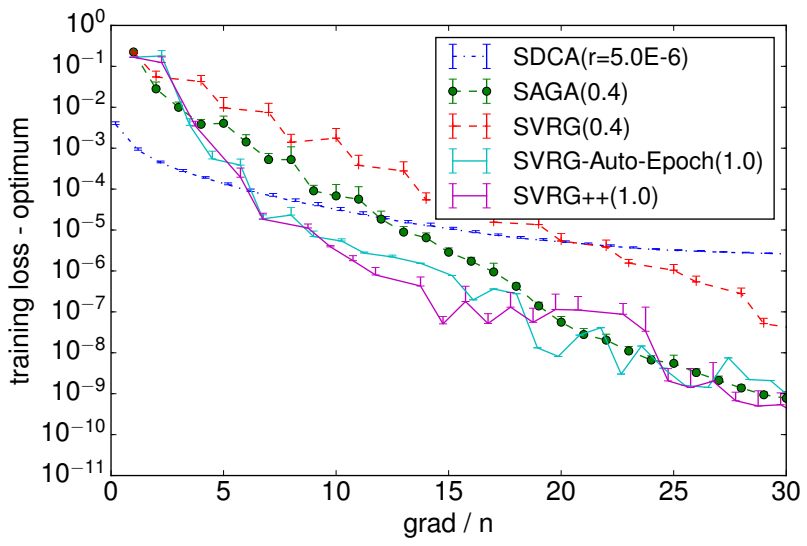

(g) Adult, Lasso $\sigma = 10^{-5}$

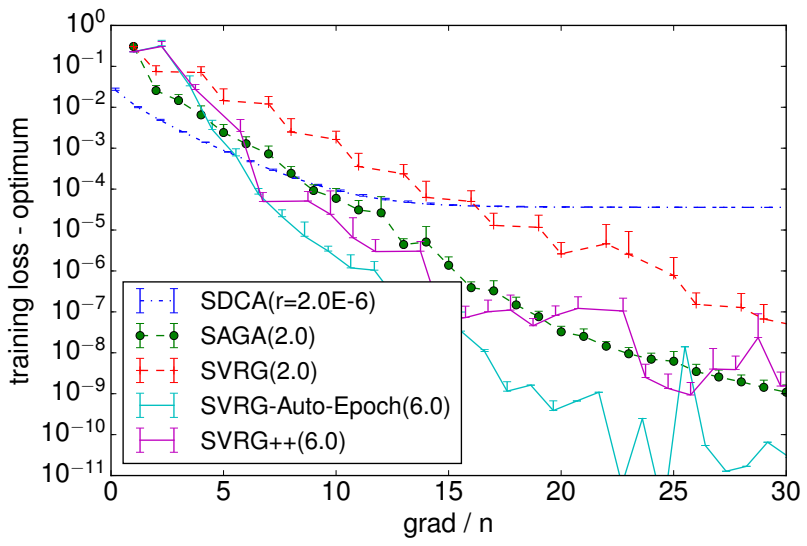

(h) Adult, Logistic $\sigma = 10^{-5}$

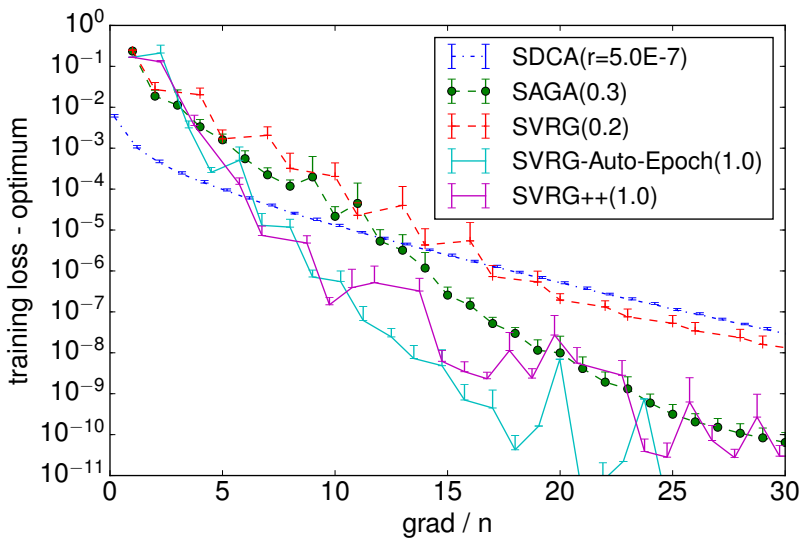

(i) Adult, Ridge $\sigma = 10^{-5}$

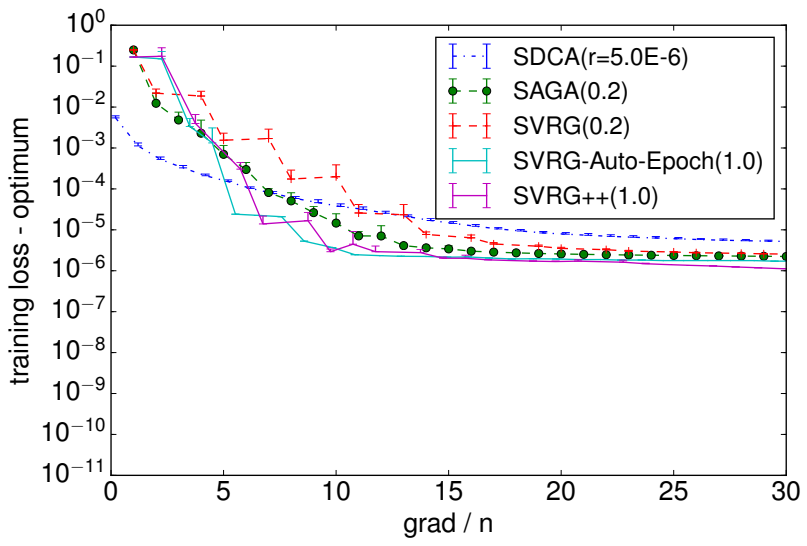

(j) Adult, Lasso $\sigma = 10^{-6}$

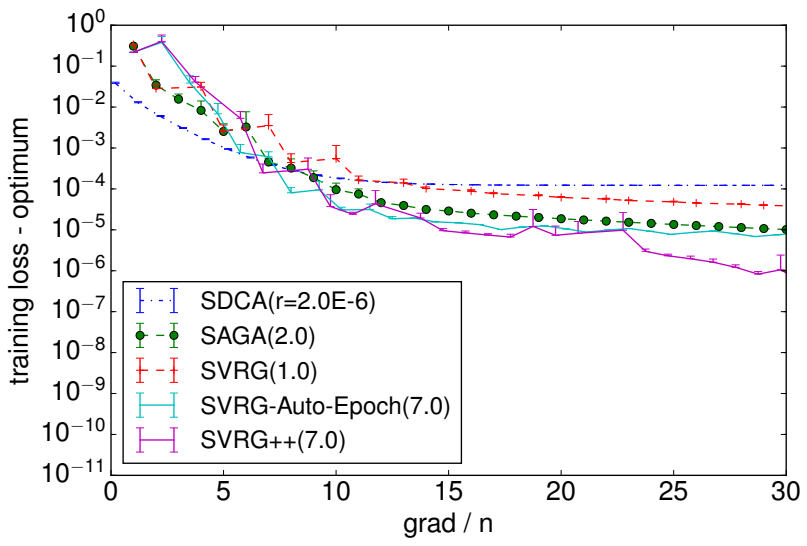

(k) Adult, Logistic $\sigma = 10^{-6}$

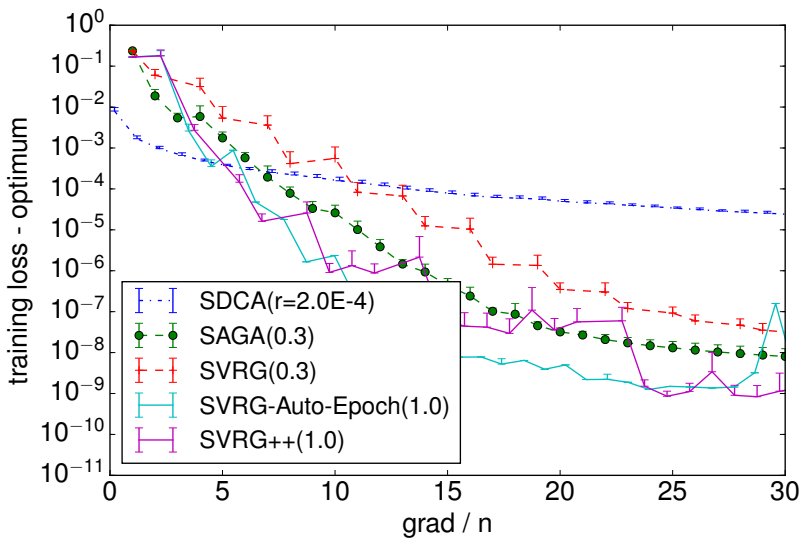

(l) Adult, Ridge $\sigma = 10^{-6}$

Figure 3: Training error comparisons on dataset Adult, using Tuning Type I.

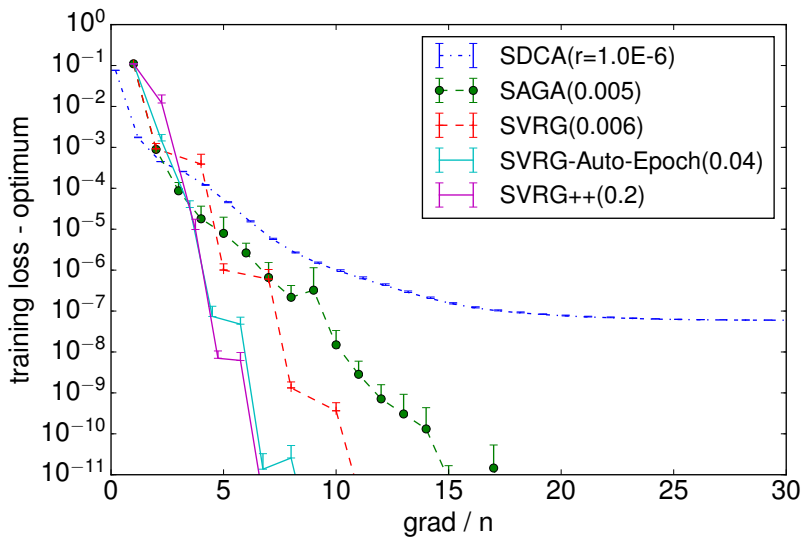

(a) Covtype, Lasso $\sigma = 10^{-3}$

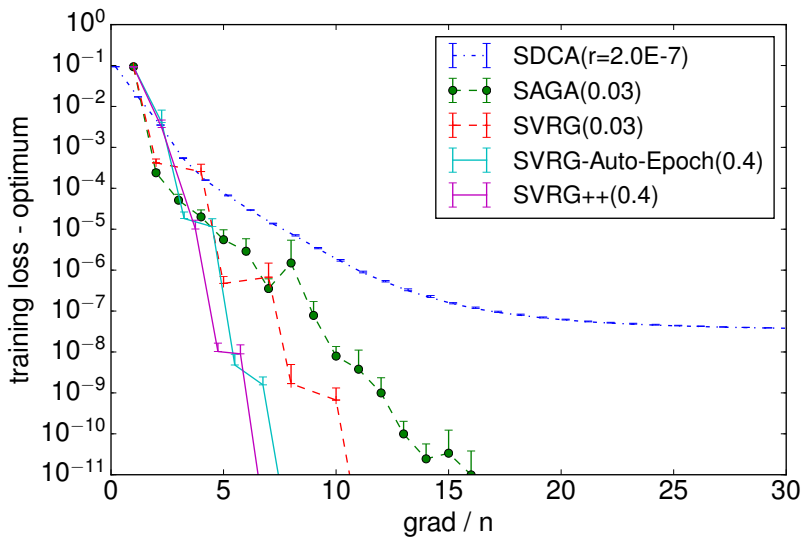

(b) Covtype, Logistic $\sigma = 10^{-3}$

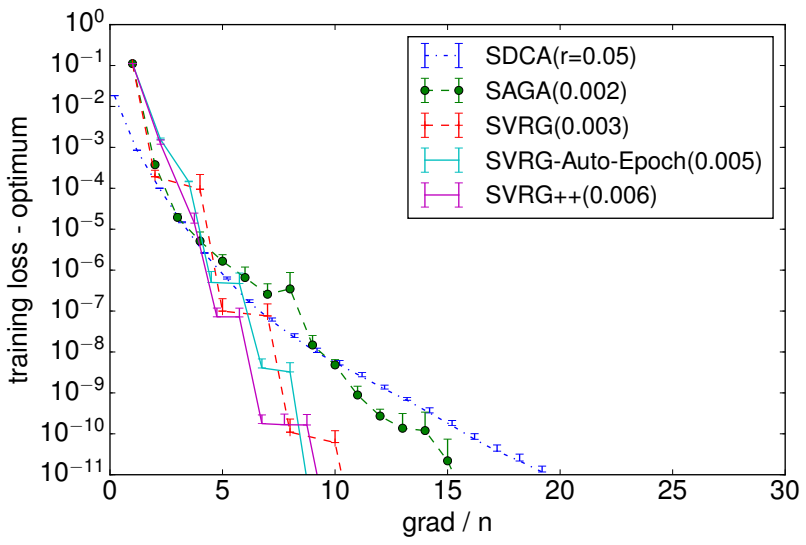

(c) Covtype, Ridge $\sigma = 10^{-3}$

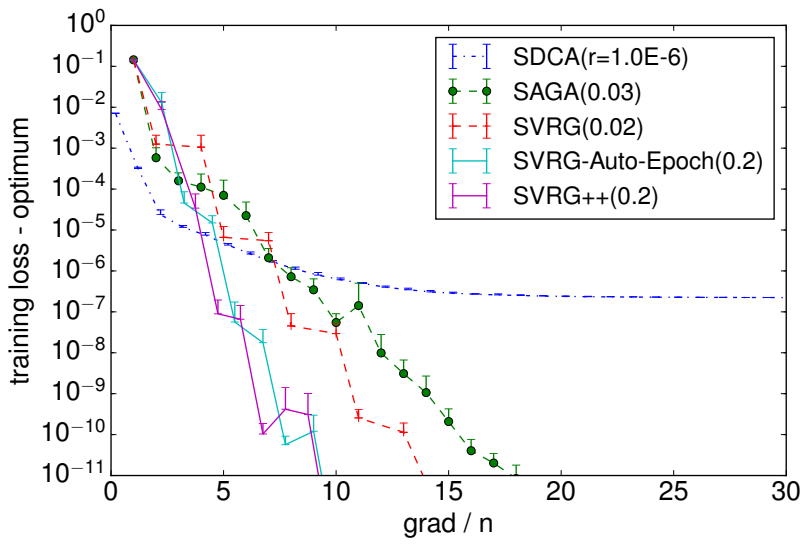

(d) Covtype, Lasso $\sigma = 10^{-4}$

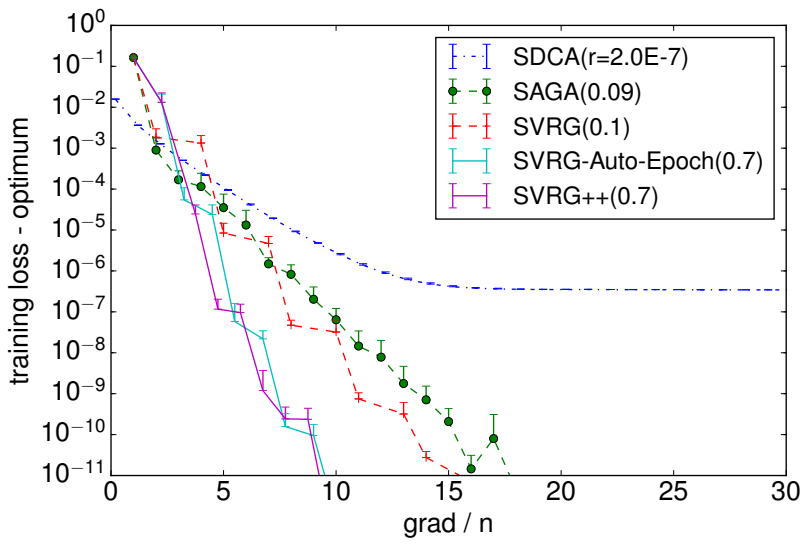

(e) Covtype, Logistic $\sigma = 10^{-4}$

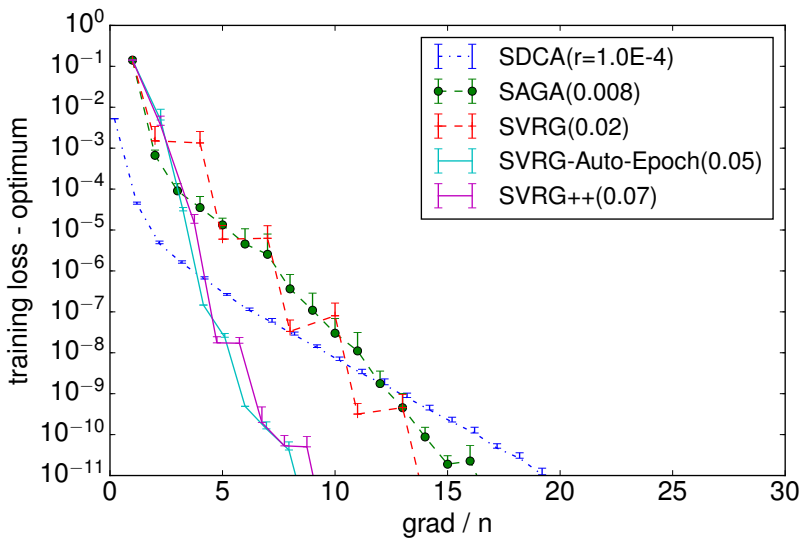

(f) Covtype, Ridge $\sigma = 10^{-4}$

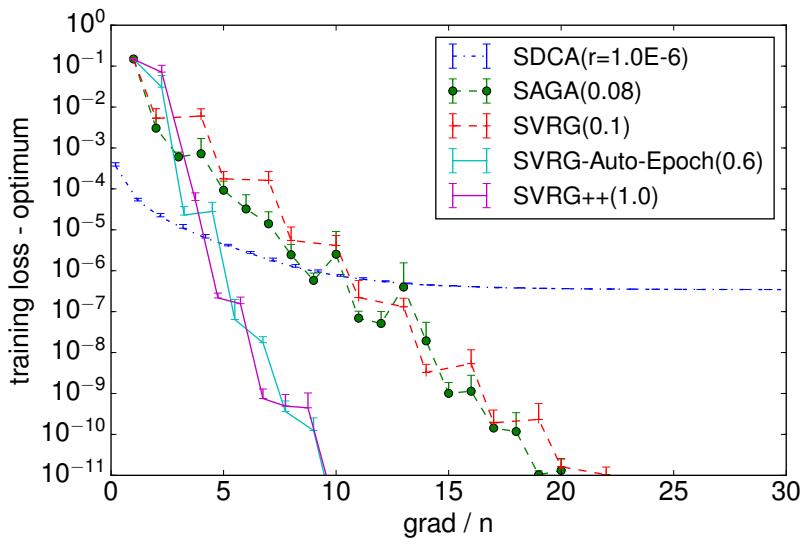

(g) Covtype, Lasso $\sigma = 10^{-5}$

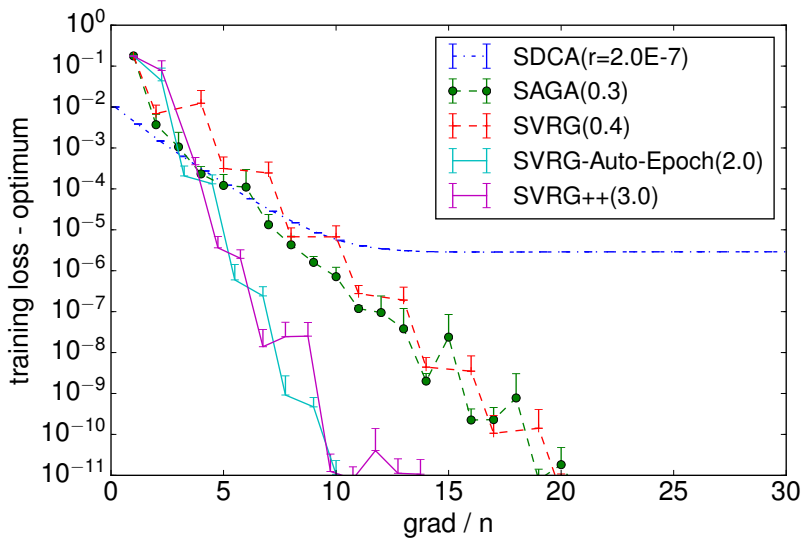

(h) Covtype, Logistic $\sigma = 10^{-5}$

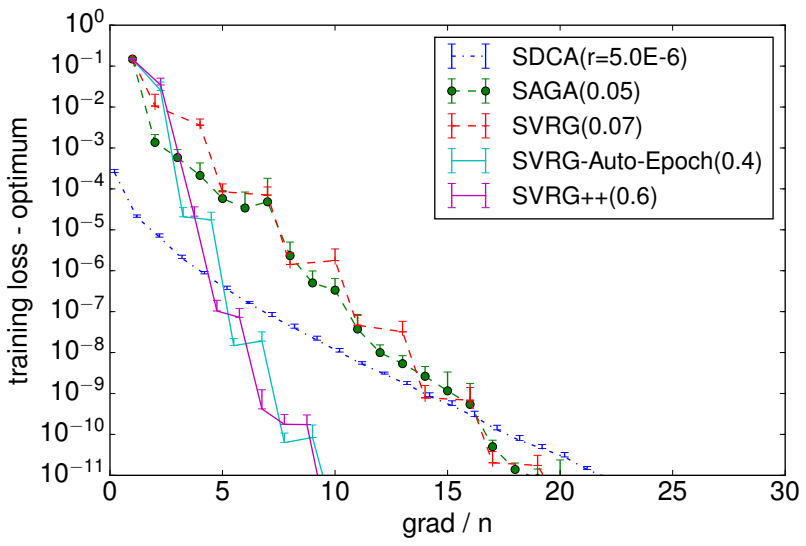

(i) Covtype, Ridge $\sigma = 10^{-5}$

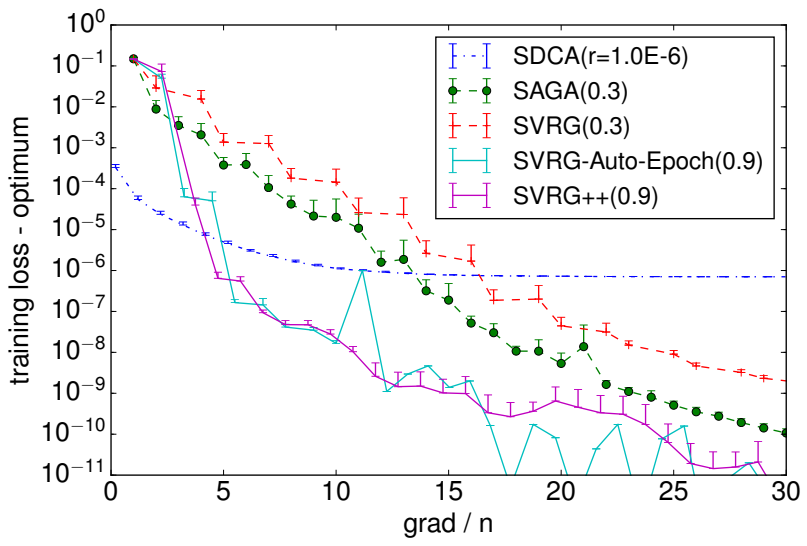

(j) Covtype, Lasso $\sigma = 10^{-6}$

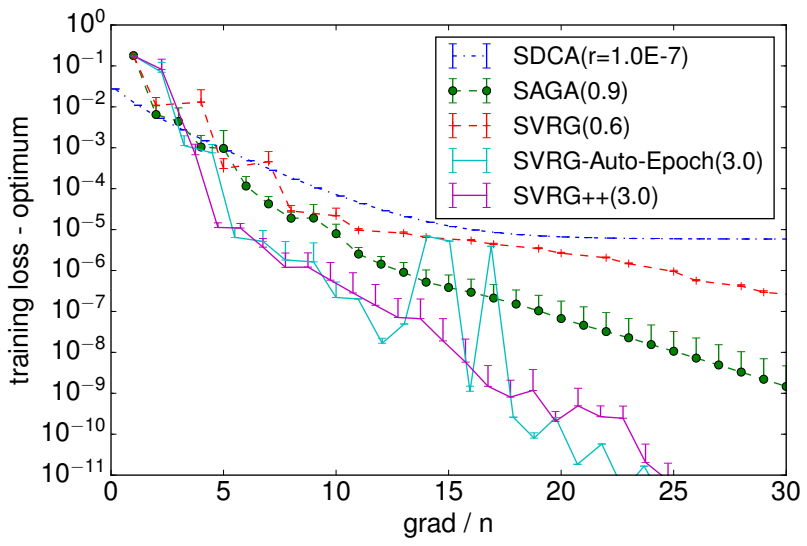

(k) Covtype, Logistic $\sigma = 10^{-6}$

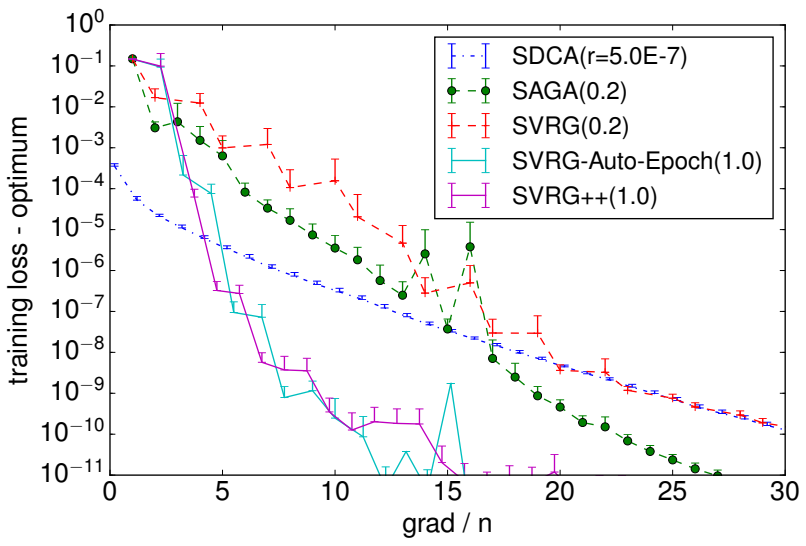

(l) Covtype, Ridge $\sigma = 10^{-6}$

Figure 4: Training error comparisons on dataset Covtype, using Tuning Type I.

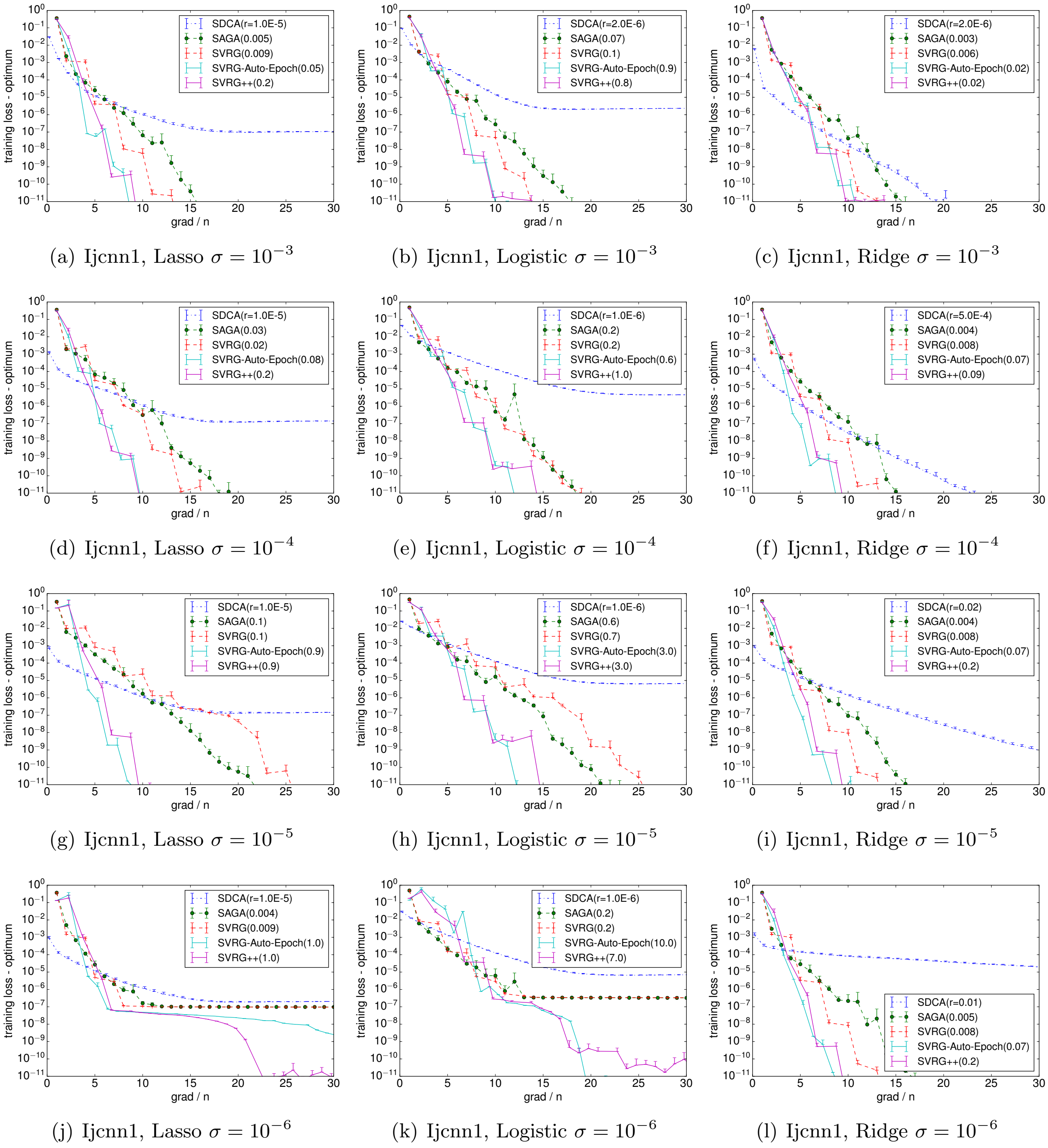

(a) Ijcnn1, Lasso $\sigma = 10^{-3}$ (b) Ijcnn1, Logistic $\sigma = 10^{-3}$ (c) Ijcnn1, Ridge $\sigma = 10^{-3}$

(d) Ijcnn1, Lasso $\sigma = 10^{-4}$ (e) Ijcnn1, Logistic $\sigma = 10^{-4}$ (f) Ijcnn1, Ridge $\sigma = 10^{-4}$

(g) Ijcnn1, Lasso $\sigma = 10^{-5}$ (h) Ijcnn1, Logistic $\sigma = 10^{-5}$ (i) Ijcnn1, Ridge $\sigma = 10^{-5}$

(j) Ijcnn1, Lasso $\sigma = 10^{-6}$ (k) Ijcnn1, Logistic $\sigma = 10^{-6}$ (l) Ijcnn1, Ridge $\sigma = 10^{-6}$

Figure 5: Training error comparisons on dataset Ijcnn1, using Tuning Type I.

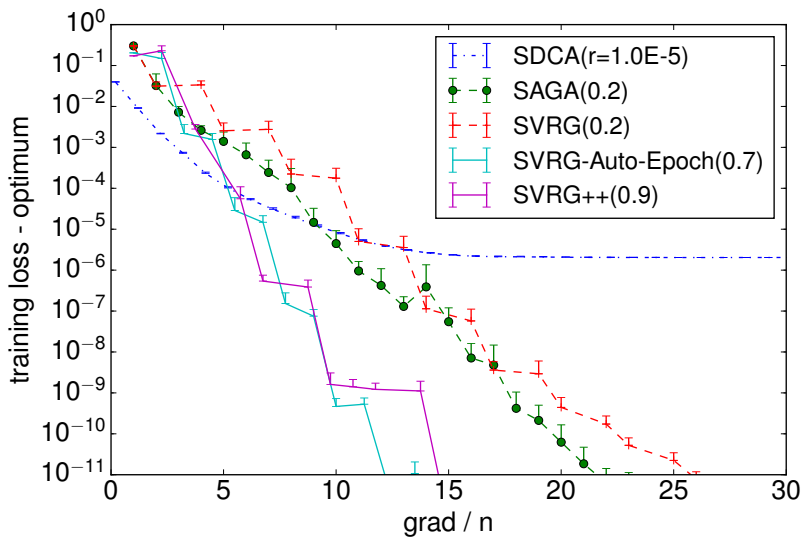

(a) Mnist, Lasso $\sigma = 10^{-3}$

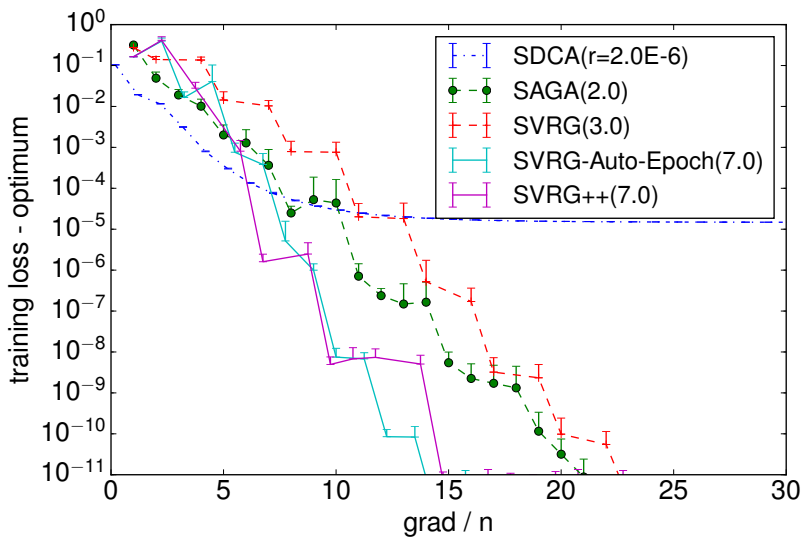

(b) Mnist, Logistic $\sigma = 10^{-3}$

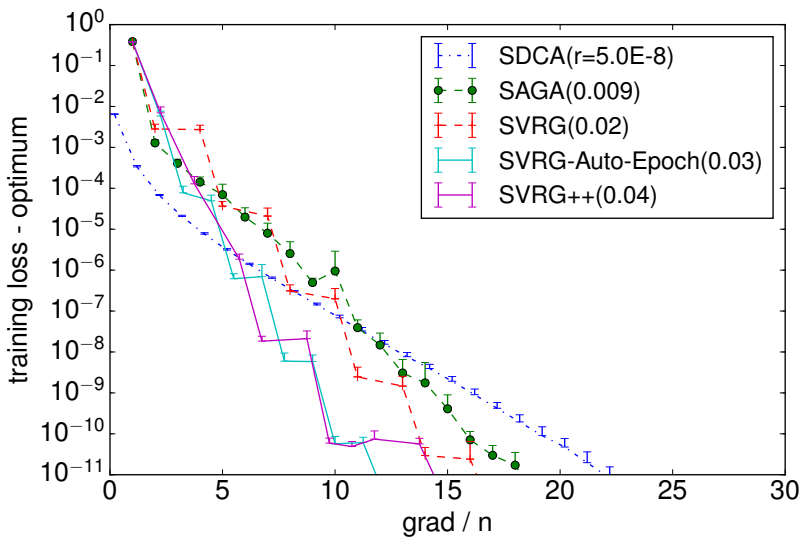

(c) Mnist, Ridge $\sigma = 10^{-3}$

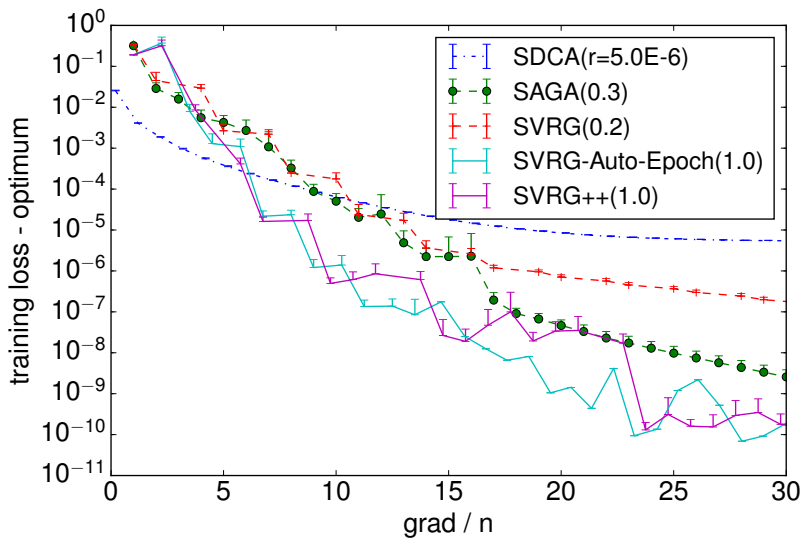

(d) Mnist, Lasso $\sigma = 10^{-4}$

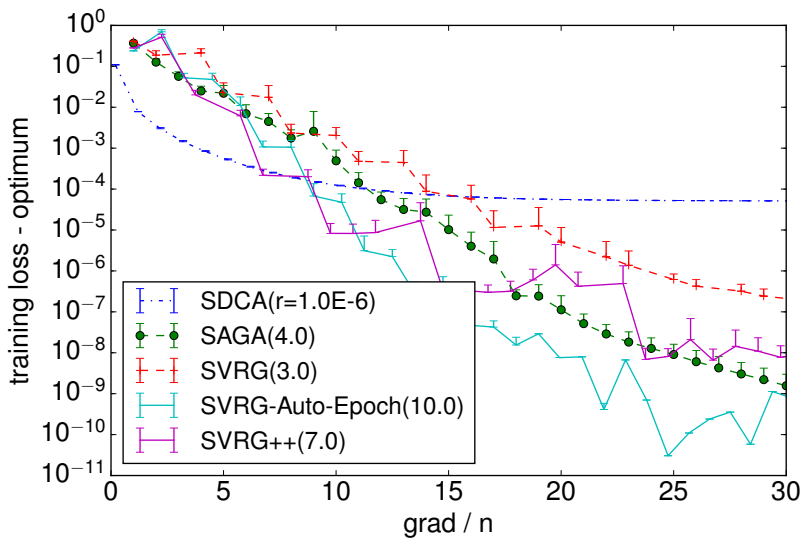

(e) Mnist, Logistic $\sigma = 10^{-4}$

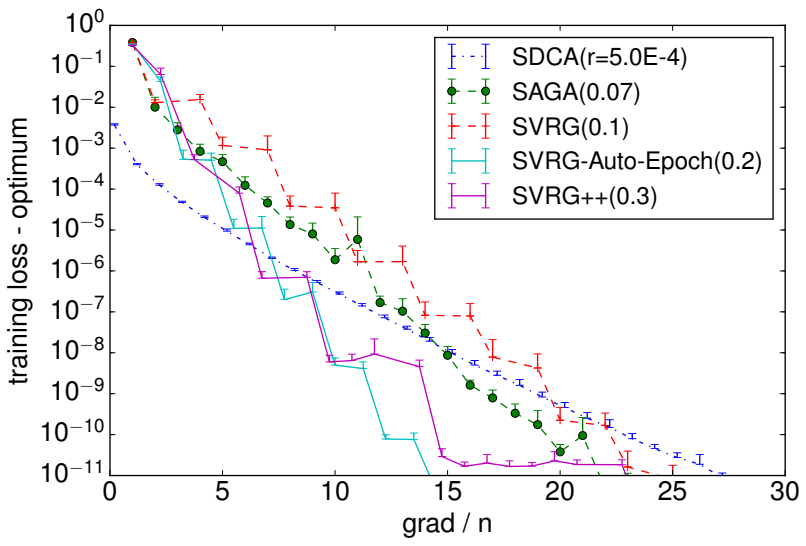

(f) Mnist, Ridge $\sigma = 10^{-4}$

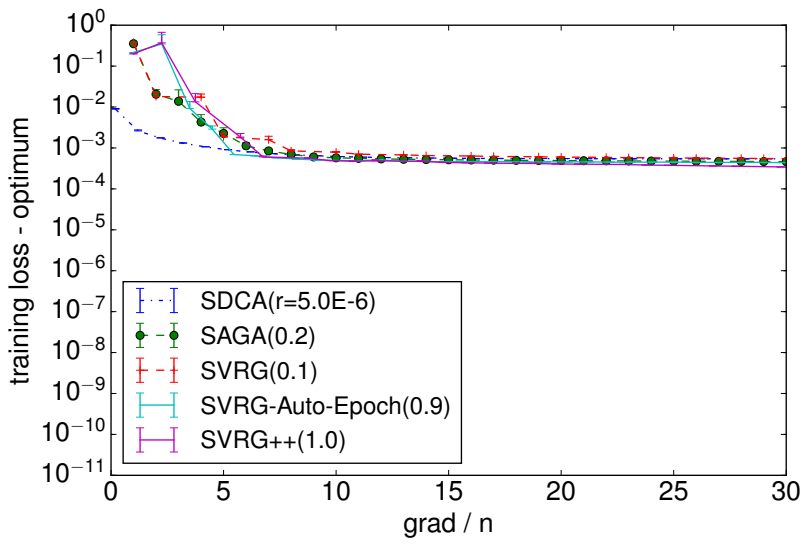

(g) Mnist, Lasso $\sigma = 10^{-5}$

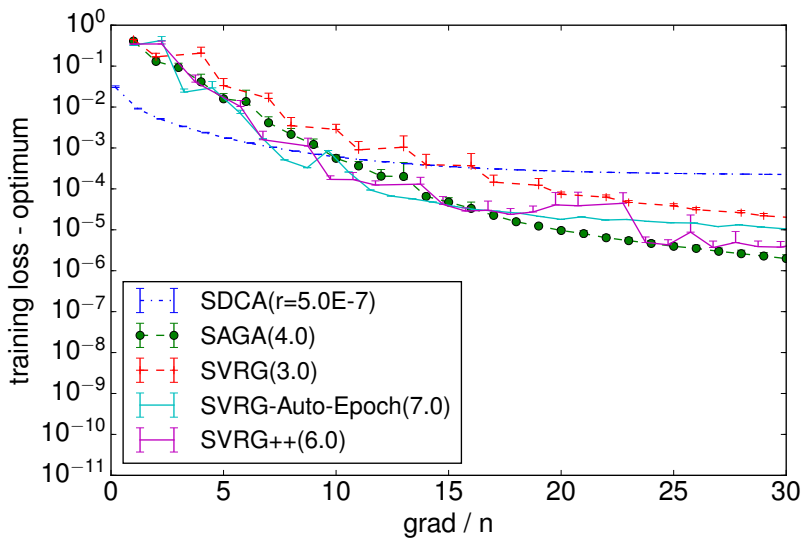

(h) Mnist, Logistic $\sigma = 10^{-5}$

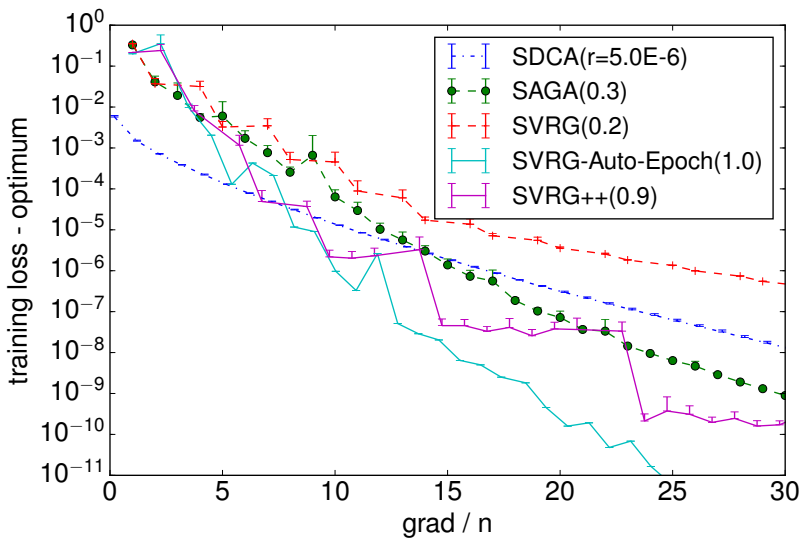

(i) Mnist, Ridge $\sigma = 10^{-5}$

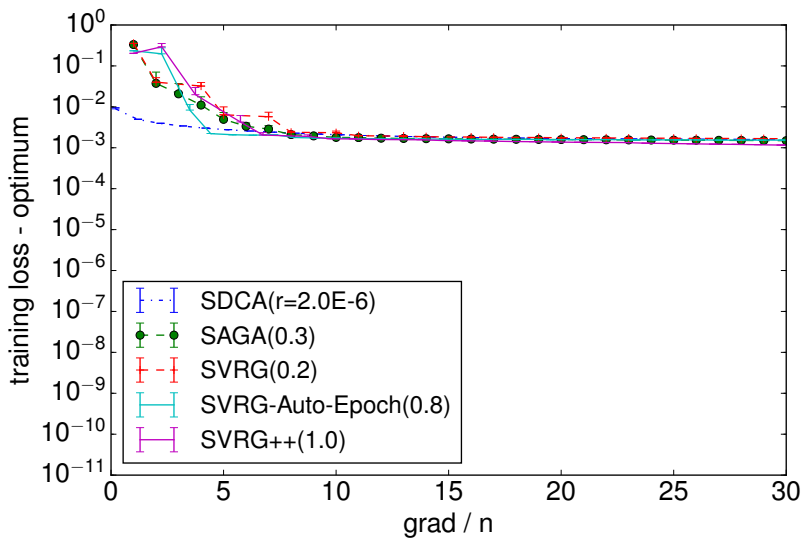

(j) Mnist, Lasso $\sigma = 10^{-6}$

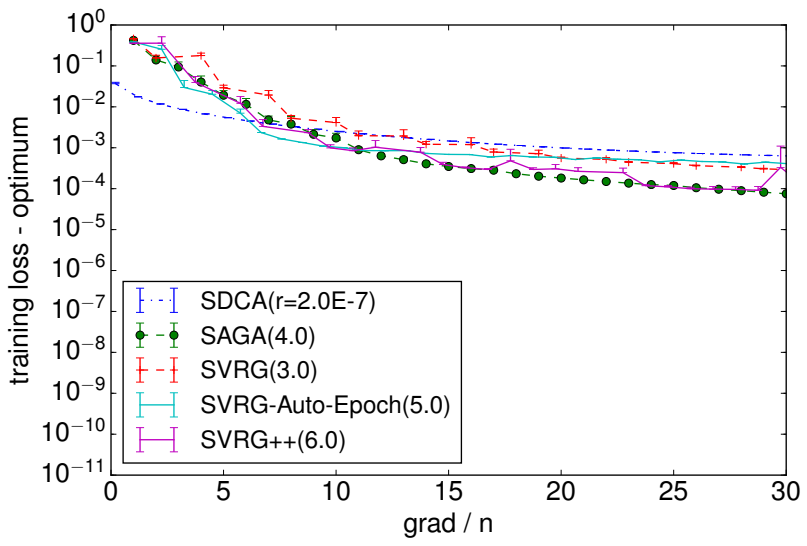

(k) Mnist, Logistic $\sigma = 10^{-6}$

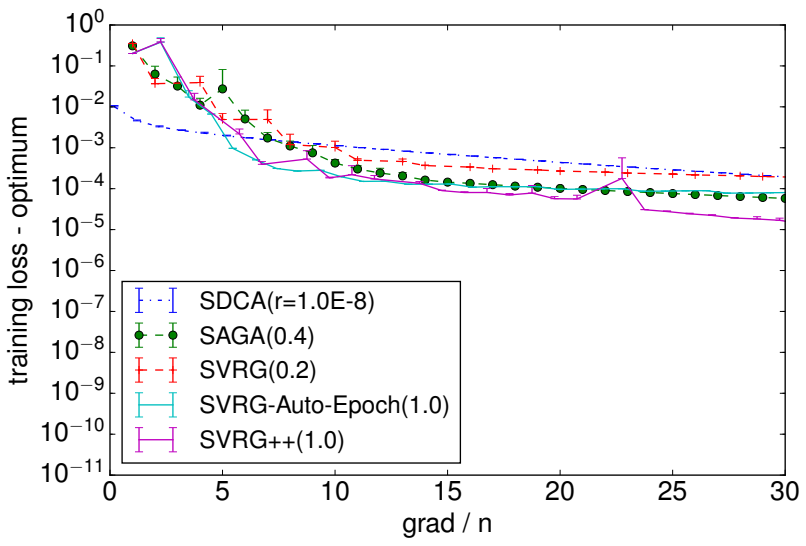

(l) Mnist, Ridge $\sigma = 10^{-6}$

Figure 6: Training error comparisons on dataset mnist, using Tuning Type I.

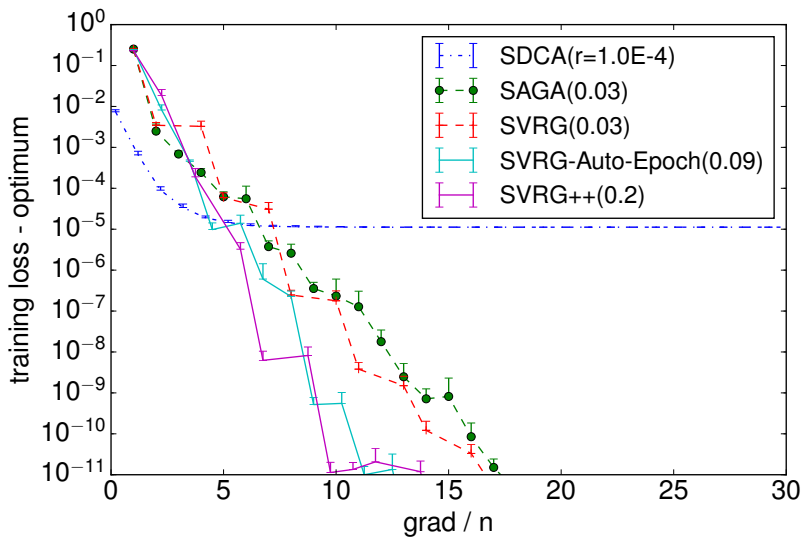

(a) Adult, Lasso $\sigma = 10^{-3}$

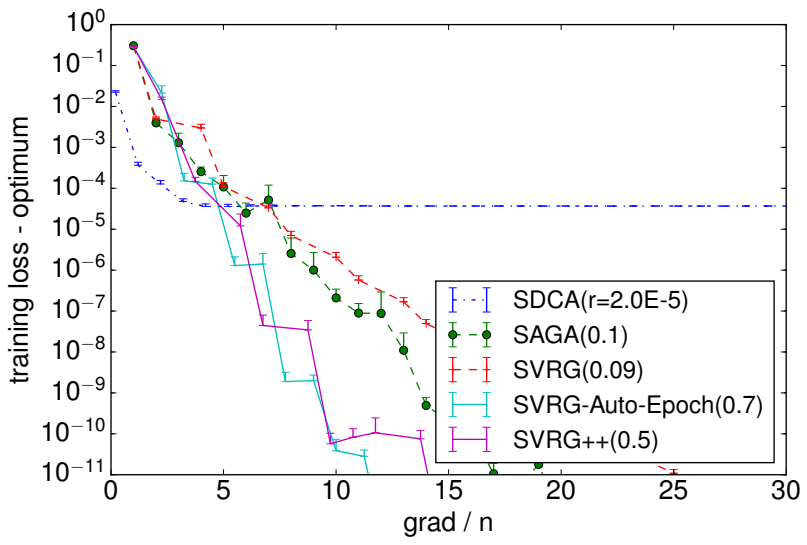

(b) Adult, Logistic $\sigma = 10^{-3}$

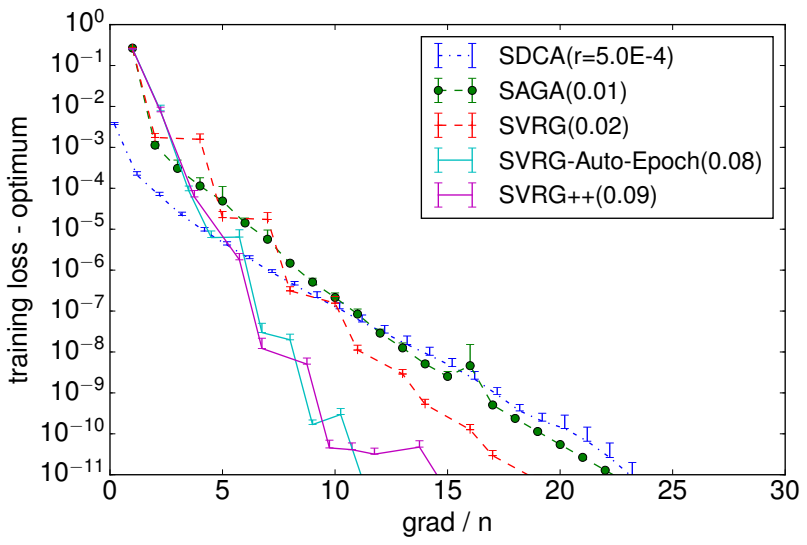

(c) Adult, Ridge $\sigma = 10^{-3}$

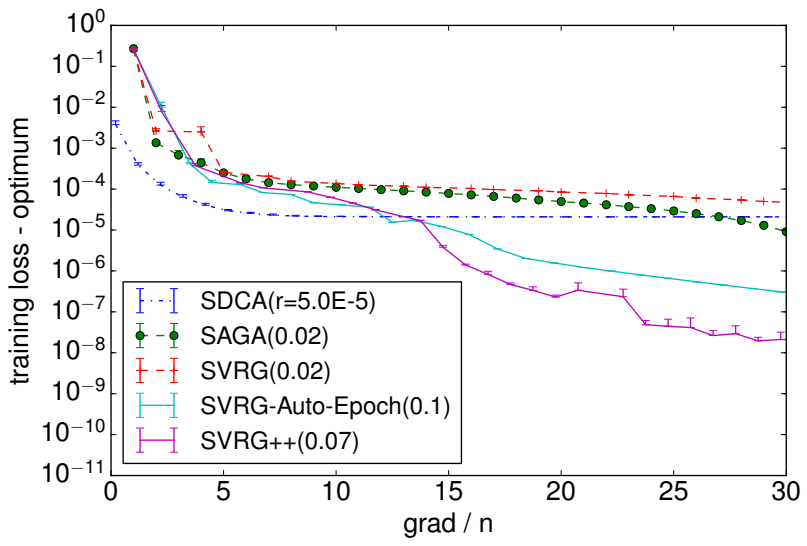

(d) Adult, Lasso $\sigma = 10^{-4}$

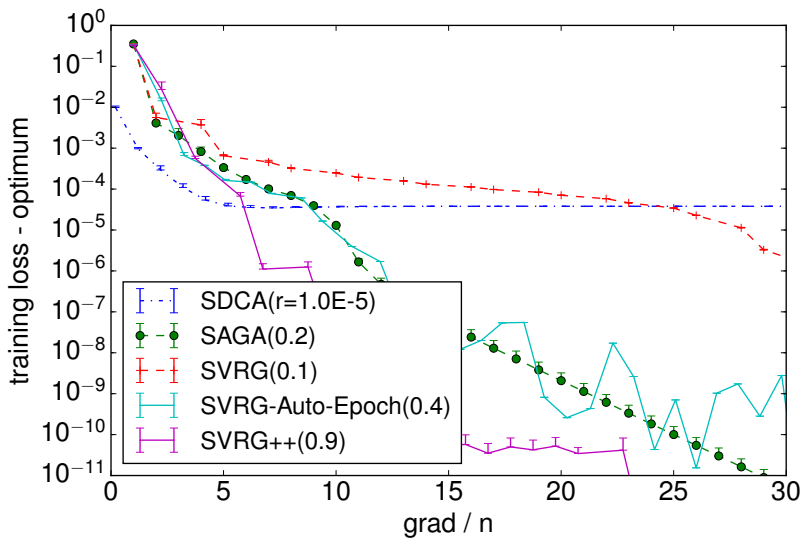

(e) Adult, Logistic $\sigma = 10^{-4}$

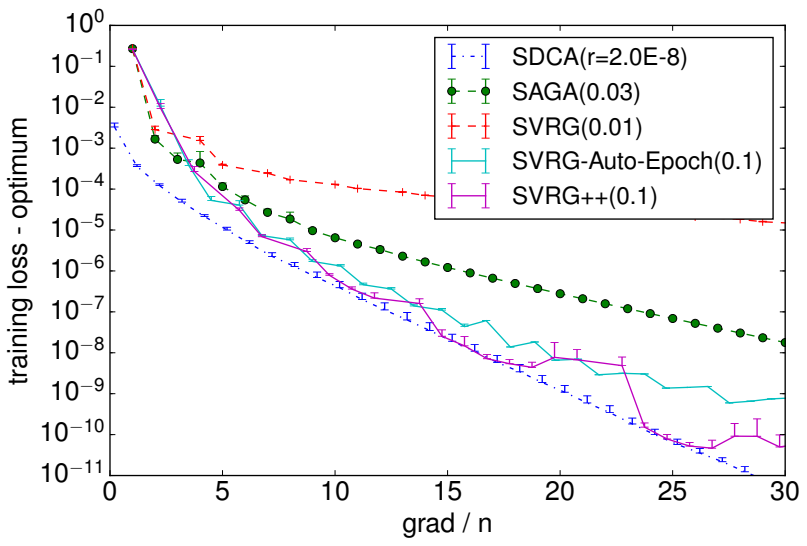

(f) Adult, Ridge $\sigma = 10^{-4}$

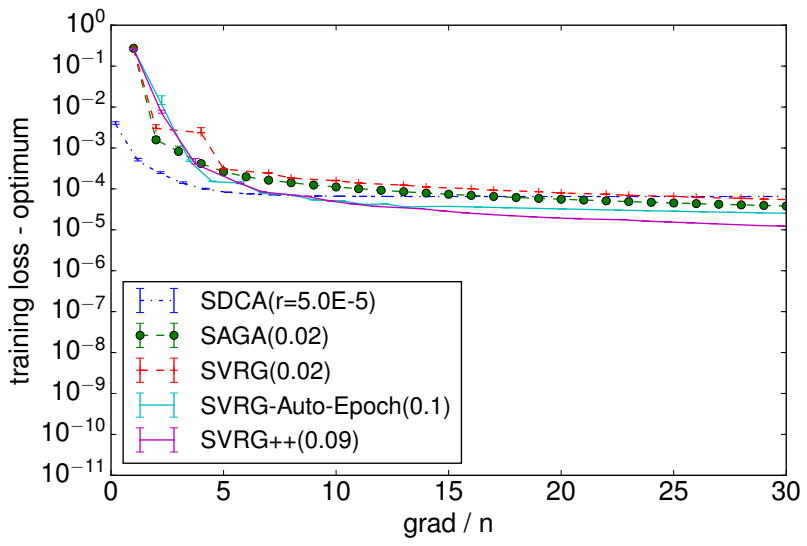

(g) Adult, Lasso $\sigma = 10^{-5}$

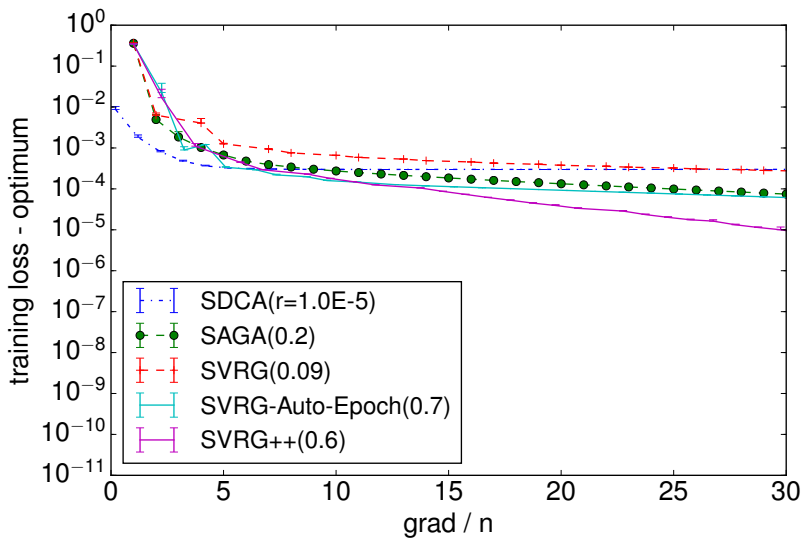

(h) Adult, Logistic $\sigma = 10^{-5}$

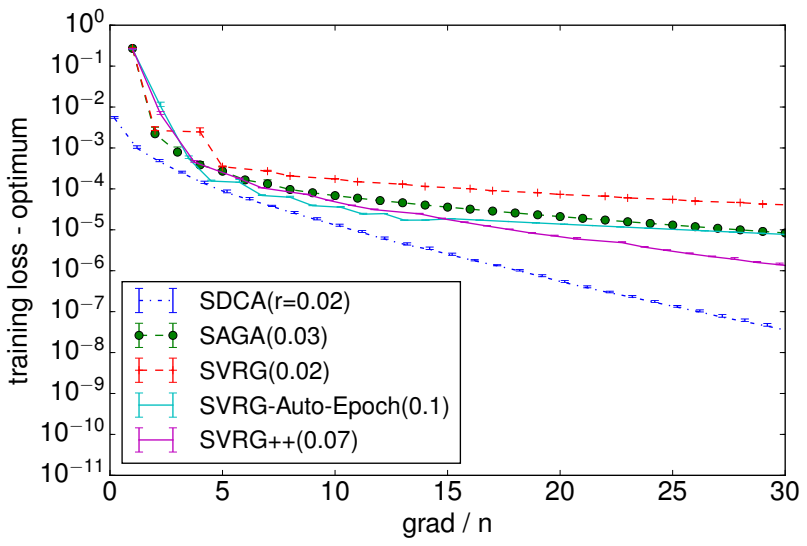

(i) Adult, Ridge $\sigma = 10^{-5}$

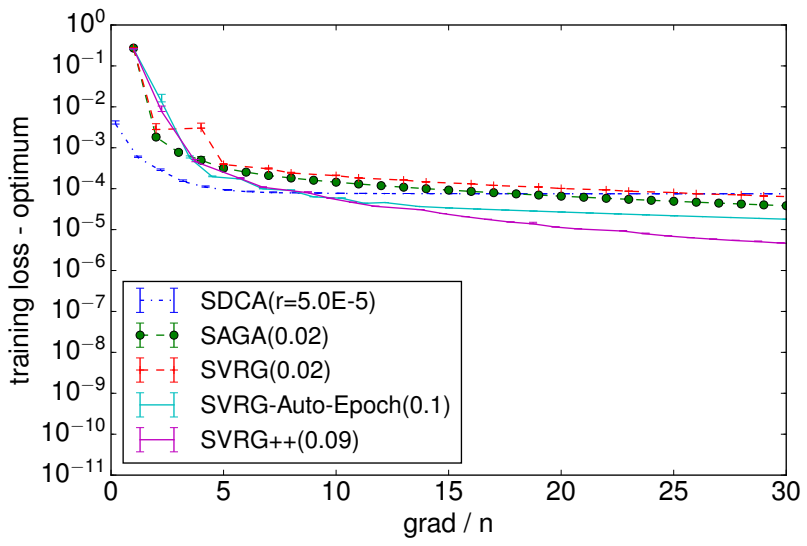

(j) Adult, Lasso $\sigma = 10^{-6}$

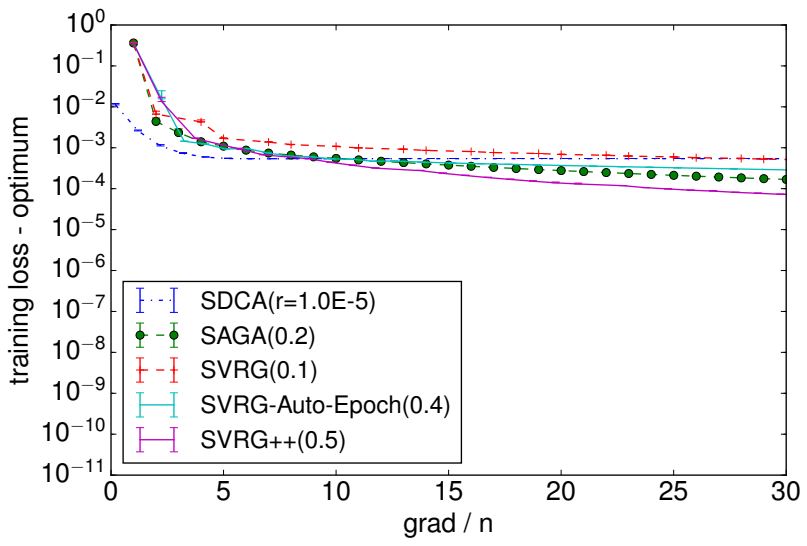

(k) Adult, Logistic $\sigma = 10^{-6}$

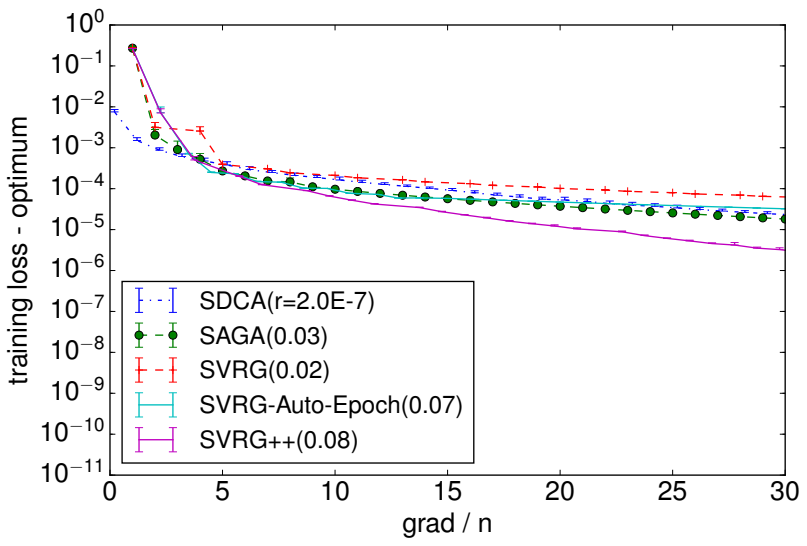

(l) Adult, Ridge $\sigma = 10^{-6}$

Figure 7: Training error comparisons on dataset Adult, using Tuning Type II.

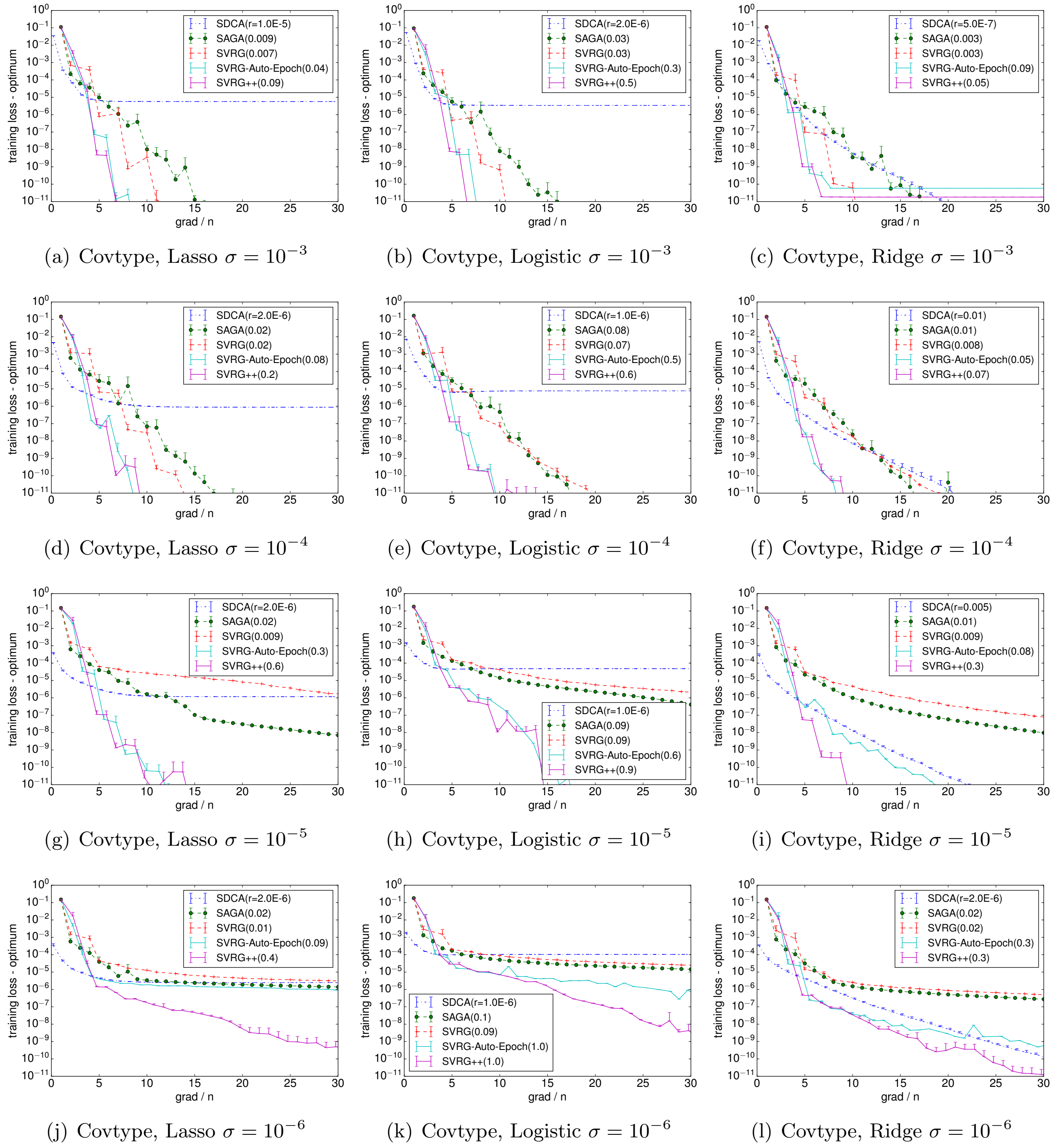

(a) Covtype, Lasso $\sigma = 10^{-3}$ (b) Covtype, Logistic $\sigma = 10^{-3}$ (c) Covtype, Ridge $\sigma = 10^{-3}$

(d) Covtype, Lasso $\sigma = 10^{-4}$ (e) Covtype, Logistic $\sigma = 10^{-4}$ (f) Covtype, Ridge $\sigma = 10^{-4}$

(g) Covtype, Lasso $\sigma = 10^{-5}$ (h) Covtype, Logistic $\sigma = 10^{-5}$ (i) Covtype, Ridge $\sigma = 10^{-5}$

(j) Covtype, Lasso $\sigma = 10^{-6}$ (k) Covtype, Logistic $\sigma = 10^{-6}$ (l) Covtype, Ridge $\sigma = 10^{-6}$

Figure 8: Training error comparisons on dataset Covtype, using Tuning Type II.

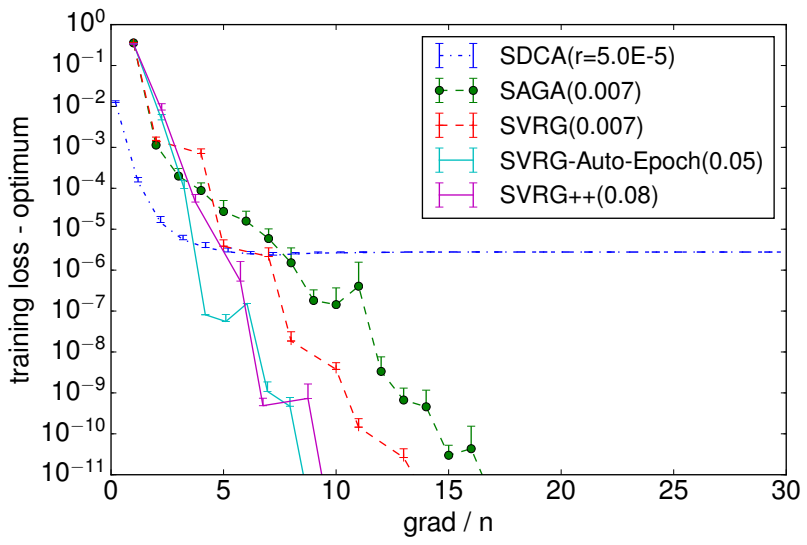

(a) Ijcnn1, Lasso $\sigma = 10^{-3}$

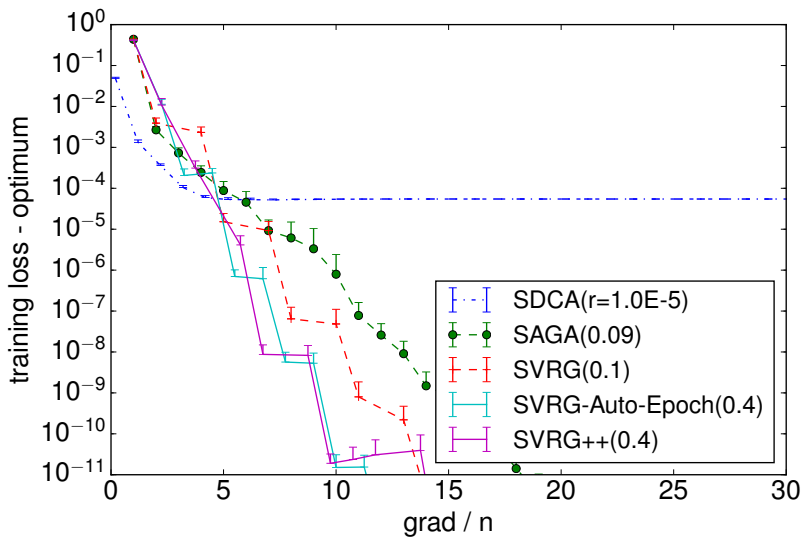

(b) Ijcnn1, Logistic $\sigma = 10^{-3}$

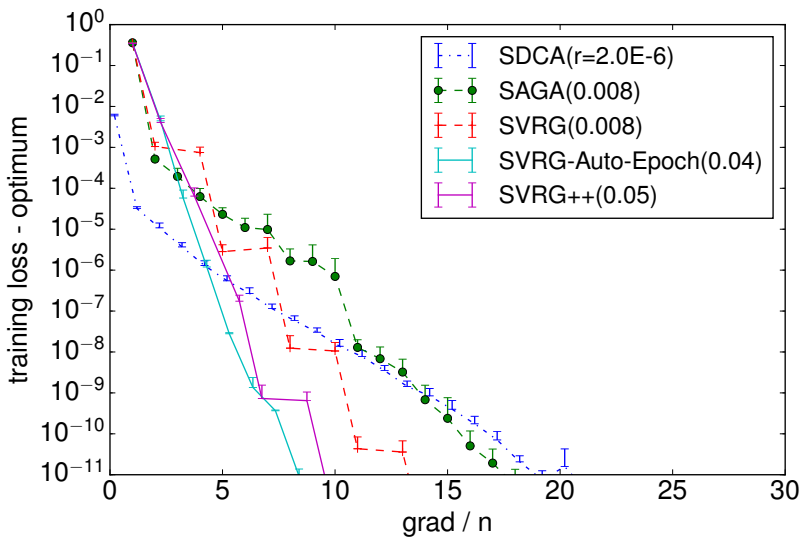

(c) Ijcnn1, Ridge $\sigma = 10^{-3}$

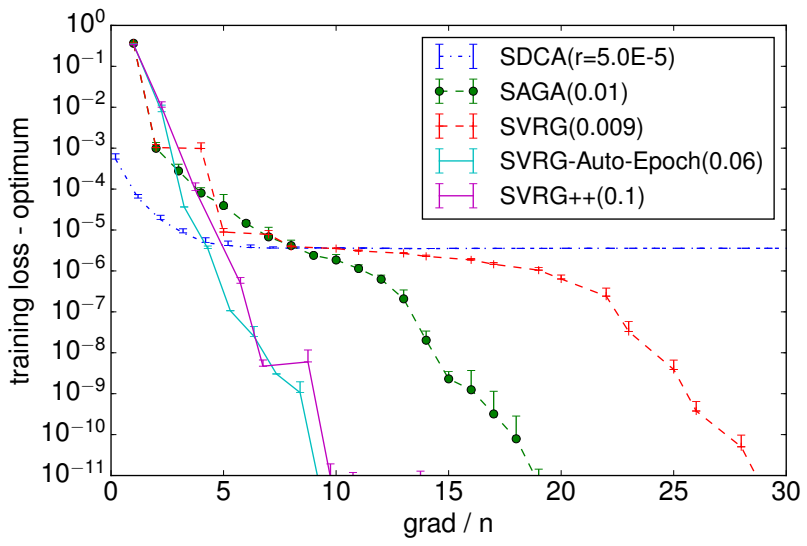

(d) Ijcnn1, Lasso $\sigma = 10^{-4}$

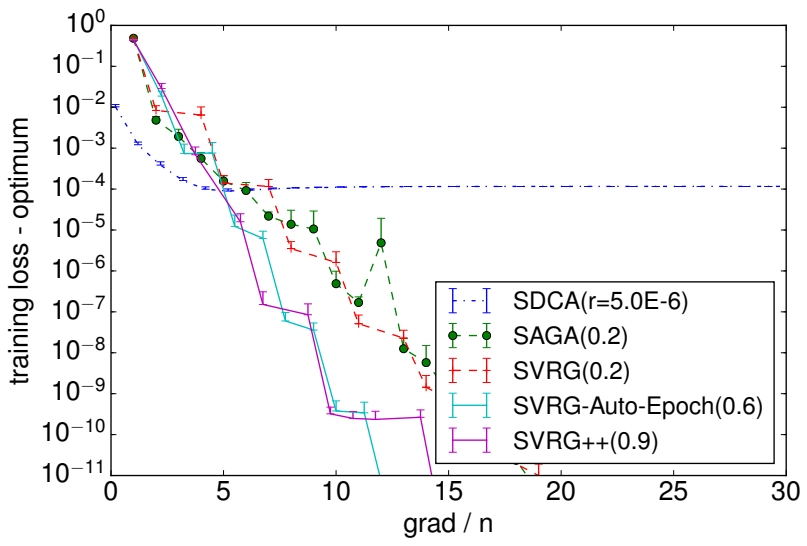

(e) Ijcnn1, Logistic $\sigma = 10^{-4}$

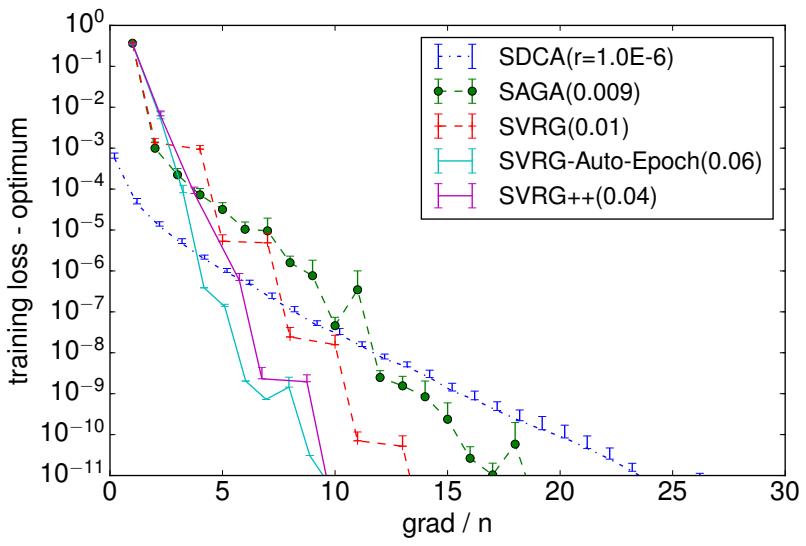

(f) Ijcnn1, Ridge $\sigma = 10^{-4}$

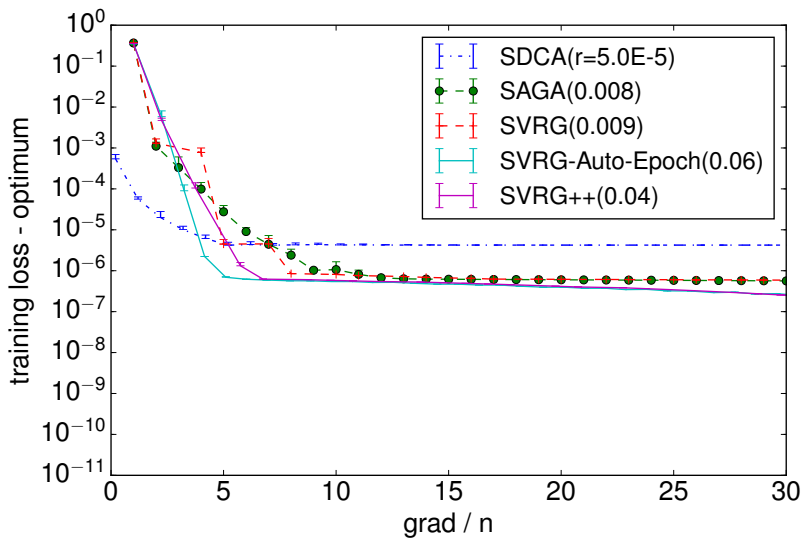

(g) Ijcnn1, Lasso $\sigma = 10^{-5}$

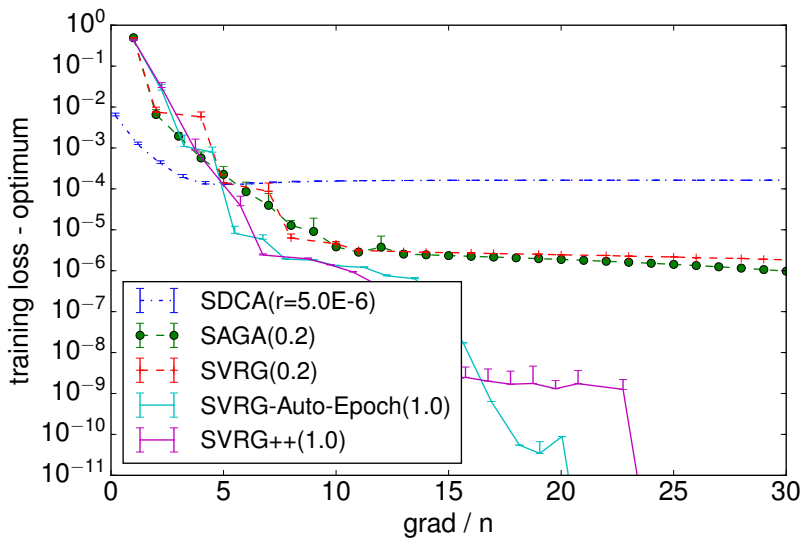

(h) Ijcnn1, Logistic $\sigma = 10^{-5}$

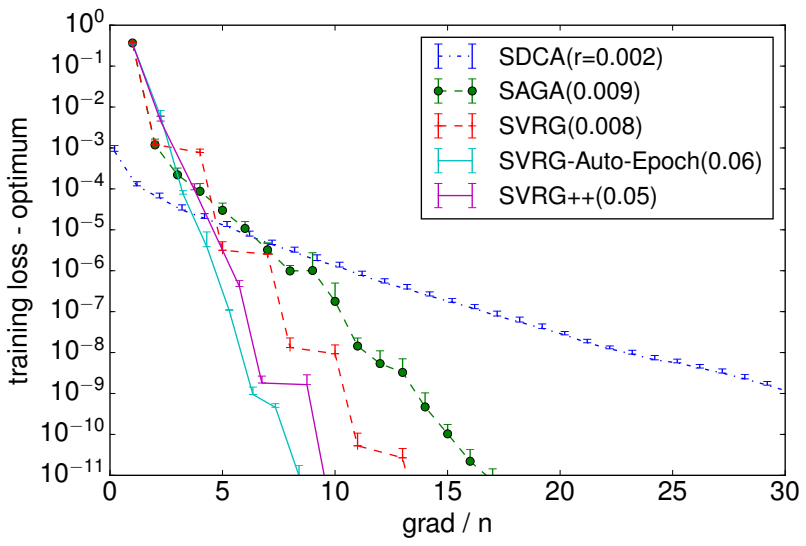

(i) Ijcnn1, Ridge $\sigma = 10^{-5}$

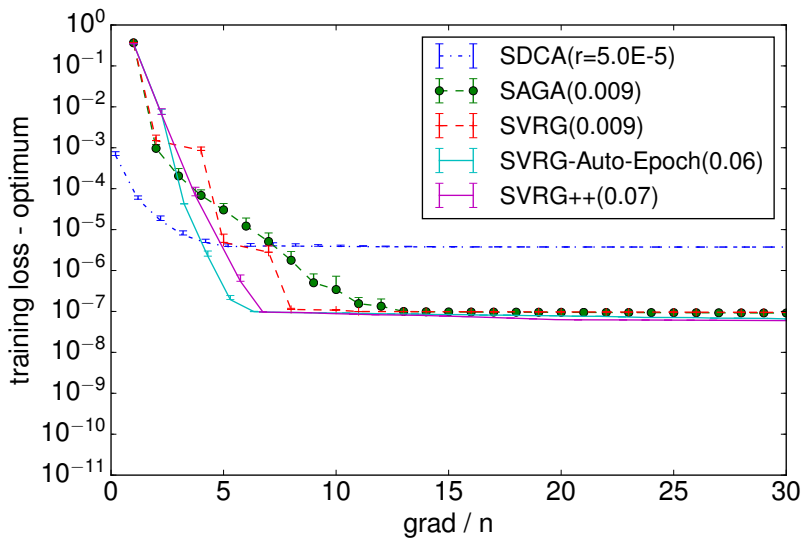

(j) Ijcnn1, Lasso $\sigma = 10^{-6}$

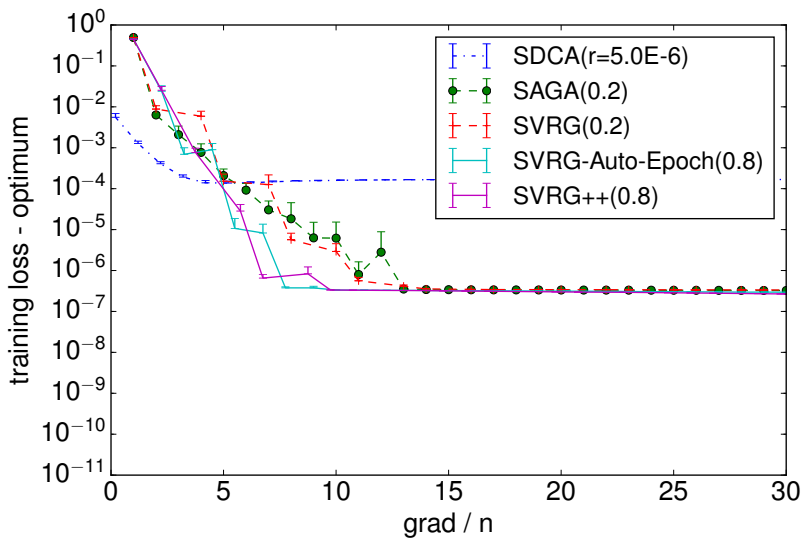

(k) Ijcnn1, Logistic $\sigma = 10^{-6}$

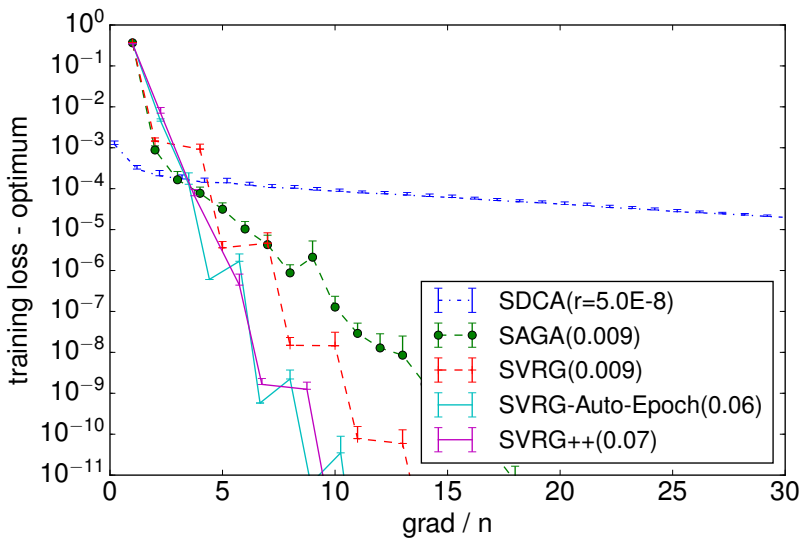

(l) Ijcnn1, Ridge $\sigma = 10^{-6}$

Figure 9: Training error comparisons on dataset Ijcnn1, using Tuning Type II.

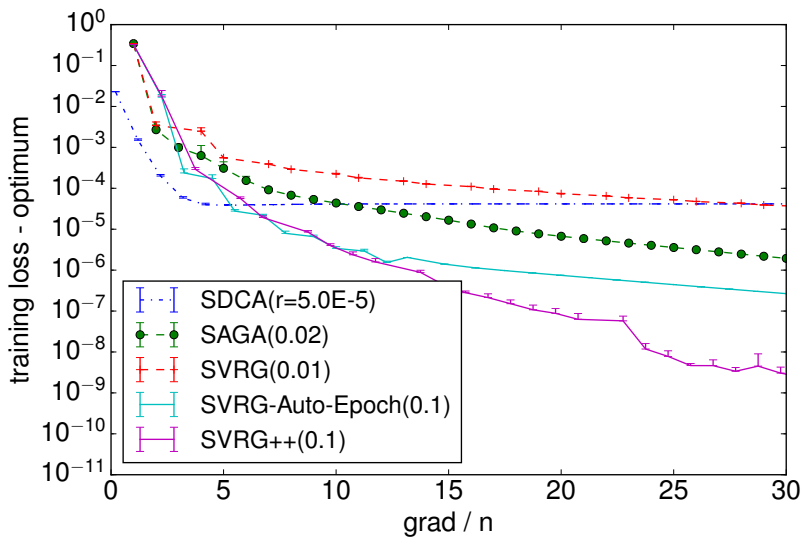

(a) Mnist, Lasso $\sigma = 10^{-3}$

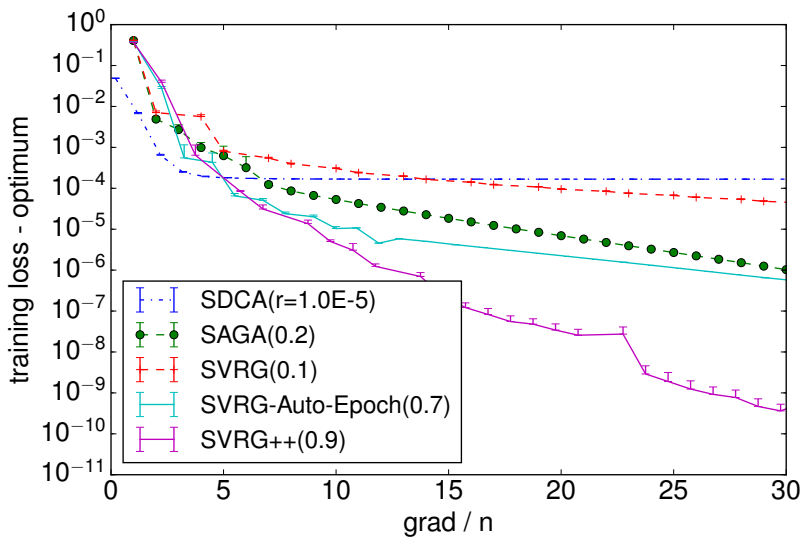

(b) Mnist, Logistic $\sigma = 10^{-3}$

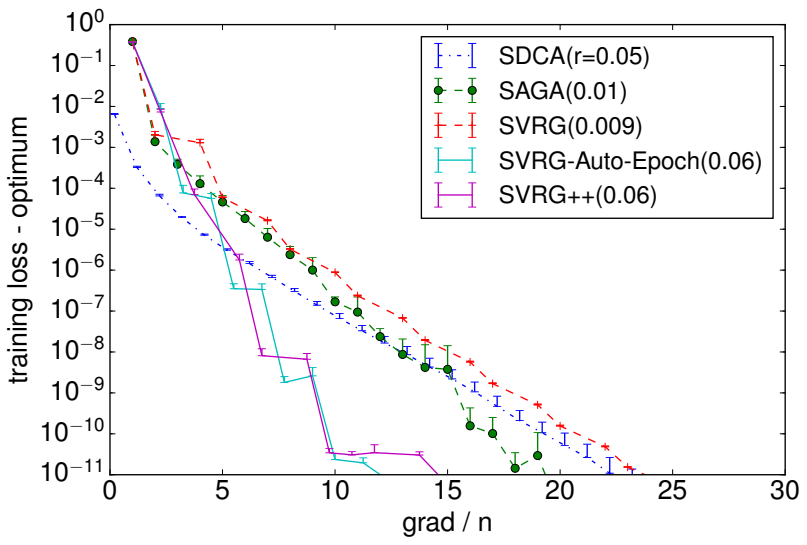

(c) Mnist, Ridge $\sigma = 10^{-3}$

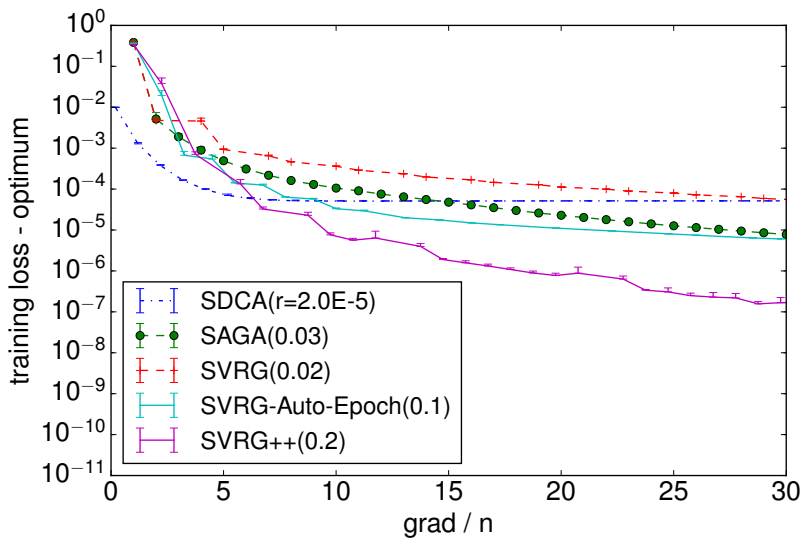

(d) Mnist, Lasso $\sigma = 10^{-4}$

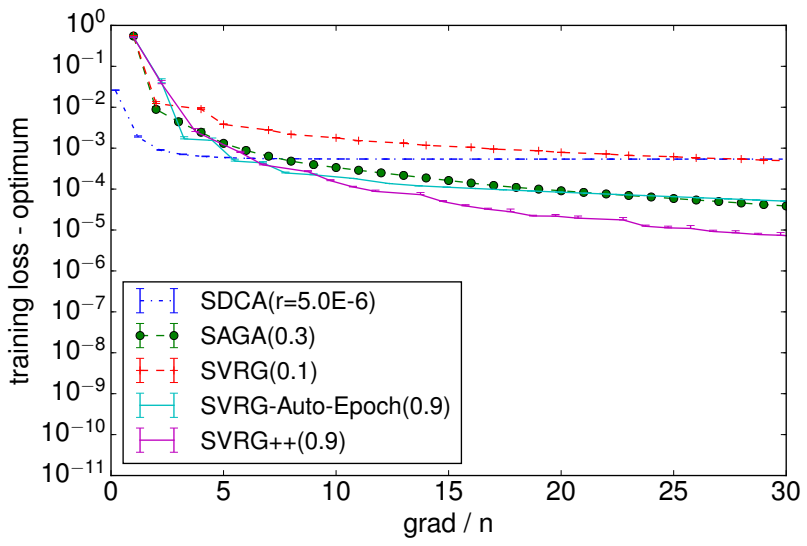

(e) Mnist, Logistic $\sigma = 10^{-4}$

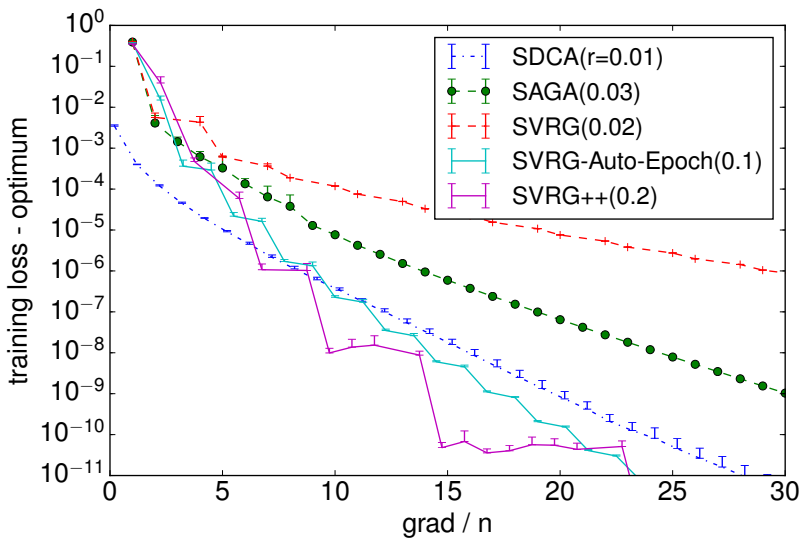

(f) Mnist, Ridge $\sigma = 10^{-4}$

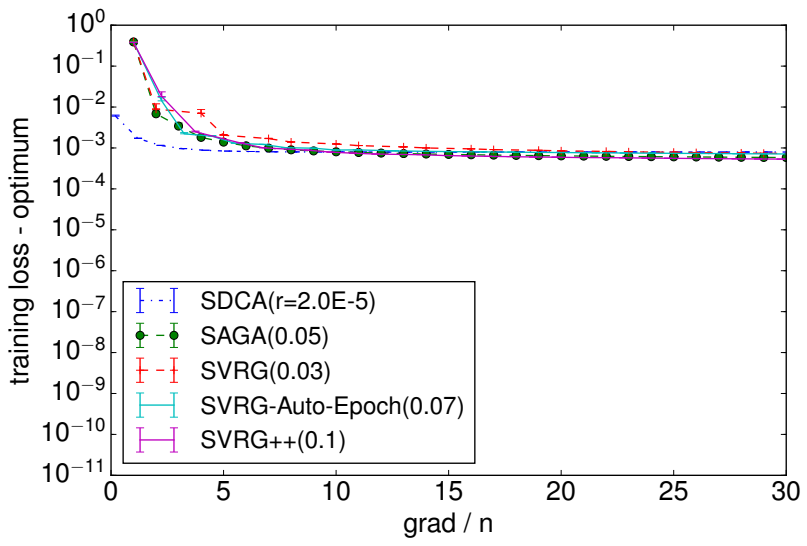

(g) Mnist, Lasso $\sigma = 10^{-5}$

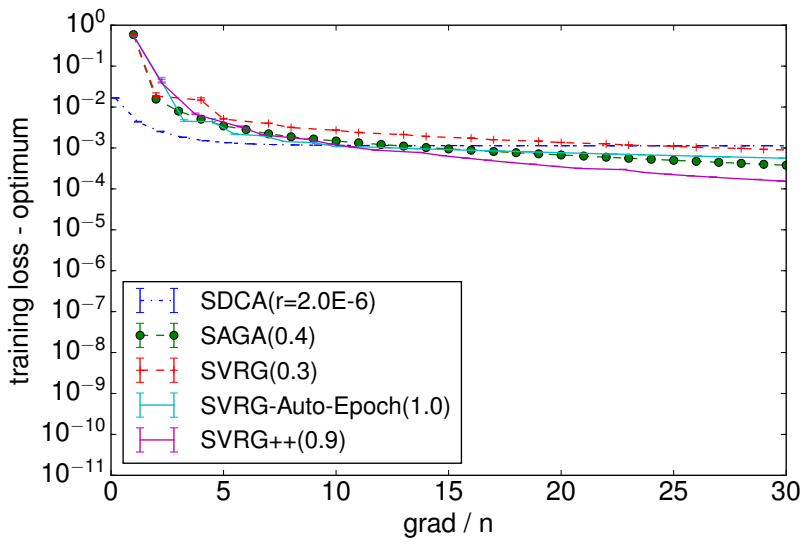

(h) Mnist, Logistic $\sigma = 10^{-5}$

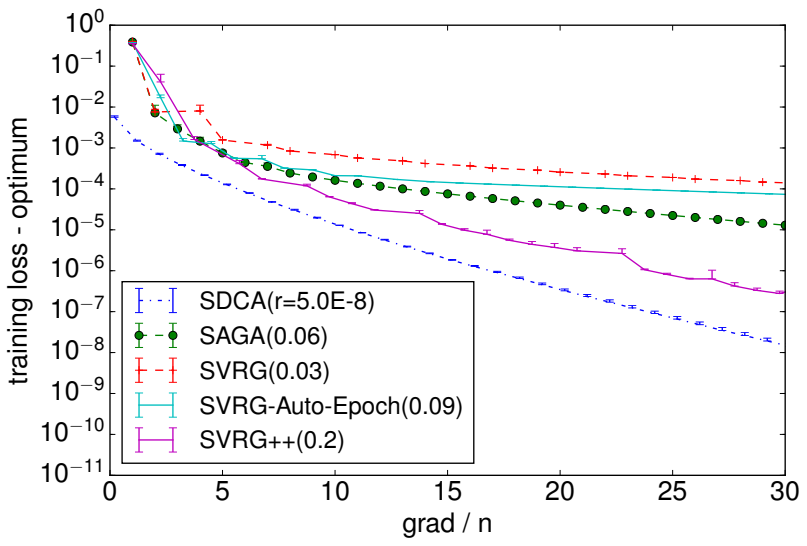

(i) Mnist, Ridge $\sigma = 10^{-5}$

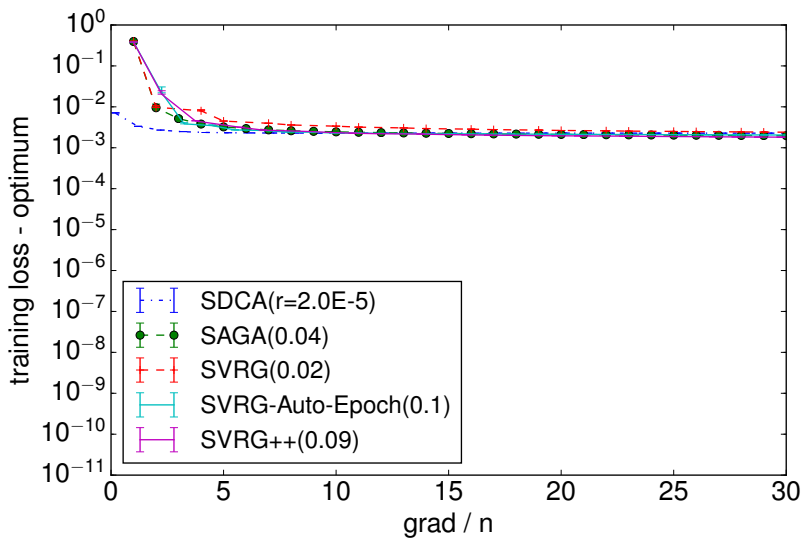

(j) Mnist, Lasso $\sigma = 10^{-6}$

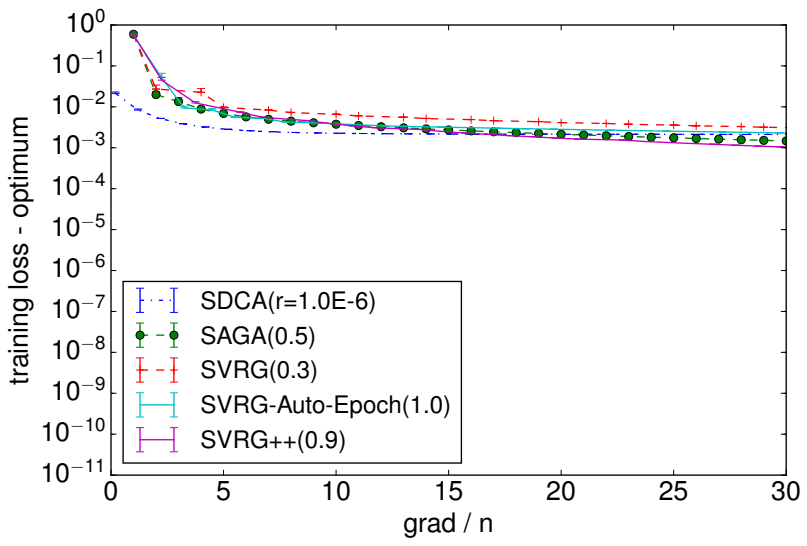

(k) Mnist, Logistic $\sigma = 10^{-6}$

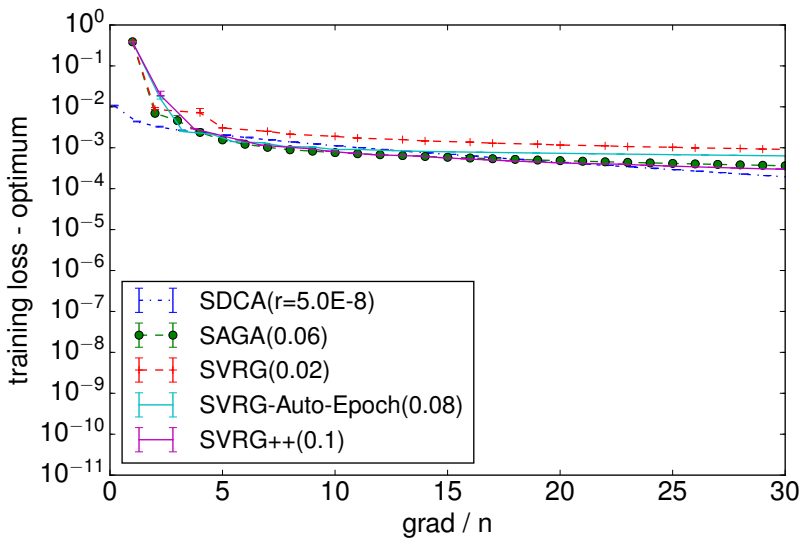

(l) Mnist, Ridge $\sigma = 10^{-6}$

Figure 10: Training error comparisons on dataset mnist, using Tuning Type II.